\documentclass{article}

\usepackage{arxiv}

\usepackage[utf8]{inputenc} 
\usepackage[T1]{fontenc}    
\usepackage{hyperref}       
\usepackage{url}            
\usepackage{booktabs}       
\usepackage{amsfonts}       
\usepackage{nicefrac}       
\usepackage{microtype}      
\usepackage{lipsum}		
\usepackage{graphicx}
\usepackage{doi}
\usepackage{caption}
\usepackage{listings}
\usepackage{xcolor}
\usepackage{fancyhdr}       
\usepackage{graphicx}       
\usepackage{fancyvrb}        
\usepackage{pdfpages}
\usepackage{hyperref} 
\usepackage{caption} 
\usepackage{amsmath}        
\usepackage{amsfonts}       
\usepackage{amssymb}        
\usepackage{algorithm}      
\usepackage{algpseudocode}
\usepackage{amsmath,amssymb}
\usepackage{adjustbox}
\usepackage{minted}

\usepackage{setspace} 
  \usepackage{enumerate}      
\usepackage{soul}
\usepackage{amsmath}
\usepackage{listings}
\usepackage{caption}
\usepackage{tabularx}

\usepackage{adjustbox}
\usepackage{subcaption}  

\usepackage{multirow}
\usepackage{helvet} 
\usepackage{tikz}
\usetikzlibrary{
  shapes.geometric,
  shapes.symbols,   
  arrows,
  arrows.meta,
  calc,
  positioning,
  fit,
  backgrounds,
  shadows,
  shapes.multipart,
}

\usetikzlibrary{decorations.pathreplacing}

\newcounter{textbox}
\renewcommand{\thetextbox}{\arabic{textbox}} 

\usepackage{sansmath} 

\tikzstyle{process} = [rectangle, rounded corners, minimum width=3cm, minimum height=1cm, text centered, draw=black, fill=orange!30]
\tikzstyle{start} = [ellipse, minimum width=3cm, minimum height=1cm, text centered, draw=black, fill=yellow!30]
\tikzstyle{decision} = [diamond, minimum width=3cm, minimum height=1cm, text centered, draw=black, fill=green!30]
\tikzstyle{arrow} = [thick,->,>=stealth]

\tikzstyle{phase} = [rectangle, rounded corners, minimum width=3cm, minimum height=1cm, text centered, draw=black, fill=orange!30]
\tikzstyle{result} = [ellipse, minimum width=3cm, minimum height=1cm, text centered, draw=black, fill=green!30]

\tikzstyle{dataset} = [rectangle, rounded corners, minimum width=2cm, minimum height=1cm, text centered, draw=black, fill=blue!30]
\tikzstyle{masking} = [rectangle, rounded corners, minimum width=2cm, minimum height=1cm, text centered, draw=black, fill=red!30]

\usepackage[skins,breakable]{tcolorbox}

\definecolor{aigold}{RGB}{244,210, 1} 
\definecolor{aigreen}{RGB}{210,244,211} 

\sethlcolor{aigreen}

\definecolor{aired}{RGB}{255,180,181}

\definecolor{aigold}{RGB}{255,180,181}

\definecolor{aiblue}{RGB}{173,216,230} 
 
\definecolor{lightred}{rgb}{1,0.9,0.9} 

\tcbset{
  customboxsmalll/.style={
    width=0.95\textwidth,  
    fonttitle=\ttfamily\scriptsize, 
    colback=white,  
    colframe=black,  
    colbacktitle=blue!75!black, 
    coltitle=white,  
    boxrule=0.5pt,  
    rounded corners,  
    boxed title style={
        colback=blue!75!black, 
        colframe=black, 
        boxrule=0.5pt  
            before={\setlength{\parskip}{0pt} \setlength{\parindent}{-6pt}},  
    after={\setlength{\parskip}{0pt} \setlength{\parindent}{-6pt}},   
    halign=center,  
    before skip=-5pt,  
    after skip=-5pt,   
    innertop=-3pt,     
    innerbottom=-3pt,  
    valign=center,    
    }
    
  }
}
\usepackage{enumitem}  

\tcbset{
  custombox/.style={
    width=512.18663pt,
    top=3pt,
    fonttitle=\bfseries\scriptsize\sffamily, 
    colback=white,
    colframe=black,
    colbacktitle=black,
    boxrule=0.5pt, 
    colback=white, 
    enhanced,
    rounded corners, 
    center,
    boxed title style={
            sharp corners,
            size=small,
            colframe=blue!75!black, 
        },
    attach boxed title to top left={yshift=-0.1in,xshift=0.15in},
    boxed title style={boxrule=0pt,colframe=white,},
    before=\setlength{\baselineskip}{2pt}, 
    before upper=\setlength{\baselineskip}{2pt}, 
    after=\setlength{\baselineskip}{2pt}, 
  }
} 

\newtcolorbox{LLMbox}[2][]{custombox, title=#2,#1}
\newtcolorbox{LLMboxSmall}[2][]{customboxsmalll, title=#2,#1}

\DefineVerbatimEnvironment{thinkingcolor}{Verbatim}{
  formatcom=\color{blue}, 
  fontsize=\scriptsize  
}
\DefineVerbatimEnvironment{reflectcolor}{Verbatim}{
  formatcom=\color{orange}, 
  fontsize=\scriptsize  
}

\title{\textbf{MusicSwarm: Biologically Inspired Intelligence for Music Composition}}

\author{ \href{https://orcid.org/0000-0002-4173-9659}{\includegraphics[scale=0.06]{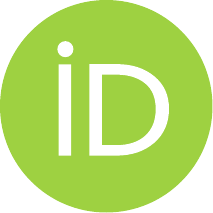}\hspace{1mm}Markus J. Buehler}\thanks{Corresponding author.} \\
	Laboratory for Atomistic and Molecular Mechanics\\Center for Computational Science and Engineering \\
        Schwarzman College of Computing \\
	Massachusetts Institute of Technology\\
	Cambridge, MA 02139, USA \\
        \\
	\texttt{mbuehler@MIT.EDU} \\
}

\newtcbox{\mybox}[1][green]{on line,
arc=0pt,outer arc=0pt,colback=#1!10!white,colframe=#1!50!black,
boxsep=0pt,left=0pt,right=0pt,top=0pt,bottom=0pt,
boxrule=0pt,bottomrule=0pt,toprule=0pt}

\tcbset{
  customboxmultipage/.style={
    width=512.18663pt,
    top=3pt,
    fonttitle=\ttfamily\scriptsize, 
    colback=white,
    breakable,
    colframe=black,
    colbacktitle=black,
    boxrule=0.5pt,
    rounded corners,
    center,
    boxed title style={
        sharp corners,
        size=small,
        colframe=blue!75!black,
    },
    attach boxed title to top left={yshift=-0.1in,xshift=0.15in},
    boxed title style={boxrule=0pt,colframe=white,},
    before={\par\noindent},
    after={\par},
    before upper={\setstretch{0.3}}, 
    after lower={\nointerlineskip},
    every page on break/.style={
      interior style={top color=white, bottom color=white},
    },
  }
}

\newtcolorbox{LLMboxmultipage}[2][]{customboxmultipage,title=#2,#1}

\renewcommand{\headeright}{}
\renewcommand{\shorttitle}{MusicSwarm: Biologically Inspired Intelligence for Composition}

\hypersetup{
pdftitle={
Autonomous Knowledge Gardening at Scale},
pdfsubject={q-bio.NC, q-bio.QM},
pdfauthor={Markus J. Buehler, MIT, mbuehler@MIT.EDU},
pdfkeywords={Language Modeling, Graph Theory, Category Theory, Materials Science, Materiomics, Symbolic Reasoning},
}

\begin{document}
\maketitle
 
\begin{abstract} 
We show that coherent, long‑form musical composition can emerge from a decentralized swarm of identical, frozen foundation models that coordinate via stigmergic, peer‑to‑peer signals, without any weight updates. We compare a centralized multi‑agent system with a global critic to a fully decentralized swarm in which bar‑wise agents sense and deposit harmonic, rhythmic, and structural cues, adapt short‑term memory, and reach consensus. Across symbolic, audio, and graph‑theoretic analyses, the swarm  yields superior quality while delivering greater diversity and structural variety and leads across creativity metrics. The dynamics contract toward a stable configuration of complementary roles, and self‑similarity networks reveal a small‑world architecture with efficient long‑range connectivity and specialized bridging motifs, clarifying how local novelties consolidate into global musical form. By shifting specialization from parameter updates to interaction rules, shared memory, and dynamic consensus, MusicSwarm provides a compute‑ and data‑efficient route to long‑horizon creative structure that is immediately transferable beyond music to collaborative writing, design, and scientific discovery.
\end{abstract}


\section{Introduction}
Much progress has been made in using Transformer-based models, including language models, attention-diffusion approaches, or flow-matching to a variety of domain-specific use cases~\cite{Vaswani2017AttentionNeed,Ouyang2022RLHF,Ziegler2020RLHF,Asghar2024,devlin2018bert,Buehler2023MechGPTModalities_updated,buehler2023biomateriomics,Ghafarollahi2024ProtAgents:Learning,Ni2023ForceGen:Model, Dhariwal2021DiffusionSynthesis,ghafarollahi2024sciagentsautomatingscientificdiscovery}. Building on these successes, artificial intelligence has demonstrated remarkable capabilities in creative domains, with large language models (LLMs) achieving significant success in text generation, code synthesis, and increasingly, musical composition \cite{Roberts2018AMusic, Briot_2018, Oore2018ThisPerformance, huang2020musenet, donahue2019lakhnes, eck2002finding, CIVIT2022118190}. However, most AI music generation systems rely on monolithic architectures where a single model learns to generate entire compositions through supervised learning on large corpora \cite{briot2017deep, ji2020comprehensive}. While such approaches can produce coherent musical outputs, they fundamentally differ from how human musical creativity emerges through collaborative processes, distributed expertise, and emergent group dynamics observed in ensembles, orchestras, and compositional partnerships \cite{sawyer2003group,  keith2011psychological}. 

In recent years, a number of researchers have explored the use of multi-agent systems in the realm of generative music. For instance, work reported in \cite{smith2020swarmmusic} demonstrated how agentic algorithms could be applied to create evolving musical compositions, leveraging the self-organizing behavior of components to produce complex soundscapes. Others~\cite{lee2018flockingmusic} explored the concept of generating music from flocking dynamics, where autonomous agents following simple behavioral rules could create musically interesting results. These studies highlight the potential of multi-agent systems to foster emergent creativity in musical composition, offering a unique approach to the generation of complex and dynamic musical pieces.

\begin{figure}[h!]
    \centering
    \includegraphics[width=.8\textwidth]{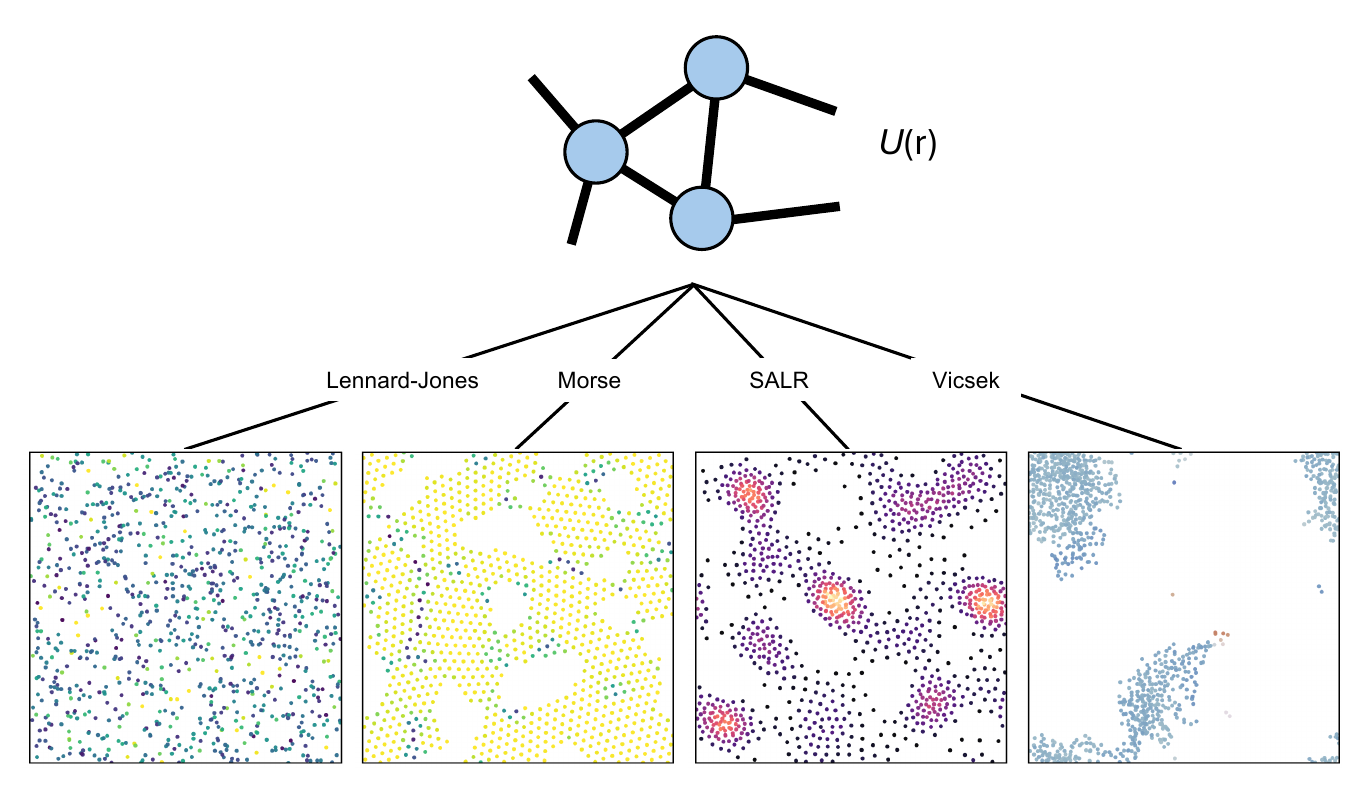}
    \caption{Self-assembly of $N=1024$ particles under identical global conditions but different local rules~\cite{Allen1987ComputerLiquids,Buehler2008AtomisticFailure}.
    Each panel shows the final configuration colored by the relevant local order parameter: hexatic order \(|\psi_6|\) for Lennard--Jones and Morse, local density for SALR, and heading angle for Vicsek (see Methods). 
    Lennard-Jones: largely disordered with weak crystallinity, reflected in a flat radial distribution function \(g(r)\) near unity \cite{LennardJones1924}. 
    Morse: extended crystalline grains with strong hexatic signatures and long-ranged oscillations in \(g(r)\) \cite{Morse1929}. 
    SALR: microphase clustering into dense spots and incipient stripes driven by competing short-range attraction and long-range repulsion, yielding a split-peak structure in \(g(r)\) \cite{SeulAndelman1995,Stradner2004}. 
   Vicsek: nonequilibrium flocking clusters formed by alignment interactions with noise, producing broad peaks in \(g(r)\) without crystalline order \cite{Vicsek1995}. 
    The comparison underscores how radically different global organizations can emerge solely from changes in local interaction laws (see Figure~\ref{fig:interactions_analogy} for a visualization of the interactions that drive the four distinct behaviors). Please see Figure~\ref{fig:g_r_result} for associated Radial Distribution Function results. }
    \label{fig:selfassembly}
\end{figure}

To go beyond currently widely used strategies (e.g., pretraining followed by task-specific fine-tuning, RLHF-style alignment, or test-time compute)~\cite{Ouyang2022RLHF,Ziegler2020RLHF,zelikman2024quietstarlanguagemodelsteach, Hu2021LoRA:Models,Buehler2025PRefLexOR,devlin2018bert,Buehler2025PRefLexOR,o1-model-card-2024,o3-mini-model-card-2025}, in this paper we revisit early traditions of distributed cognition, such as Minsky’s Society of Mind and the rule-based MYCIN expert system, framed intelligence as cooperation among simple parts \cite{minsky1986society,shortliffe1976mycin}. We reinterpret that thesis with foundation models, canonically based on LLMs here, as basic constituents: these form role-conditioned agents, shared memory, and peer/environmental feedback yield coordinated behavior without weight updates, shifting the locus of learning from parameters to system organization~\cite{brown2020language,ouyang2022training,moor2023foundation}.
This  resonates also with Gödel’s incompleteness theorem~\cite{godel1931formal}, which showed that no formal system can be both consistent and complete; by analogy, no monolithic model can exhaust the space of creative possibilities, motivating architectures that extend themselves through interaction and feedback organized in a larger set.
This strategy also relates with earlier work~\cite{Supper2001AComposition,Schuijer2008AnalyzingContexts,Tymoczko2011APractice,Franjou2021ANanostructures,Yu2019AAI,Milazzo0BioinspiredModeling,Buehler2023UnsupervisedDesigns} that focused on musical composition interpreted via atomic-to-composition processes inspired by category theory~\cite{Giesa2011ReoccurringAnalogies,Spivak2011CategoryNetworks,Eilenberg1945GeneralEquivalences_updated,Marquis2019CategoryTheory}.  
We speculate that a swarm approach based on stigmergic coordination - a principle of indirect communication observed in social insects, where individuals leave traces in a shared environment (e.g., pheromone trails) that shape the actions of others - can be a useful strategy to realize self-organizing behavior. In the work presented here, a swarm of agents deposit and sense musical cues in a similar way, enabling decentralized yet coherent organization without central control.

Recent advances in multi-agent modeling have indeed shown that complex behaviors can emerge from simple local interactions without centralized control, demonstrating principles of swarm intelligence observed in biological systems~\cite{tampuu2017multiagent, foerster2018emergent, sunehag2017value, ghafarollahi2025sparksmultiagentartificialintelligence,ghafarollahi2025autonomousinorganicmaterialsdiscovery,buehler2023biomateriomics}. These distributed approaches contrast sharply with traditional fine-tuning paradigms that adapt pre-trained models through gradient-based optimization on task-specific datasets~\cite{howard2018universal, devlin2018bert}. Instead of relying on massive parameter updates across entire neural networks, multi-agent systems enable specialized agents  to develop distinct competencies while learning to coordinate through environmental interactions and peer feedback \cite{stone2000multiagent, weiss2013multiagent}. This paradigm shift from monolithic learning to distributed intelligence offers promising avenues for understanding emergent creativity and collaborative problem-solving in artificial systems. 

\subsection{Motivation from Statistical Mechanics}
To motivate our framework we draw on an analogy from statistical physics: a simple box of interacting particles. When all global conditions are fixed - number of particles, density, integration scheme, and annealing schedule -the only determinant of large-scale organization is the local interaction rule (Figure~\ref{fig:interactions_analogy} shows visualizations of the interactions, further details, see Supplementary Materials). Changing this rule alone suffices to yield dramatically different global outcomes, as shown in Figure~\ref{fig:selfassembly}: The Lennard-Jones system remains largely disordered under the chosen parameters, with weak signatures of hexatic order and a nearly flat radial distribution function \(g(r)\) close to unity, reflecting the absence of strong crystallization. In contrast, the Morse potential yields extended crystalline patches with pronounced hexatic order, confirmed by sharp oscillations in \(g(r)\) that persist to long range. The SALR case shows the expected clustering into microphases driven by competing short--range attraction and long--range repulsion, manifested both as localized density hotspots and as a split--peak structure in \(g(r)\), consistent with prior reports of stripe and cluster morphologies. Finally, the Vicsek model produces nonequilibrium flocking clusters: particles align into transient bands, and the corresponding \(g(r)\) displays a broad peak structure indicative of intermediate-range correlations without crystalline order. Together, these outcomes underscore the central message of the analogy: the same ``substrate'' of agents or particles can yield radically different global organizations solely through changes in their local update rules.

\begin{figure}
\centering
\sffamily
\scriptsize

\begin{tikzpicture}[
    >=Latex,
    node distance=0.8cm and 1cm,
    box/.style={rectangle, draw, rounded corners, align=center, minimum height=1.1cm, inner sep=3pt, text width=3.8cm},
    stepA/.style={box, fill=blue!15},
    stepB/.style={box, fill=green!15},
    stepC/.style={box, fill=yellow!20},
    stepD/.style={box, fill=orange!25},
    hyp/.style={rectangle, draw, very thick, rounded corners, align=center, inner sep=6pt, text width=5.0cm, fill=white},
    flow/.style={-Stealth, thick}
]

\begin{scope}[shift={(-2.2cm,0)}]   

\def\r{3.2cm}
\node[stepA] (S1) at (0,1.5) 
  {\textbf{Swarm of Agents}\\(instances with unique features)};
\node[stepB] (S2) at (3.8,0)
  {\textbf{Agents Act}\\(compose)};
\node[stepC] (S3) at (0,-1.5) 
  {\textbf{Feedback}\\(peer or critic evaluation\\ interact, sense, ...)  };
\node[stepD] (S4) at (-3.8,0) 
  {\textbf{Policy Update}\\(adapt roles, prompts, ...)};

\draw[flow] (S1) -- (S2);
\draw[flow] (S2) -- (S3);
\draw[flow] (S3) -- (S4);
\draw[flow] (S4) -- (S1);

\begin{pgfonlayer}{background}
  \node[draw=orange, dashed, very thick, rounded corners,
        fit=(S1)(S2)(S3)(S4), inner sep=12pt,
        label={[text=orange]below:\textbf{Online orchestration loop (no weight updates)}}] (ORCH) {};
\end{pgfonlayer}

\node[align=center, text=violet!80!black, font=\bfseries] (EM)
  at (0,0) {Emergent Specialization};
  
\node[shape=cylinder, draw=black, thick, shape border rotate=90, aspect=0.28,
      minimum height=1.9cm, minimum width=1.7cm, fill=blue!10, right=2cm of S2, inner sep=1.5pt] (MEM)
      {\parbox{1.5cm}{\centering \scriptsize \textbf{Episodic}\\[-1pt]\scriptsize \textbf{Memory}}};

\draw[flow] (MEM.west) -- (ORCH.east);

\node[box, fill=gray!15, text width=3.8cm, above=1.6cm of ORCH.north] (LLM)
  {\textbf{Foundation LLM}\\(static, pretrained)};
\draw[flow] (LLM.south) -- (S1.north);
\end{scope}
\node[align=left, font=\bfseries] (Tconv) at (-6.2,-4.5) {Conventional pipeline:};

\node[box, fill=gray!15, below=0.8cm of Tconv.east, anchor=east, text width=3.4cm] (P0)
  {Pre-train on large corpora};
\node[box, fill=gray!15, right=0.6cm of P0, text width=3.4cm] (P1)
  {Post train or \\ fine-tune on task data};
\node[box, fill=gray!15, right=0.6cm of P1, text width=3.4cm] (P2)
  {Deploy monolithic model};

\draw[flow] (P0) -- (P1);
\draw[flow] (P1) -- (P2);

\end{tikzpicture}

\rmfamily
\caption{Key hypothesis and simplified orchestration cycle to explain the overall strategy explored in this paper. The pretrained foundation LLM spawns a swarm of agents with unique features, and which a cabability to evolve their behavior. Agents act (compose) and produce outputs. Their outputs undergo feedback (peer and critic evaluation), which drives policy updates (adapting roles, prompts, ...). This in turn updates the behavior of the swarm. Through repeated iterations, emergent specialization arises as a property of the cycle. Episodic memory informs the entire loop. For contrast, bottom shows a conventional pre-train$\rightarrow$fine-tune$\rightarrow$deploy pipeline.}
\label{fig:key_hypothesis}
\end{figure}
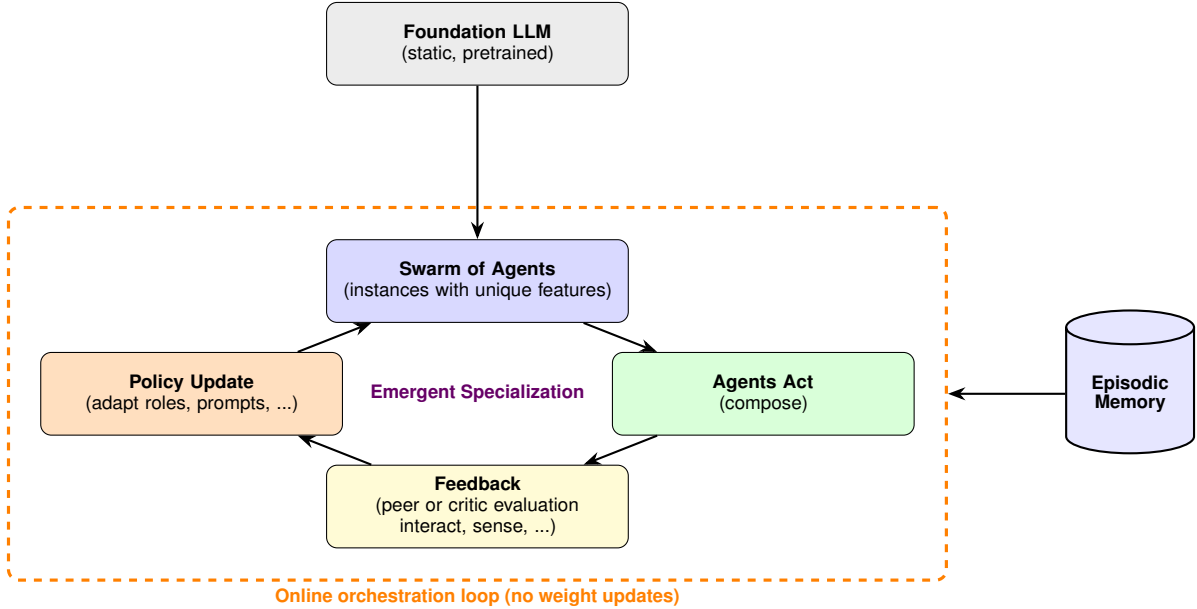

\subsection{Research Design}

The central motivation of this work is to explore whether general-purpose foundation models, such as frontier  large language models (LLMs) such as \texttt{GPT-5} and its variants~\cite{OpenAI2025GPT5SystemCard}, can be harnessed for highly specialized creative tasks without the need 
for costly fine-tuning. Current approaches to AI music generation often assume that achieving quality 
and stylistic control requires training or adapting a monolithic model on domain-specific data. Yet, 
foundation models already encode vast latent knowledge of structure, pattern, and reasoning that may 
be repurposed in real time if orchestrated appropriately. The key challenge, then, is how to mobilize 
these models in a way that allows for adaptation, role specialization, and emergent coordination 
without modifying their internal weights.

This leads us to the hypothesis that swarms of agents—each powered by the same general-purpose 
model but situated in distinct roles, exposed to different contexts, and guided by feedback and rewards—
can achieve domain-specific expertise on the fly. Rather than relying on parameter updates, 
specialization emerges through iterative reflection, interaction, and reinforcement within the swarm. 
In this view, creativity is not hard-coded into the model itself, but arises dynamically from the 
system-level organization of agents and the adaptive feedback loops they enact. 
The dynamics of role differentiation and coordination in such swarms can be interpreted as a form of distributed game, where each agent adapts relative to the strategies of others, converging toward equilibrium states reminiscent of Nash’s formulation of non-cooperative games \cite{nash1950equilibrium,nash1951noncooperative}.

By testing this hypothesis in the context of music composition, we probe a broader question about 
artificial intelligence: can distributed collectives of general-purpose models self-organize into expert 
systems for specialized tasks, thereby transforming how we think about learning, adaptation, and 
creativity in AI (Figure~\ref{fig:key_hypothesis})? In this view, emergent specialization mirrors Nash-like equilibria, that is, stable configurations of roles and strategies that arise without central control, while the continual generation of novelty echoes Gödel’s insight that no system is ever fully closed or complete.

\subsection{Paper Outline}

\begin{figure}[t]
\centering
\sffamily
\scriptsize

\def\NodeW{3.55cm} 
\def\PanelW{0.21\textwidth}

\tikzset{every picture/.style={scale=0.8, transform shape}} 

\begin{subfigure}[t]{\PanelW}
\centering
\scriptsize

\begin{tikzpicture}[
    node distance=0.45cm and 0.7cm, auto,
    line/.style={draw, -{Stealth}, shorten >=1pt},
    task/.style={rectangle, draw, fill=red!30, text width=\NodeW, text centered, rounded corners, minimum height=0.9cm},
    generation/.style={rectangle, draw, fill=cyan!30, text width=\NodeW, text centered, rounded corners, minimum height=0.9cm},
    assessment/.style={rectangle, draw, fill=green!20, text width=\NodeW, text centered, rounded corners, minimum height=0.95cm},
    merge_graph/.style={rectangle, draw, fill=violet!20, text width=\NodeW, text centered, rounded corners, minimum height=0.95cm},
    visualization/.style={rectangle, draw, fill=yellow!30, text width=\NodeW, text centered, rounded corners, minimum height=0.85cm},
    final_output/.style={rectangle, draw, fill=orange!30, text width=\NodeW, text centered, rounded corners, minimum height=0.85cm}
  ]

  \node[task] (init) {\textbf{Initialize}\\ Seed piece $\mathcal{P}^{(0)}$, agent states $\{\mathcal{M}_i\}$};
  \node[generation, below=of init] (propose) {\textbf{Local Composition (per bar)}\\ \textsc{ExtractContext} $\rightarrow$ \textsc{GenerateBar}};
  \node[merge_graph, below=of propose] (assemble) {\textbf{Assemble Bars}\\ $\mathcal{P}^{(t)} \leftarrow \{b_i^{(t)}\}_{i=1}^N$};
  \node[assessment, below=of assemble] (critic) {\textbf{Global Critic}\\ Score $\sigma^{(t)}$, NL feedback $\psi^{(t)}$};
  \node[task, below=of critic] (update) {\textbf{Update Memories/Objectives}\\ $\{\mathcal{M}_i\}\!\leftarrow\!\textsc{Update}(\psi^{(t)})$};
  \node[visualization, below=of update] (persist) {\textbf{Save/Visualize}\\ JSON/MusicXML/WAV};
  \node[final_output, below=of persist] (finala) {\textbf{Composition} $\mathcal{P}_\star$};

  \path[line] (init) -- (propose);
  \path[line] (propose) -- (assemble);
  \path[line] (assemble) -- (critic);
  \path[line] (critic) -- (update);
  \path[line] (update) -- (persist);
  \path[line] (persist) -- (finala);
  \path[line] (update.west) to[out=180,in=180,looseness=1.05] (propose.west);

  \begin{pgfonlayer}{background}
    \node[draw=orange, thick, dashed, fit=(propose) (assemble) (critic) (update),
          inner sep=0.25cm, label={[text=orange]right:\textbf{Iteration}}] {};
  \end{pgfonlayer}
\end{tikzpicture}
\caption{Swarm with central critic.}
\label{subfig:central}
\end{subfigure}
\hfill
\begin{subfigure}[t]{\PanelW}
\centering
\scriptsize
\begin{tikzpicture}[
    node distance=0.45cm and 0.7cm, auto,
    line/.style={draw, -{Stealth}, shorten >=1pt},
    init/.style={rectangle, draw, fill=red!30, text width=\NodeW, text centered, rounded corners, minimum height=0.9cm},
    task/.style={rectangle, draw, fill=blue!20, text width=\NodeW, text centered, rounded corners, minimum height=0.9cm},
    generation/.style={rectangle, draw, fill=cyan!30, text width=\NodeW, text centered, rounded corners, minimum height=0.9cm},
    assessment/.style={rectangle, draw, fill=green!20, text width=\NodeW, text centered, rounded corners, minimum height=0.95cm},
    merge_graph/.style={rectangle, draw, fill=violet!20, text width=\NodeW, text centered, rounded corners, minimum height=0.95cm},
    visualization/.style={rectangle, draw, fill=yellow!30, text width=\NodeW, text centered, rounded corners, minimum height=0.85cm},
    final_output/.style={rectangle, draw, fill=orange!30, text width=\NodeW, text centered, rounded corners, minimum height=0.85cm}
  ]

  \node[init] (int) {\textbf{Initialize}\\ Agent states $\{\mathcal{M}_i\}$, traits $\mathbf{p}_i$};
  \node[task, below=of int] (env) {\textbf{Environment Signals}\\ Pheromones, energy, theme memory};
  \node[generation, below=of env] (sensecompose) {\textbf{Local Composition (per bar)}\\ \textsc{Sense} $\rightarrow$ \textsc{ComposeLocally}};
  \node[merge_graph, below=of sensecompose] (assembleb) {\textbf{Assemble Bars}\\ $\mathcal{P}^{(t)} \leftarrow \{\hat{b}_i^{(t)}\}$};
  \node[assessment, below=of assembleb] (peer) {\textbf{Peer Assessment \& Consensus}\\ Local NL feedback $\rightarrow$ \textsc{Aggregate}};
  \node[task, below=of peer] (evolve) {\textbf{Adapt}\\ Update environment \& agent personalities};
  \node[visualization, below=of evolve] (persistb) {\textbf{Save/Visualize}\\ JSON/MusicXML/WAV};
  \node[final_output, below=of persistb] (finalb) {\textbf{Composition} $\mathcal{P}_\star$};

  \path[line] (int) -- (env);
  \path[line] (env) -- (sensecompose);
  \path[line] (sensecompose) -- (assembleb);
  \path[line] (assembleb) -- (peer);
  \path[line] (peer) -- (evolve);
  \path[line] (evolve) -- (persistb);
  \path[line] (persistb) -- (finalb);
  \path[line] (evolve.west) to[out=180,in=180,looseness=1.05] (sensecompose.west);
  \path[line] (peer.east)  to[out=0,in=0,looseness=1.05] (env.east);

  \begin{pgfonlayer}{background}
    \node[draw=orange, thick, dashed, fit=(env) (sensecompose) (assembleb) (peer) (evolve),
          inner sep=0.25cm, label={[text=orange, anchor=south east, xshift=35pt, yshift=2pt]south east:\textbf{Iteration}}] {};
  \end{pgfonlayer}
\end{tikzpicture}
\caption{Decentralized swarm without central critic.}
\label{subfig:swarm}
\end{subfigure}
\hfill
\begin{subfigure}[t]{\PanelW}
\hspace{0.6em}
\centering
\scriptsize
\begin{tikzpicture}[
    node distance=0.45cm and 0.7cm, auto,
    line/.style={draw, -{Stealth}, shorten >=1pt},
    task/.style={rectangle, draw, fill=blue!20, text width=\NodeW, text centered, rounded corners, minimum height=0.9cm},
    generation/.style={rectangle, draw, fill=cyan!30, text width=\NodeW, text centered, rounded corners, minimum height=0.9cm},
    extraction/.style={rectangle, draw, fill=violet!40, text width=\NodeW, text centered, rounded corners, minimum height=0.9cm},
    visualization/.style={rectangle, draw, fill=yellow!30, text width=\NodeW, text centered, rounded corners, minimum height=0.85cm},
    final_output/.style={rectangle, draw, fill=orange!30, text width=\NodeW, text centered, rounded corners, minimum height=0.85cm}
  ]

  \node[task] (prompt) {\textbf{Single-Shot Prompt}};
  \node[generation, below=of prompt] (call) {\textbf{One-Pass Composition (all bars)}\\ No reflection or critique};
  \node[visualization, below=of call] (persistc) {\textbf{Save/Visualize}\\ JSON/MusicXML/WAV};
  \node[final_output, below=of persistc] (finalc) {\textbf{Final Composition} $\mathcal{P}$};

  \path[line] (prompt) -- (call);
  \path[line] (call) -- (persistc);
  \path[line] (persistc) -- (finalc);
\end{tikzpicture}
\caption{Single-shot.}
\label{subfig:single}
\end{subfigure}

\caption{Overview of three composition paradigms explored in this paper. (a) Swarm system with a central critic: a swarm of agents compose bars under a global critic that scores and provides NL feedback; agent memories/objectives update each iteration. (b) Decentralized swarm: agents sense shared environmental signals, compose locally, exchange peer feedback to reach consensus, and adapt personalities and signals—no central planner. (c) Single-shot: build a prompt, call once, generate, parse to a score; no iteration.}
\label{fig:alg_overview}
\end{figure}
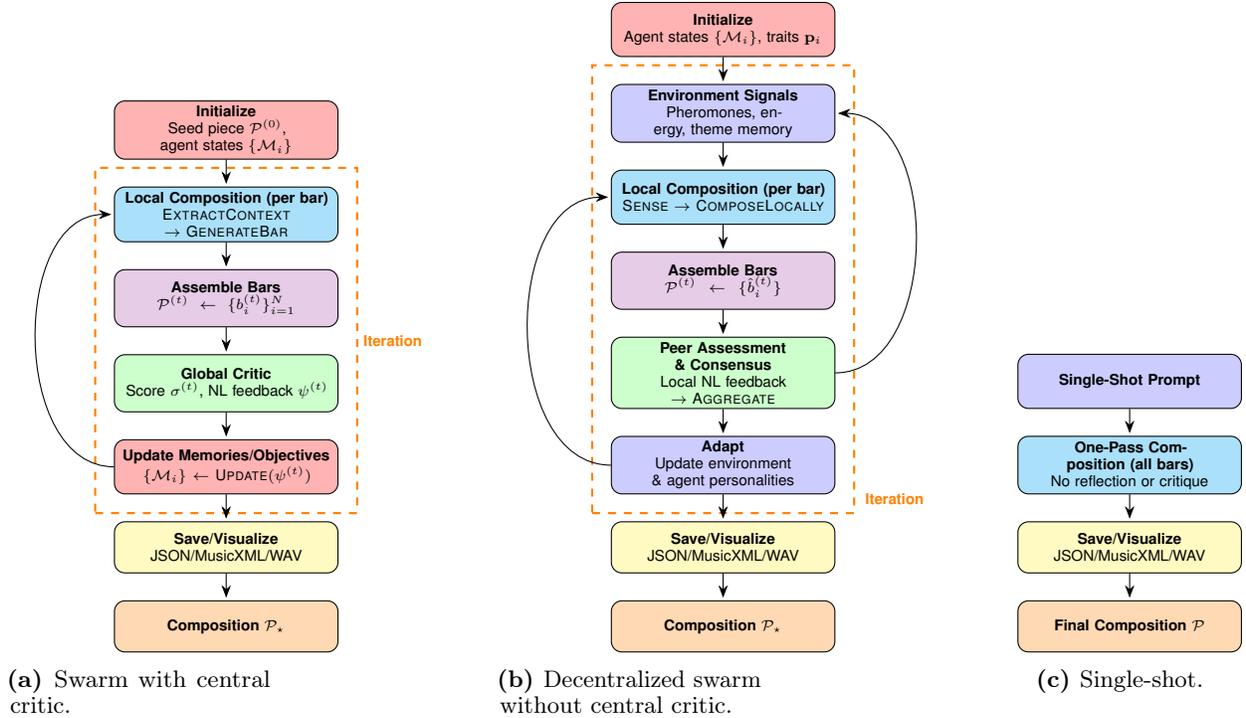

The plan of this paper is as follows. We introduce two swarm-based architectures for AI music composition that embody fundamentally different approaches to distributed creativity: a traditional multi-agent system that features a swarm of agents with centralized evaluation via a critic, and a biologically-inspired swarm intelligence system featuring peer-to-peer communication through pheromone-like signals \cite{kennedy2001swarm, bonabeau1999swarm}. Both systems eschew conventional fine-tuning approaches in favor of iterative learning through structured interaction, environmental feedback, and memory accumulation. By comparing these architectures against comprehensive computational musicology analysis \cite{cuthbert2010music21, lerdahl2001tonal}, we demonstrate how emergent musical intelligence can arise from distributed agent coordination, offering insights into the relationship between individual competency, collective behavior, and creative output quality in artificial intelligence systems. We compare these two highly agentic systems with a single-shot compositional approach. A sketch of the three primary architectures is shown in Figure~\ref{fig:alg_overview}. Additional details are provided in the Materials and Methods section. 

\section{Results and Discussion}

We present a series of experiments conducted with these three models introduced in Figure~\ref{fig:alg_overview}, along with deeper analyses of the results and comparative benchmarking. 

We use three fundamentally different computational paradigms that embody contrasting philosophies of  composition that will be described in the following sections. The first approach follows traditional multi-agent principles with centralized coordination, where a swarm of composition agents generate musical content under the guidance of a global critic system (Figure~\ref{fig:alg_overview}A). This hierarchical architecture employs episodic memory to accumulate compositional knowledge and uses directed feedback loops to refine agent behavior through supervised learning. The system operates through discrete phases: generation, centralized evaluation, memory consolidation, and selection of optimal solutions based on expert assessment. (Episodic memory is a context-rich record of agent experiences that enable credit assignment, prevent repetition of failed strategies, and support retrieval of successful patterns in later iterations. By contrast, regular memory would denote the transient prompt/context window.)

Our biological swarm intelligence system eliminates centralized control entirely, instead relying on emergent collective behavior arising from local agent interactions (Figure~\ref{fig:alg_overview}B). Individual agents possess randomized personality traits that create behavioral diversity, while communication occurs through chemical-inspired pheromone trails that encode musical patterns and preferences. Rather than hierarchical oversight, the system employs distributed peer assessment where agents evaluate each other's contributions through localized consensus formation. This decentralized architecture mirrors biological swarms—ant colonies, bee hives, flocking birds—where sophisticated group intelligence emerges from simple interaction rules without any central coordinator.

These contrasting implementations are compared with a single-shot compositional approach (Figure~\ref{fig:alg_overview}C) to benchmark against a more conventional strategy. Comparing the three approaches allows us to systematically investigate fundamental questions about the nature of creative collaboration: Does consistent musical  quality require centralized evaluation and expert guidance, or can sophisticated artistic decisions emerge from distributed consensus among autonomous agents? Can peer-to-peer communication and environmental feedback generate coherent musical narratives comparable to those produced by hierarchical systems? By comparing outcomes  from both paradigms across identical compositional tasks, we can empirically assess whether emergent self-organization or directed coordination yields superior musical creativity, coherence, and innovation in distributed artificial intelligence systems.

\subsection{Traditional Multi-Agent Composition System}

The traditional multi-agent architecture implements a hierarchical distributed intelligence framework where a swarm of specialized compositional agents operate under centralized coordination to achieve musical coherence through supervised learning and expert evaluation (Figure~\ref{fig:alg_overview}A). This system embodies the computational hypothesis that consistent artistic quality emerges from centralized evaluation and directed learning, leveraging established teacher-student paradigms proven effective in artificial intelligence  to create a compositional ecosystem capable of continuous refinement and knowledge accumulation.

The architecture's computational foundation rests on the integration of episodic memory systems with reinforcement learning mechanisms, creating what we term cumulative compositional intelligence. Each agent $A_i$ specializes in generating musical content $M_i(t)$ for temporal segment $t$ while maintaining global awareness through centralized feedback $F_c(M_{1:n})$ from a sophisticated critic system implemented using large language models. The critic provides multi-dimensional musical analysis encompassing harmonic progression evaluation, voice leading assessment, and structural coherence measurement—analytical capabilities that transcend individual agent perspectives and capture complex inter-voice relationships that emerge only at the ensemble level.

The system's learning dynamics operate through iterative policy refinement, where agent strategies $\pi_i$ evolve based on centralized evaluation scores $S_c$ and accumulated episodic experiences $E$. The episodic memory component $M_{ep}$ stores successful compositional patterns while maintaining negative examples to avoid  previously identified problematic structures, implementing a form of artificial musical experience analogous to human compositional development. This creates a
progressive learning trajectory where the collective system demonstrably improves musical sophistication across iterations, with measurable convergence toward expert-defined quality metrics.

Our hypothesis posits that centralized evaluation enables rapid convergence to high-quality musical solutions through consistent expert-level  assessment. Unlike distributed consensus mechanisms that may converge to mediocre local optima, the centralized critic provides stable quality gradients that guide agent exploration toward musically sophisticated territories. This pedagogical approach mirrors human compositional education, where students develop expertise through  structured teacher guidance, suggesting fundamental principles of creative skill acquisition that optimize both human and artificial learning systems.

 The computational uniqueness of this implementation transcends simple ensemble methods by creating genuine collaborative intelligence where agents adapt individual strategies $\pi_i \rightarrow \pi_i'$ based on collective performance metrics. This leads to emergent specialization phenomena where agents spontaneously develop complementary roles—melodic leadership, harmonic support, rhythmic anchoring—without explicit task assignment. The resulting musical outputs exhibit coherent artistic  vision while maintaining the exploratory diversity inherent in distributed generation, representing a synthesis of creative exploration with quality control mechanisms.

\subsection{Biological Swarm Intelligence System}

The swarm intelligence architecture (Figure~\ref{fig:alg_overview}B) represents a fundamental paradigm shift toward truly distributed creativity, where musical intelligence emerges spontaneously from  local agent interactions without centralized control, expert oversight, or predetermined quality metrics. This system tests the radical computational hypothesis that
sophisticated collective musical intelligence can arise from simple interaction rules and environmental communication, implementing principles observed in  biological swarms where complex group behaviors emerge from individual organisms following basic local protocols.

 The system's revolutionary computational framework centers on chemical-inspired communication through dynamic pheromone fields $\Phi(x,t)$ that encode musical patterns,  agent preferences, and emergent themes within a shared environmental space. Unlike traditional inter-agent message passing, this environmental communication creates a  persistent, evolving musical landscape that influences future compositional decisions through spatiotemporal pattern reinforcement. Agents deposit pheromones  $\phi_{i,j}(t)$ representing successful musical elements—melodic motifs, harmonic progressions, rhythmic patterns—with decay rates $\lambda_{decay}$ and reinforcement  strengths $\alpha_{reinforce}$ determined by peer validation, creating a natural selection mechanism for musical ideas that operate without external evaluation criteria.

  Each agent $A_i$ possesses personality vectors $\mathbf{p}_i = [r_i, h_i, \rho_i, \theta_i, s_i]$ representing risk-taking propensity, harmonic sensitivity,  rhythmic drive, thematic loyalty, and social influence susceptibility. These traits create behavioral diversity that prevents premature convergence while enabling personality-driven compositional exploration. The absence of global optimization objectives forces agents to develop consensus through distributed negotiation processes  $C_{consensus}(\mathbf{A}, \Phi)$, where musical coherence emerges from peer-to-peer assessment rather than expert evaluation.

  Our central hypothesis proposes that peer-driven consensus combined with environmental feedback can generate musical coherence comparable to expert evaluation  while discovering novel creative territories. The system's learning dynamics operate through personality evolution mechanisms where successful agents influence neighbors through pheromone strength modulation and direct social interaction. This creates adaptive swarm behavior $\mathbf{S}(t+1) = f(\mathbf{S}(t), \Phi(t),  \mathbf{I}_{peer}(t))$ where collective musical intelligence continuously evolves without external supervision, potentially accessing creative possibilities that  centralized evaluation might constrain through inherent biases toward established musical conventions.

  The computational significance of this approach extends beyond algorithmic novelty to address fundamental questions about creativity's emergent nature. By eliminating  human-designed evaluation criteria and implementing purely peer-driven consensus formation, we create experimental conditions for investigating whether artificial   systems can exhibit genuine creative emergence—the spontaneous generation of novel artistic patterns that transcend recombination of existing materials. The swarm's  distributed decision-making process $D_{swarm} = \bigcup_{i=1}^n d_i(\mathbf{p}_i, \Phi_{local}, \mathbf{N}_i)$ aggregates individual agent decisions based on local  personality, environmental sensing, and neighbor influence, creating collective intelligence that may discover unconventional musical relationships impossible under
  centralized oversight.

  The system's theoretical foundation rests on complexity science principles suggesting that creative intelligence emerges most naturally from decentralized networks with  sufficient diversity, interaction density, and selection pressure. Our implementation provides empirical testing grounds for these theories within the creative domain,
  offering insights into whether computational creativity requires human-imposed aesthetic criteria or can develop autonomous artistic sensibilities through distributed
  consensus formation and environmental adaptation.

\subsection{Single-Shot Approach}

The single-shot approach (Figure~\ref{fig:alg_overview}C) provides a baseline for comparison against both centralized and decentralized multi-agent frameworks. In this paradigm, a frozen foundation model is tasked with generating an entire $N$-bar composition in a single forward pass, without iterative refinement, memory, or feedback. The process is defined by a carefully constructed prompt that encodes the global objective. The model then produces a complete composition $P = \{b_1, b_2, \dots, b_N\}$, where each bar $b_i$ is represented as a JSON-structured collection of pitches and durations, immediately parsable into symbolic formats such as MusicXML or MIDI.

This architecture eliminates the reinforcement-style learning loop and bypasses the mechanisms of peer or critic evaluation that are central to the other two systems. As a result, it offers no opportunity for emergent specialization, role differentiation, or correction of local inconsistencies. Once the prompt is issued, the composition is fixed: there is no trajectory of adaptation or cumulative refinement across iterations. The absence of episodic memory $M_{ep}$ or environmental feedback $\Phi(x,t)$ ensures that the output reflects only the latent distribution of the pretrained model, conditioned on the single input.

\subsection{Experimental Results}

The first set of experiments is conducted based on this prompt: 
\begin{tcolorbox}[
  colback=gray!3,
  colframe=black!20,
  boxrule=0.6pt,
  left=6pt,right=6pt,top=6pt,bottom=6pt,
  title={Global objective \#1}
]
\small
Compose a coherent and emotionally expressive two-voice piano composition across 8 bars in A harmonic minor, with proper scale transitions. The upper voice should carry a lyrical melodic line, the lower voice provides rhythmic grounding and harmonic support. When musically appropriate, develop the motif A4-C5-E5-G\#5-A5. Play with expressive rhythms and ideas, but make sure the arc closes at the end.
\end{tcolorbox}

\begin{figure}[h!]
    \centering
    \includegraphics[width=.8\textwidth]{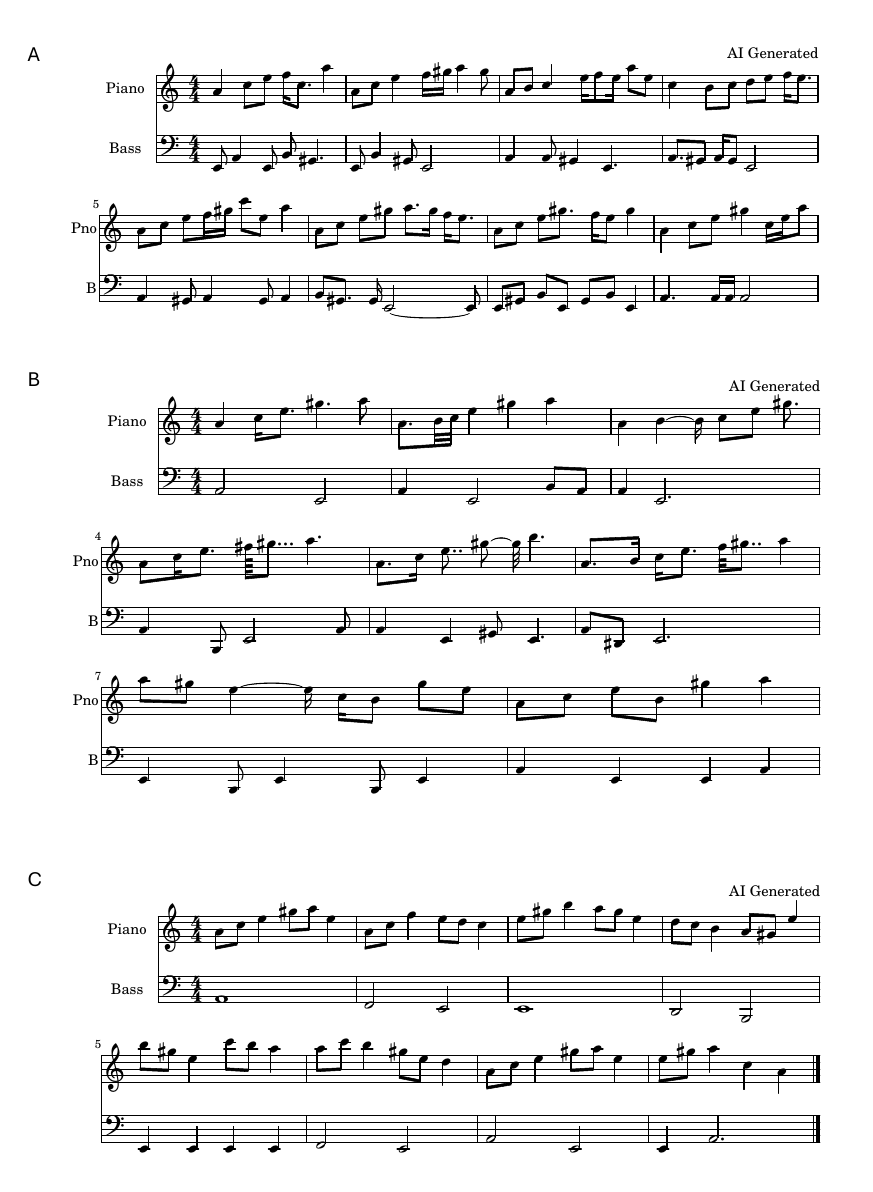}
    \caption{Three sample scores generated by the algorithms, for Global objective \#1. A, traditional multi-agent system. B, swarm system. C, single shot. In the multi-agent system (A), bar-wise agents coordinated by a global evaluator yield clear call-and-response between both parts, recurrent statements according to the prompt, and a more articulated cadence in the closing bars. In the swarm composition (B) we see longer melodic arcs, sustained tones, and sparser bass motion; motif fragments emerge implicitly without explicit top-down control. In the single-shot baseline (C) we see how a single pass produces locally plausible phrases but fewer motif returns, flatter phrase-level dynamics, and weaker long-range closure. In all cases, scores are shown on two staves (Piano treble; Bass).}
    \label{fig:score_example}
\end{figure}

Figure~\ref{fig:score_example} shows three sample scores generated using the three algorithms (for associated audio files, see Supplementary Information). 
To give a sense of the kind of objectives developed during the process, Text Boxes~S1-S3 show the raw output of evolving local per-agent objectives for the first bar, fifth bar, and final bar for the traditional multi-agent system. Text Box~1 shows the final objectives over all eight agents to give exact representations of the kind of musical reasoning conducted. 

\begin{tcolorbox}[
  float,                 
  floatplacement=tp,     
  colback=gray!2,
  colframe=black!20,
  boxrule=0.5pt,
  enhanced, 
  title={Text Box~1: Final local objectives of all eight agents, multi-agent system, for Global objective \#1.}
]
\begin{minted}[fontsize=\footnotesize, breaklines]{json}
{
  "Agent1": "Open with a smooth, lyrical statement of the A4–C5–E5 motif (with a gentle F passing neighbor) while the bass establishes a clear tonic foundation (sustained A2 with subtle rhythmic variation rather than large leaps) and moves purposefully toward G#2 at the bar’s close to prepare Bar 2’s dominant.",

  "Agent2": "Lyrically restate the A–C–E–G#–A motif with the F neighbor/appoggiatura and slight rhythmic displacement while the bass sustains a clear E2 dominant pedal with complementary syncopation and only tasteful passing tones (no isolated B2) to clarify harmony and set up Bar 3.",

  "Agent3": "Make Bar 3 a smooth, stepwise lyrical variation of the A–C–E–G#–A motif using F natural as a clear passing/neighbor tone, while the bass outlines A2→G#2→E2 with a lightly syncopated rhythm to clarify harmonic function toward the dominant (preparing Bar 4), ensure smooth voice-leading and avoid parallel octaves or abrupt register leaps.",

  "Agent4": "In Bar 4, present the inverted, stepwise variant of the A–C–E–G#–A motif in a slightly higher lyrical register with F as a passing/neighbor tone while clarifying bass function via a syncopated A2→G#2→E2 line that outlines iv→V preparation, introduces rhythmic contrast to break arpeggio repetition, and ensures smooth stepwise voice-leading into Bar 5 to better prepare the dominant arrival.",

  "Agent5": "In Bar 5, sequence/invert the A–C–E–G#–A motif in the right hand with increased spacing and an F-natural neighbor/pass tone to emphasize the harmonic-minor color, while the left hand provides a clearer preparatory bass line (A2 pulse moving toward E2, optionally via D2) using syncopation and tied rhythms to smooth the 0.75+0.25 placement, avoid parallel octaves/unisons, and shape voice-leading to firmly prepare the dominant for bars 6–7.",

  "Agent6": "Shape a lyrical climactic ascent of the motif by stepwise/arpeggiated motion around A5 (no notes above A5) with an F natural as a passing/neighbor tone, while making the bass more independent—using a B→G#→E approach with varied rhythmic placement and occasional ties—to clarify motion toward a dominant prep for bars 7–8 and avoid parallel octaves/unisons.",

  "Agent7": "Make Bar 7 a clear dominant preparation by anchoring the bass on E2 (varying its inversion/rhythm with E→G#→B to avoid static root-position repeats), sustain a dominant pedal with complementary syncopation, fragment/sequence the A–C–E–G# motif in the upper voice so G# is prominently voiced at the phrase-end, and add a short V–IV–V passing gesture that colors with F natural (no D in the bass) to increase forward momentum into bar 8.",

  "Agent8": "Deliver a decisive i cadence: place a grounded A2 on the downbeat (with a subtle anticipation or short rest for rhythmic interest), have the upper voice resolve G#→A stepwise and arpeggiate C5–E5 before a final A5 to confirm a clear A–C–E tonic sonority, use contrary motion to avoid parallel octaves, and briefly allude to the A–C–E–G#–A motif to close the arc."
}
\end{minted}
\end{tcolorbox}

\begin{figure}[h!]
    \centering
    \includegraphics[width=1\textwidth]{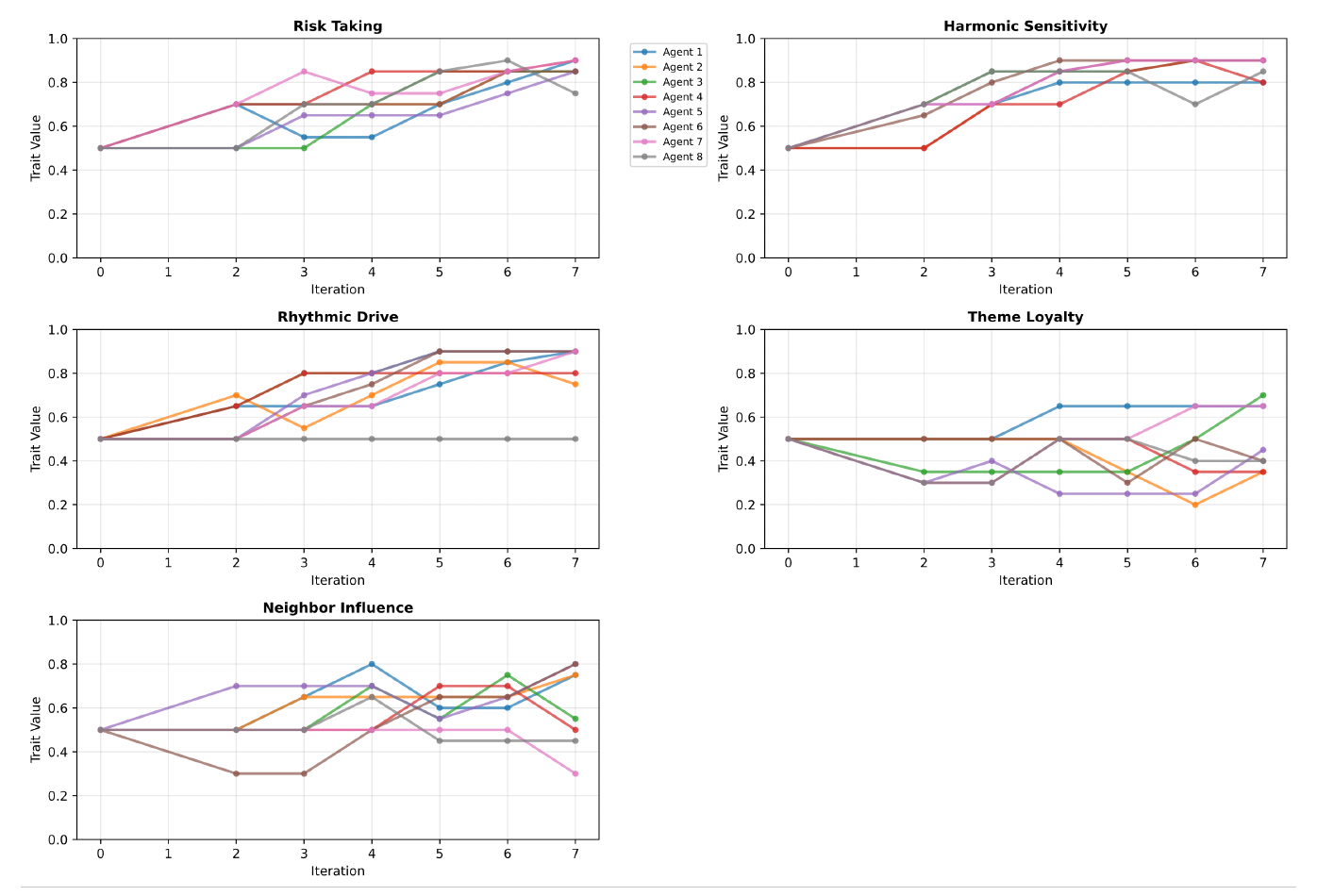}
    \caption{Trait evolution during swarm composition, for Global objective \#1. Each line is one of eight agents (legend) over global iterations. Risk Taking and Harmonic Sensitivity rise quasi-monotonically, approaching  0.85  as the ensemble explores then locks into the tonal scaffold. Rhythmic Drive ramps and saturates near 0.85–0.9, indicating tightening meter. Theme Loyalty is heterogeneous and non-monotonic, reflecting a trade-off between motif reuse and contrast as roles differentiate across agents. Neighbor Influence fluctuates rather than steadily increasing, showing periods of local coordination and autonomy with late-iteration partial convergence. The trajectories reveal progressive specialization and coordination emerging without a central planner; showing how trait trajectories reflect emergent role differentiation, not just convergence.}
    \label{fig:trait_eval}
\end{figure}

Figure~\ref{fig:trait_eval} illustrates how traits evolve over the course of swarm composition algorithm. Each curve corresponds to one of the eight agents, with values normalized to $[0,1]$ across global iterations (0–7). Several consistent patterns emerge. Risk Taking and Harmonic Sensitivity increase in a near-monotonic fashion, both approaching values around 0.85 as the ensemble first explores and then consolidates around a tonal scaffold. Rhythmic Drive ramps quickly and saturates between 0.85–0.9, signaling stabilization. In contrast, Theme Loyalty displays heterogeneous and non-monotonic behavior, reflecting the inherent tension between motif reuse and contrast as agents differentiate their roles. Finally, Neighbor Influence exhibits oscillations rather than steady growth, with alternating phases of local coordination and independence, and partial convergence emerging only in later iterations. Together, these trajectories demonstrate how progressive specialization and coordination can arise in a decentralized system without a central planner.

For a deeper analysis, Figure~\ref{fig:trait_stats} depicts the agent personality landscape at convergence. Several consistent trends emerge. Risk Taking, Harmonic Sensitivity, and Rhythmic Drive are not only high in magnitude but also tightly clustered across agents, suggesting that the swarm collectively stabilizes around a confident and harmonically grounded performance mode. These shared high values indicate that exploration is active yet bounded, and that rhythmic structure is firmly consolidated.
In contrast, Theme Loyalty is systematically lower and displays a broader distribution. This variability reflects a persistent tension between motif reuse and the need for contrast, pointing to an internal differentiation of roles: some agents reinforce thematic continuity while others deliberately introduce novelty. Such heterogeneity supports the ensemble’s ability to avoid stagnation while still maintaining coherence.
The greatest dispersion is observed in Neighbor Influence, which exhibits wide variance across agents. This indicates heterogeneous coupling strengths, with some agents responding strongly to their peers while others operate more autonomously. For instance, as highlighted in the per-agent heatmap (Figure~\ref{fig:trait_stats}B), Agent 8 shows reduced Rhythmic Drive but elevated Neighbor Influence, suggesting a compensatory role in sensing and aligning with the collective rather than driving its temporal structure.

These statistics reveal a swarm in which core musical competencies—exploration, harmonic modeling, rhythmic stability—are robustly shared, while higher-order coordination traits are differentiated across agents. This division of labor emerges without centralized control, underscoring the capacity of decentralized orchestration to generate both cohesion and diversity. Such findings reinforce the hypothesis that complex musical structures can arise through local adaptation and feedback loops, rather than from top-down planning, paralleling dynamics observed in natural collective systems.

\begin{figure}[h!]
    \centering
    \includegraphics[width=1\textwidth]{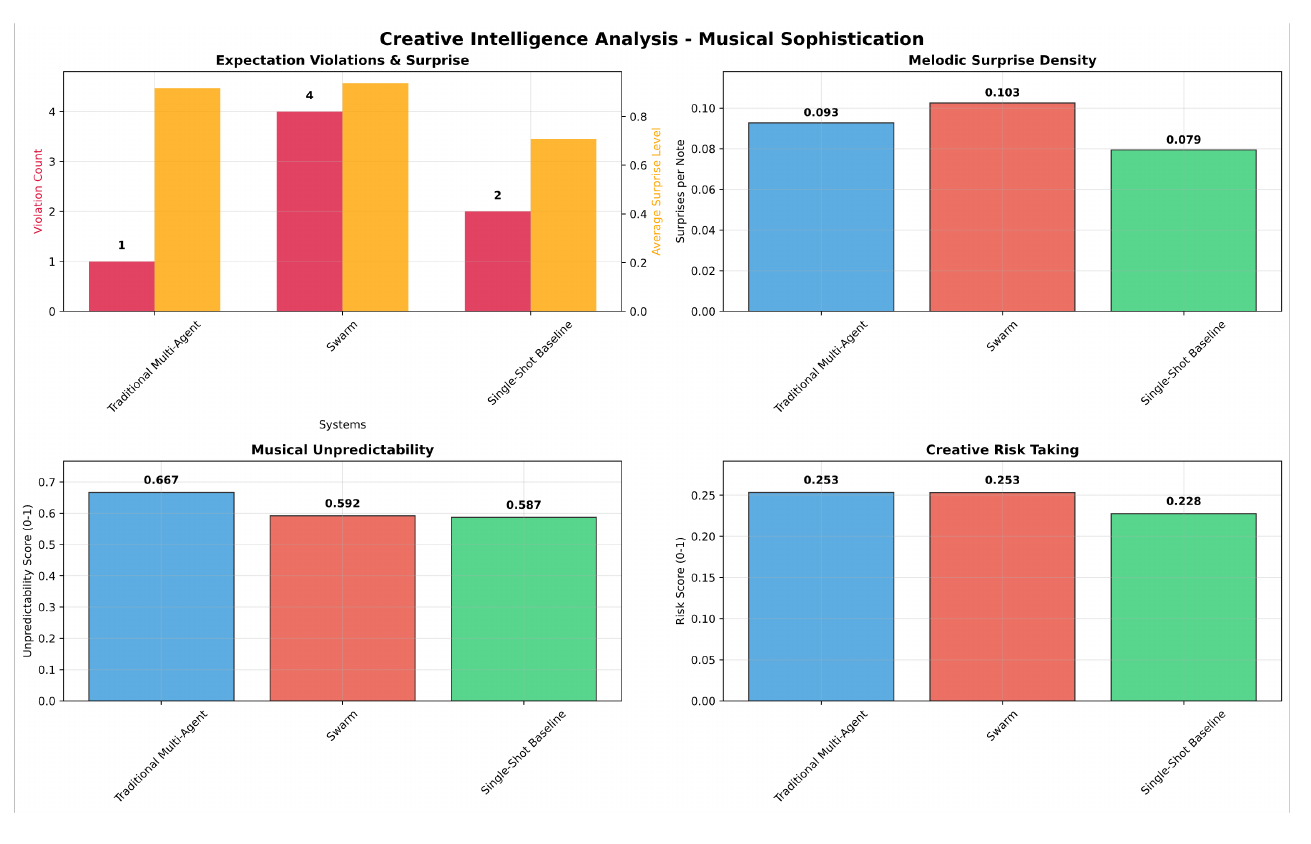}
    \caption{Creative-intelligence metrics comparing the three systems, for Global Objective \#1. Top-left: Expectation violations (left axis) and mean surprise per note (right axis) computed from a predictive pitch/rhythm model—swarm triggers the most violations (4) with high surprise, multi-agent shows 3, and single-shot 2. Top-right: Melodic surprise density (information content per note): swarm is highest (0.103), followed by multi-agent (0.093) and single-shot (0.079). Bottom-left: Musical unpredictability is greatest for the multi-agent system (0.667), then swarm (0.592) and single-shot (0.587). Bottom-right: Creative risk-taking (overall averaged measure of the three other qualities) is highest for multi-agent  =and swarm (0.253) and smallest for the single-shot case (0.228). The swarm yields more local surprises and rule-breaking events, while the multi-agent system pursues higher global unpredictability and risk; the single-shot baseline is the most conservative across all measures.}
    \label{fig:creative_surprise}
\end{figure}

We next compare the results across all three compositional models, as depicted in Figure~\ref{fig:creative_surprise}. Several distinct signatures emerge. The swarm model produces the highest number of expectation violations (4 on average) and elevated per-note surprise, indicating its tendency to probe and occasionally break the learned tonal and rhythmic scaffold. This is consistent with its decentralized orchestration, where local interactions generate bursts of novelty. The multi-agent model, by contrast, achieves fewer localized violations (3 on average) but exhibits the greatest global unpredictability and risk-taking index. This reflects a system that does not emphasize small-scale surprise events as much as broader shifts in trajectory and structure. The single-shot model is the most conservative across all metrics, with only two expectation violations, the lowest melodic surprise density, and the smallest composite risk-taking score.

These results suggest a trade-off between local and global creativity across model classes. The swarm architecture excels at generating micro-level deviations—subtle surprises that accumulate into rich textures—while the multi-agent system pushes toward macro-level unpredictability and structural risk. The single-shot baseline, lacking iterative feedback and adaptation, remains the most stable but also the least inventive, and follows relatively simple conventional compositional schemes. These findings highlight how decentralized orchestration can amplify local novelty without destabilizing the system, while coordinated multi-agent designs channel creative risk into global structural choices. Both approaches yield more adventurous outcomes than single-shot composition, underscoring the importance of interaction and feedback for emergent musical creativity.

\begin{figure}[h!]
    \centering
    \includegraphics[width=1\textwidth]{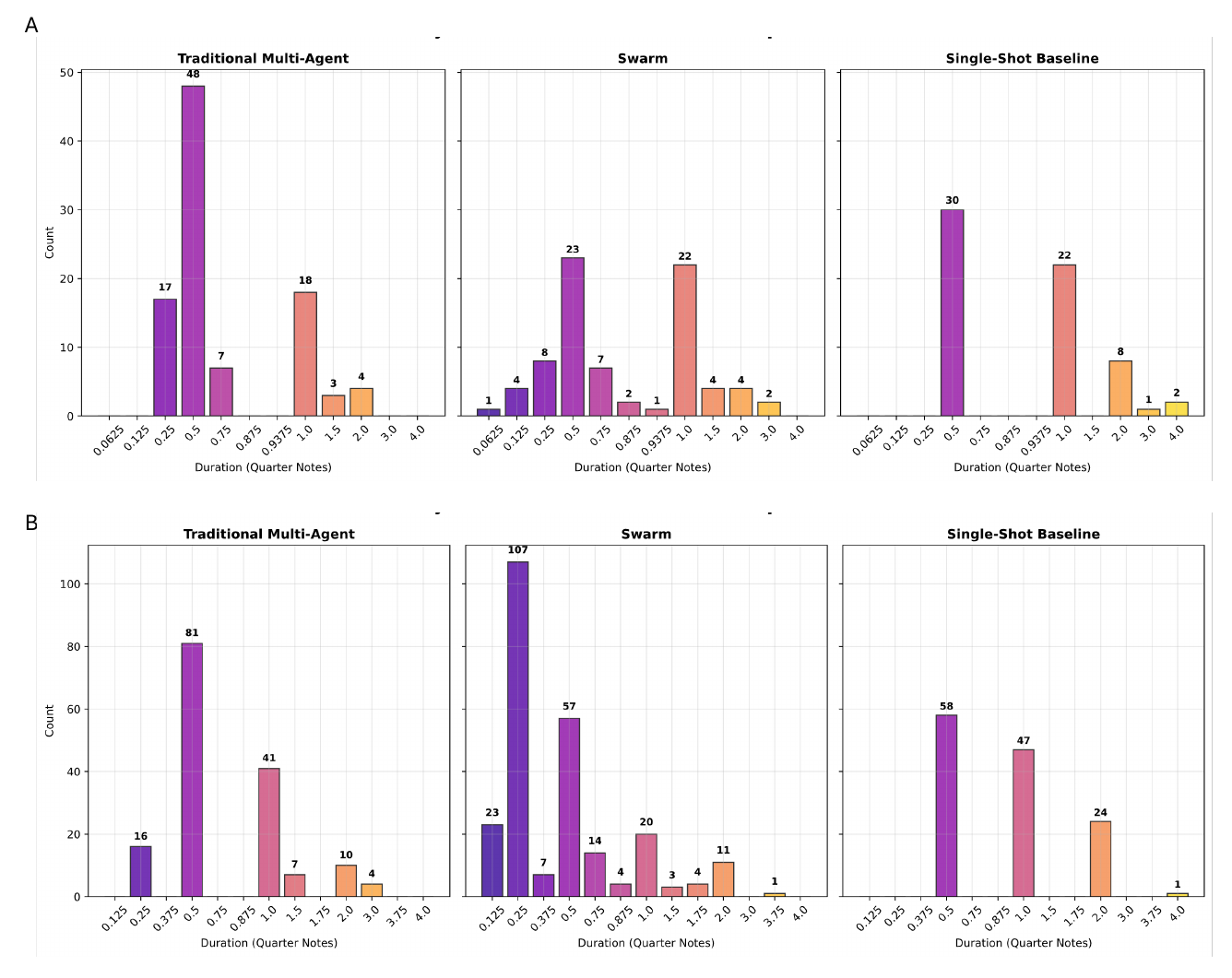}
    \caption{Rhythmic palette across systems, for Global Objective \#1 (A) and Global Objective \#2 (B). Histograms show the distribution of note durations (in quarter-note units) for Traditional Multi-Agent, Swarm, and Single-Shot Baseline. A: Multi-agent is strongly bimodal—eighths (0.5) and quarters (1.0) dominate (with few long holds) yielding a tight pulse. Swarm exhibits the broadest spread, including short values (around 0.25), dotted/subdivided notes (around 0.75–0.9375), and occasional longer sustains (1.5–3.0), indicating richer subdivision and syncopation. The single-shot baseline favors steady eighths/quarters but uses more long notes (1.5–4.0) and fewer fine subdivisions, producing a flatter rhythmic texture. Numeric labels above bars give counts per bin. B: Results for Global Objective \#2, showing similar general trends.}
    \label{fig:rhythmic_analysis}
\end{figure}

We next examine rhythmic structure across the three systems, as summarized in Figure~\ref{fig:rhythmic_analysis}. Clear differences emerge in the ``palette'' of note durations each model explores. The multi-agent system shows a strongly bimodal profile, with eighths (0.5) and quarters (1.0) dominating the distribution. The scarcity of longer notes produces a tightly pulsed texture, emphasizing rhythmic regularity over variation. By contrast, the swarm system exhibits the broadest and most heterogeneous spread of durations. It generates not only short values (around 0.25) but also dotted and subdivided figures (0.75–0.9375) alongside occasional extended sustains (1.5–3.0). This distribution points to a richer use of subdivision, syncopation, and rhythmic layering, consistent with the swarm’s capacity for emergent complexity. The single-shot baseline falls between these extremes: it favors eighths and quarters like the multi-agent model, but introduces more long notes (1.5–4.0) and fewer fine subdivisions. The result is a flatter, less articulated rhythmic texture that lacks both the swarm’s syncopated variety and the multi-agent model’s crisp regularity.

These findings resonate with the earlier trait analysis (Figure~\ref{fig:trait_eval}), where rhythmic drive consistently ramped and saturated near convergence. In the multi-agent case, this drive manifests as a strong pulse and disciplined metric stability; in the swarm case, it fuels a more exploratory search of rhythmic subdivisions and syncopated figures; and in the single-shot model, the absence of iterative feedback yields a more static deployment of long values that weakens articulation. Together, the two analyses demonstrate how rhythmic drive, quantified as an evolving trait, translates directly into the realized distribution of note durations—and how the orchestration mechanism (swarm, multi-agent, or single-shot) biases the way that drive is expressed.

\begin{figure}[h!]
    \centering
    \includegraphics[width=.8\textwidth]{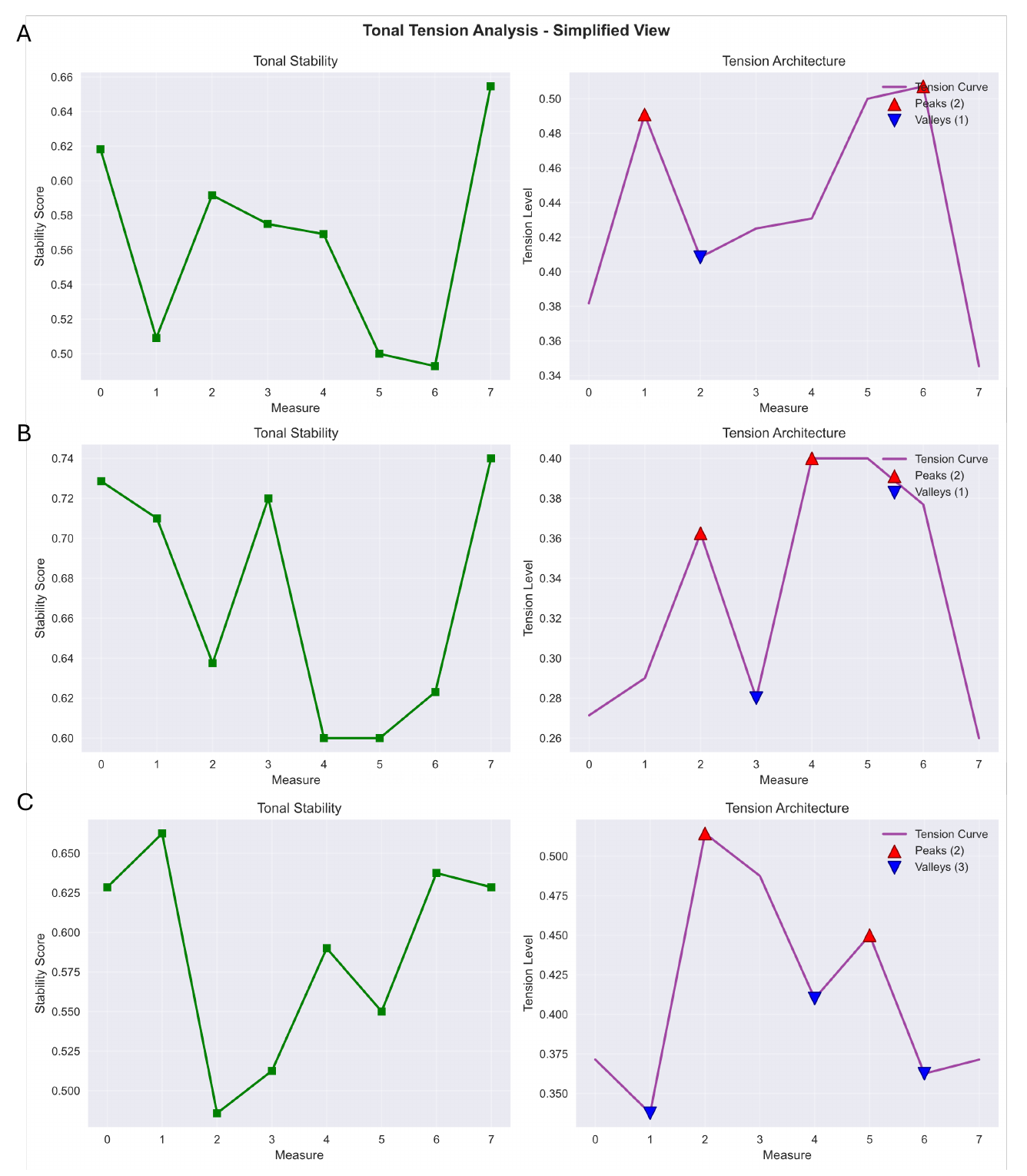}
    \caption{Tonal stability and tension architecture across measures for the three composition models for Global Objective \#1: agents with central critic (A), swarm system (B), and single-shot baseline (C). Left panels show tonal stability as a function of measure: higher values reflect closer alignment to tonal centers, while dips indicate excursions or chromatic deviation. The central-critic model (top) maintains moderate stability with fluctuations but culminates in a sharp rise at the cadence, suggesting guided resolution. The swarm (middle) exhibits overall higher stability punctuated by pronounced dips (e.g., measure 4–5), reflecting exploratory divergence followed by coordinated return. The single-shot baseline (bottom) demonstrates greater volatility, alternating peaks and troughs, consistent with its lack of iterative correction and more 
    fragile tonal grounding.}
    \label{fig:tonal_stability_analysis}
\end{figure}

Extending this view to harmonic structure, Figure~\ref{fig:tonal_stability_analysis} shows tonal stability and tension architecture across measures for the three compositional models: agents with a central critic (top), the swarm system (middle), and the single-shot baseline (bottom). The central-critic model maintains relatively stable tonal grounding with well-defined peaks and valleys, producing a conventional arch-like tension profile. The swarm, by contrast, exhibits smoother global organization punctuated by exploratory dips, suggesting decentralized coordination that integrates local divergences into coherent arcs. The single-shot baseline is the most volatile, with alternating peaks and troughs that reflect the absence of iterative correction or adaptive feedback. These patterns mirror the dynamics of the orchestration loop (Figure~\ref{fig:key_hypothesis}): swarm agents, instantiated from the same foundation model, adapt roles through feedback and policy updates, yielding emergent specialization that balances novelty and coherence. The central-critic system channels adaptation into more globally guided resolution, while the single-shot approach lacks the loop entirely, resulting in fragile tonal stability. Together, these results underscore how emergent specialization in decentralized swarms can generate expressive tension architectures without centralized control, while preserving coherence through iterative adaptation.
We can see that these results complement the creative-intelligence metrics (Figure~\ref{fig:creative_surprise}). The swarm’s exploratory dips and local fluctuations in tonal stability parallel its higher rates of expectation violations and melodic surprises, demonstrating how emergent specialization drives micro-level novelty. By comparison, the central-critic model mirrors the multi-agent system in emphasizing global guidance and producing broader, more predictable tension arcs—consistent with its higher risk-taking index and global unpredictability. The single-shot baseline, conservative across all surprise and risk metrics, shows the least coherent tension architecture, underscoring its inability to adaptively balance novelty and stability.

We proceed with a second example, this time composing a longer piece with 16 bars, and using \texttt{GPT-5} for the composing agents. The objective is:
\begin{tcolorbox}[
  colback=gray!3,
  colframe=black!20,
  boxrule=0.6pt,
  left=6pt,right=6pt,top=6pt,bottom=6pt,
  title={Global objective \#2}
]
\small
Compose a flowing and organic two-voice piano composition across 16 bars, inspired by biological growth and natural patterns. The upper voice should evoke the unfolding of a vine or the branching of neurons, with lyrical, evolving melodic lines that spiral outward. The lower voice should embody the grounding pulse of a heartbeat or the cycles of respiration, providing rhythmic stability and harmonic nourishment. Introduce motifs that resemble natural symmetries (e.g., ascending/descending stepwise motions, wave-like arpeggios, or cellular-like repetition with variation). Allow themes to propagate, mutate, and recombine as in evolution, but ensure the overall piece retains coherence. Close with a sense of resolution, as if the system has reached a natural equilibrium.
\end{tcolorbox}

\begin{figure}[h!]
    \centering
    \includegraphics[width=.8\textwidth]{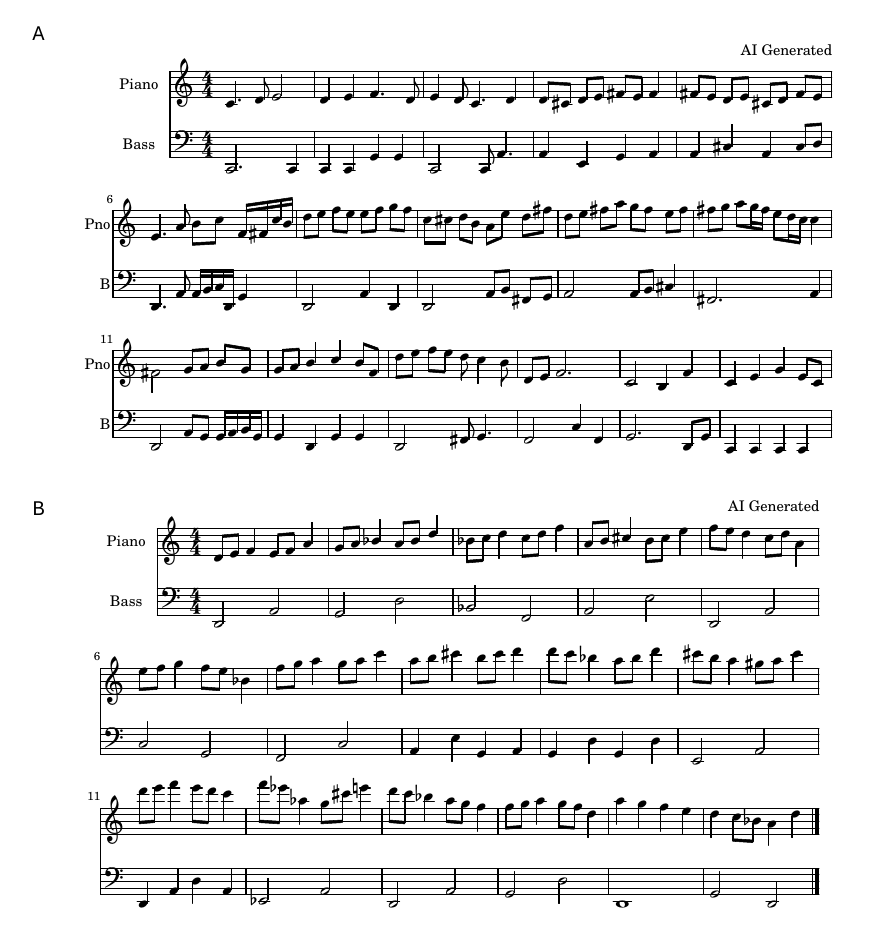}
    \caption{Results for the traditional multi-agent system (A), and the single-shot composition (B), for Global objective \#2. Note: Results for the swarm case are depicted in a separate figure due to the score's complexity.}
    \label{fig:score_example_102}
\end{figure}

\begin{figure}[h!]
    \centering
    \includegraphics[width=.67\textwidth]{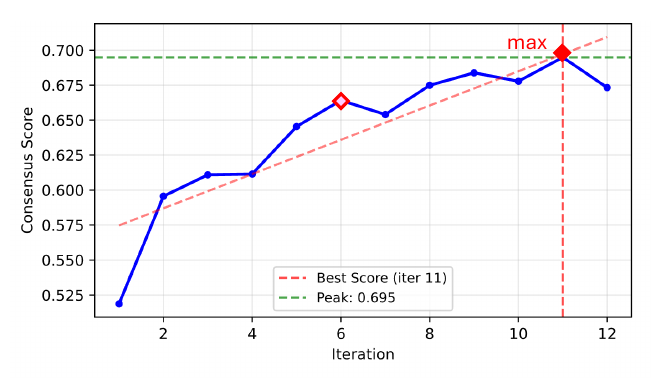}
    \caption{Score evolution for the swarm system under Global Objective~\#2. The blue trajectory shows the consensus score $\sigma(t)$ across 12 global iterations, reflecting how peer assessment and pheromone-mediated feedback progressively shape the ensemble’s output. The early iterations (1--3) display rapid gains as agents converge from random initializations toward a coherent scaffold, followed by smaller but steady improvements as local adaptation, personality evolution, and emergent theme reinforcement refine the composition. The lightly shaded red diamond denotes an intermediate local maximum at iteration 6, corresponding to a transient basin of stability where trait adjustments plateau before resuming exploratory adaptation. The red star at iteration 11 marks the overall maximum, accompanied by the dashed red vertical and horizontal lines indicating the iteration index and score level, respectively. The dashed red trend line illustrates the fitted improvement trajectory, while the dashed green line highlights the global peak consensus score attained by the system. The slight decline in iteration 12 suggests saturation and over-exploration, consistent with convergence toward a stable equilibrium point rather than indefinite improvement. This dynamic reflects the balance between exploration and exploitation in swarm-based coordination: early divergence builds diversity, while later convergence stabilizes complementary roles and fosters coherent musical structure.}

    \label{fig:score_example_200}
\end{figure}

We ran the objective with all three systems, but received the best result by far for the swarm model. Resulting scores are shown in Figure~\ref{fig:score_example_102} for the traditional multi-agent system and single-shot model. Results for the swarm model are far more complex, and are depicted in two separate figures. Figure~\ref{fig:score_example_100} shows the overall highest scoring result, and Figure~\ref{fig:score_example_101} shows the first local maximum. See Figure~\ref{fig:score_example_200} for the evolution of the score assessment for this case.  

It is also worth reflecting on the listening experience, as we found an intriguing dimension of how the swarm-generated composition unfolded for human listeners. On first encounter, the piece’s complexity may feel abstract or opaque. Yet, with repeated listening and especially with a simple rhythmic overlay (like a simple 808 drumbeat), the intricate rhythmic diversity and hidden structures became vividly clear. This process of initial unfamiliarity followed by deeper engagement underscores the emergent complexity of the composition. It highlights how the listener’s perception evolves—transforming an initially abstract piece into a richly intelligible and rhythmically engaging experience.

\begin{figure}[h!]
    \centering
    \includegraphics[width=1\textwidth]{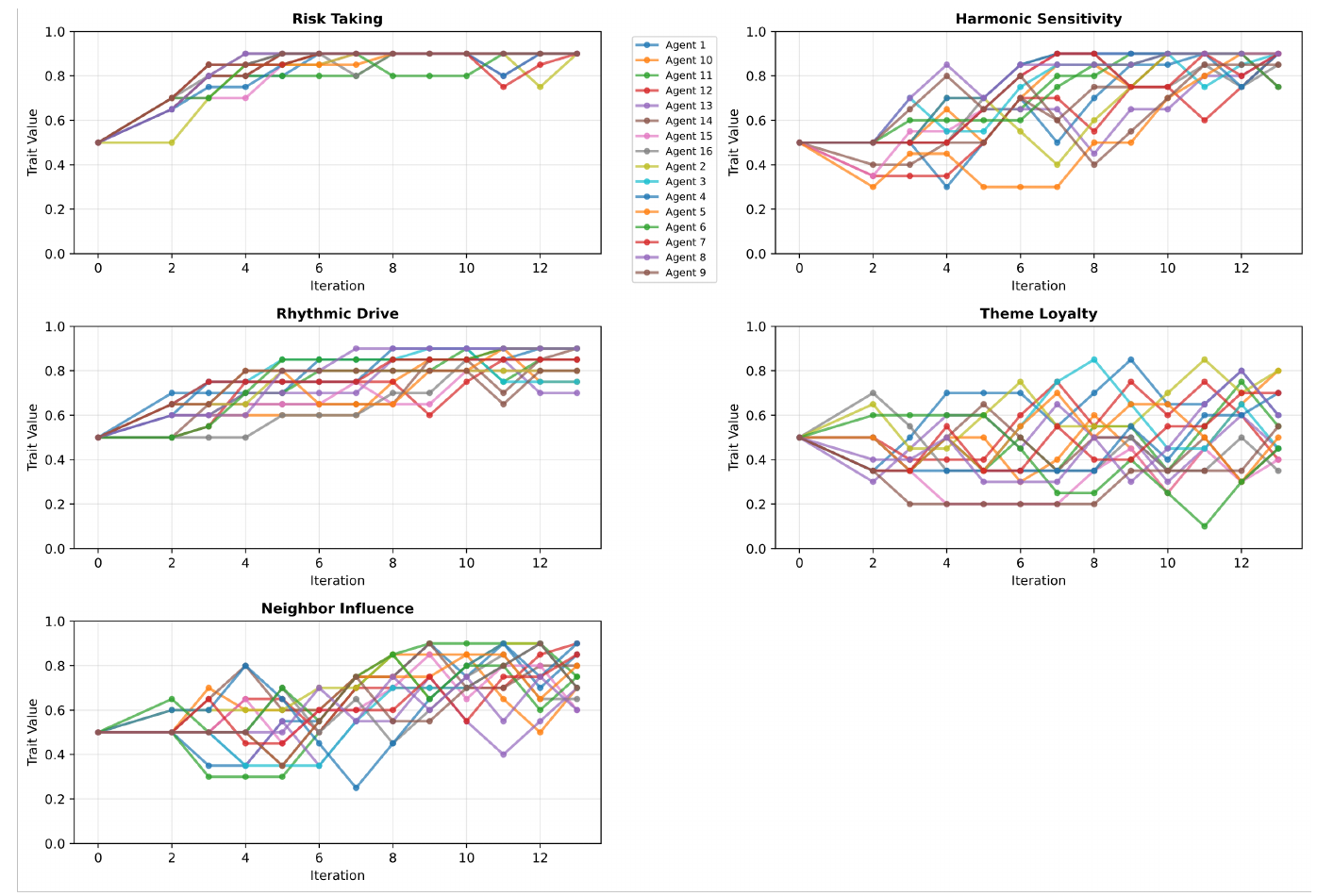}
    \caption{Trait evolution during swarm composition, for Global objective \#2. Each line is one of eight agents (legend), and values are normalized to 
$[0,1]$ over global iterations. Each line represents an individual agent adapting its behavioral parameters. Risk Taking shows rapid convergence toward high values, indicating agents increasingly embrace exploratory choices. Rhythmic Drive similarly stabilizes near saturation, reflecting strong alignment on rhythmic coherence. In contrast, Harmonic Sensitivity exhibits larger fluctuations, suggesting that agents continuously adjust their harmonic balance as the composition evolves. Theme Loyalty remains diverse across agents, with no clear convergence, highlighting the persistence of variation in motif development strategies. Finally, Neighbor Influence shows gradual increases with substantial variability, capturing the tension between individuality and local peer adaptation. These observations demonstrate the coexistence of convergence in global stylistic traits and sustained heterogeneity in local strategies, a feature often found in emergent swarm creativity.} 
    \label{fig:trait_eval_objective2}
\end{figure}

\begin{figure}[h!]
    \centering
    \includegraphics[width=.8\textwidth]{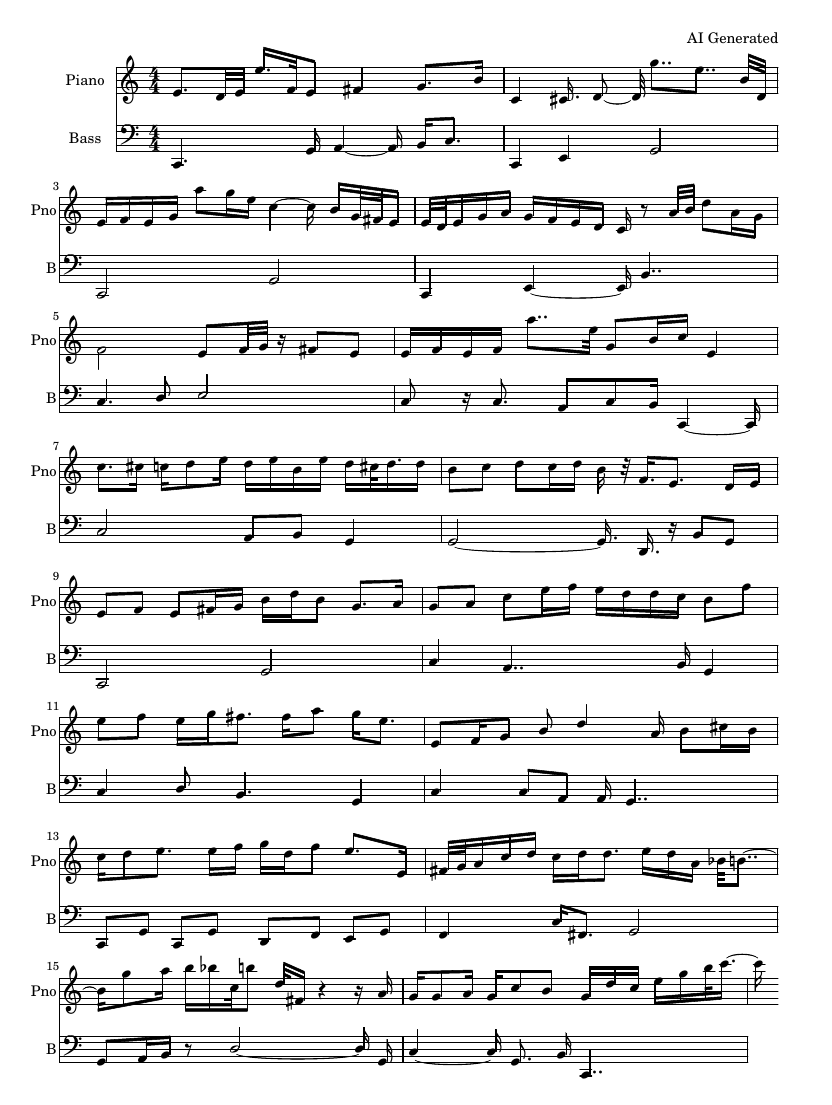}
    \caption{Resulting musical score at the best-scoring iteration, for the swarm case, for Global objective \#2.}
    \label{fig:score_example_100}
\end{figure}

Figure~\ref{fig:trait_eval_objective2} shows the evolution of five emergent traits across iterations in the multi-agent composition system, for Global objective \#2. Each subplot illustrates how individual agents adapt their behavioral parameters over time. Risk taking and rhythmic drive converge rapidly toward high values, indicating alignment on exploration and rhythmic coherence. Harmonic sensitivity displays sustained fluctuations, reflecting ongoing adaptation to harmonic context. Theme loyalty remains highly diverse across agents, highlighting persistent variation in motif development strategies. Neighbor influence gradually increases but with large variability, capturing the balance between individuality and local peer adaptation. These results demonstrate how convergence in global stylistic dimensions coexists with sustained heterogeneity in local strategies, a hallmark of emergent swarm creativity over iterations. Figure~\ref{fig:trait_stats} depicts a detailed visualization of the traits in the final iteration, akin to the earlier result. This is consistent with the earlier result of emergence of specialization without weight updates, forming a paradigm for specialization that is cost-efficient and domain-transferable.

Figure~\ref{fig:nash_analysis_delta} plots observed $\overline{\Delta}_t$ against $t{+}1$ (the ``next'' iteration). Therein, small values indicate that, on average, agents would \emph{barely change} their traits on the next round; sustained lows (plateaus) and local minima of
$\overline{\Delta}_t$ therefore mark low-motion basins or local rest points in the dynamics.
In our experiments, early dips (e.g., $t{+}1\approx 4\text{–}5$) show transient stabilization, while later,
deeper dips (e.g., $t{+}1\approx 9\text{–}11$) indicate convergence to a more stable configuration.
This is consistent with a behavioral fixed-point interpretation (no unilateral change at the profile),
though a payoff-based Nash certificate requires explicit utilities. We observe two performance maxima (around iteration 6 and in the second-to-last iteration, at 11). The first follows a brief stabilization in an early basin, while the second coincides with the later, deeper low-motion basin; this indicates the dynamics transition from a transient local optimum to a higher-quality equilibrium. Score peaks may be offset from motion minima due to exploration noise, evaluation lag, and the fact that the agent behavior is not solely determined by the numerical traits but additional natural language delineations. Nonetheless, the later peak aligns with the lowest sustained motion, consistent with convergence to a better local equilibrium. 

Figure~\ref{fig:nash_analysis_delta} also shows the calibrated overlay between the observed step–to–step mean absolute trait change $\overline{\Delta}_t$ and the learned best–response model is coherent with the dynamics inferred from the data. The fitted line–topology maps $x^{(t+1)}_{:,k}=M_k x^{(t)}_{:,k}+c_k$ are locally contractive (spectral radii $\rho(J_k)=\rho(M_k)<1$), so iterating the model drives the system toward a fixed point; empirically, $\overline{\Delta}_t$ decays into low–motion plateaus with a small residual floor. The affine calibration $(\lambda,\delta)$ reconciles the idealized, synchronous model with real execution (exploration, clipping, asynchrony): $\lambda$ rescales the effective step size, while $\delta$ captures persistent jitter. Two basins are evident (early $\sim 3$–$5$, later $\sim 9$–$11$), and the later, deeper basin coincides with the higher score peak, indicating stabilization at a better configuration. Together with small fixed–point residuals $\lvert f(x)-x\rvert$ at the terminal iteration, these results provide a behavioral $\varepsilon$–Nash certificate: holding others fixed, each agent’s learned best–response would change its traits only minimally, and local shocks decay. We do not claim a payoff–based Nash equilibrium here, since utilities are not modeled explicitly; establishing a utility $\tau$–NE would require per–agent scores and a unilateral–deviation test. Nonetheless, the convergence pattern, calibrated agreement, and local stability jointly substantiate the equilibrium interpretation of the final swarm state.
For additional details, see Supplementary Information.

A deeper dive into the details of the composition shows similar trends as identified earlier. When we examine rhythmic structure across the three systems, for Global Objective \#2 as shown in Figure~\ref{fig:rhythmic_analysis}, clear differences emerge in the ``palette'' of note durations.

\begin{figure}[h!]
    \centering
    \includegraphics[width=1\textwidth]{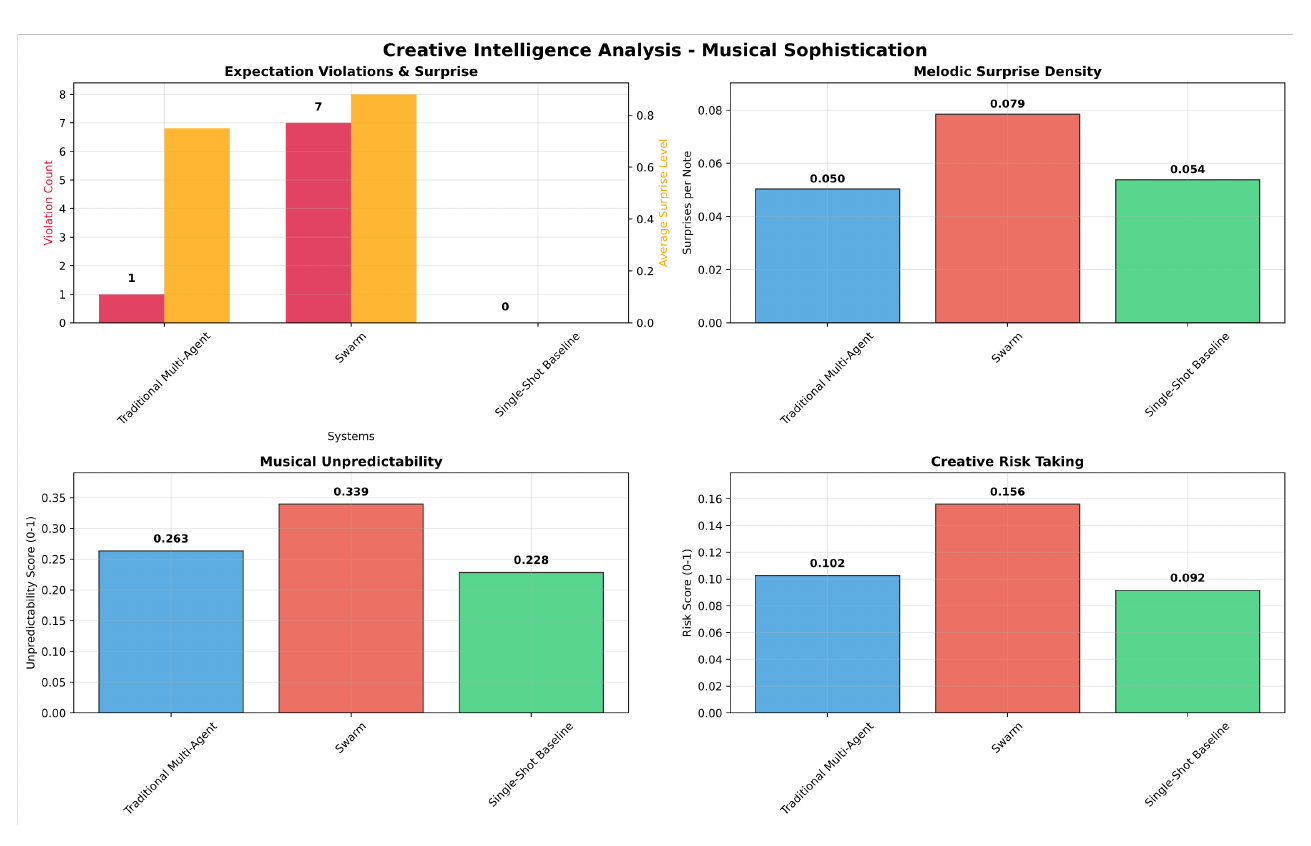}
    \caption{Creative-intelligence metrics comparing the three systems under Global Objective~\#2. 
    Top-left: Expectation violations and mean surprise per note. The swarm produces the highest number of violations (7) with elevated surprise levels, while the traditional multi-agent shows only 1 violation with modest surprise, and the single-shot baseline shows none. 
    Top-right: Melodic surprise density (information content per note), where the swarm again dominates compared to the single-shot baseline and multi-agent system. 
    Bottom-left: Musical unpredictability is greatest for the swarm , followed by multi-agent  and single-shot, indicating richer variation in musical trajectories. 
    Bottom-right: Creative risk-taking, a composite metric of the other measures, is also highest in the swarm, with multi-agent and single-shot  trailing last. 
    This data suggests that the swarm emphasizes local creativity through rule-breaking and high-information content, the multi-agent system provides moderate global unpredictability with fewer local disruptions, and the single-shot baseline remains the most conservative across all dimensions.}
    \label{fig:creative_surprise_objective2}
\end{figure}

Next, assessing two evaluations of creative-intelligence metrics (Figures~\ref{fig:creative_surprise} and \ref{fig:creative_surprise_objective2}) reveals a striking shift in how the swarm and multi-agent systems express creativity. In the earlier results (Figure~\ref{fig:creative_surprise}), the multi-agent framework slightly outperformed the swarm in global unpredictability and overall risk-taking, while the swarm led in expectation violations and melodic surprise density. This suggested a division of labor: the swarm as a driver of local novelty and rule-breaking, and the multi-agent system as a generator of broader-scale unpredictability and structural risk.  
By contrast, the analysis for Global Objective~\#2(Fig.~\ref{fig:creative_surprise_objective2}) shows the swarm dominating across all four metrics: generating the most expectation violations (7 vs. 1 in multi-agent, 0 in single-shot), the highest melodic surprise density (0.079), the greatest unpredictability (0.339), and the strongest overall risk-taking (0.156). Here, the multi-agent system is relegated to an intermediate role, contributing modest unpredictability (0.297) but little local surprise or risk. The single-shot baseline remains consistently conservative across both evaluations.  
This shows that the balance between swarm-driven local creativity and multi-agent global unpredictability is sensitive to evaluation method and metric definition. Results obtained for Global Objective~\#2 underscores the swarm’s ability to dominate both domains simultaneously, reinforcing its role as the most generative and exploratory system under this objective.

\begin{figure}[h!]
    \centering
    \includegraphics[width=1\textwidth]{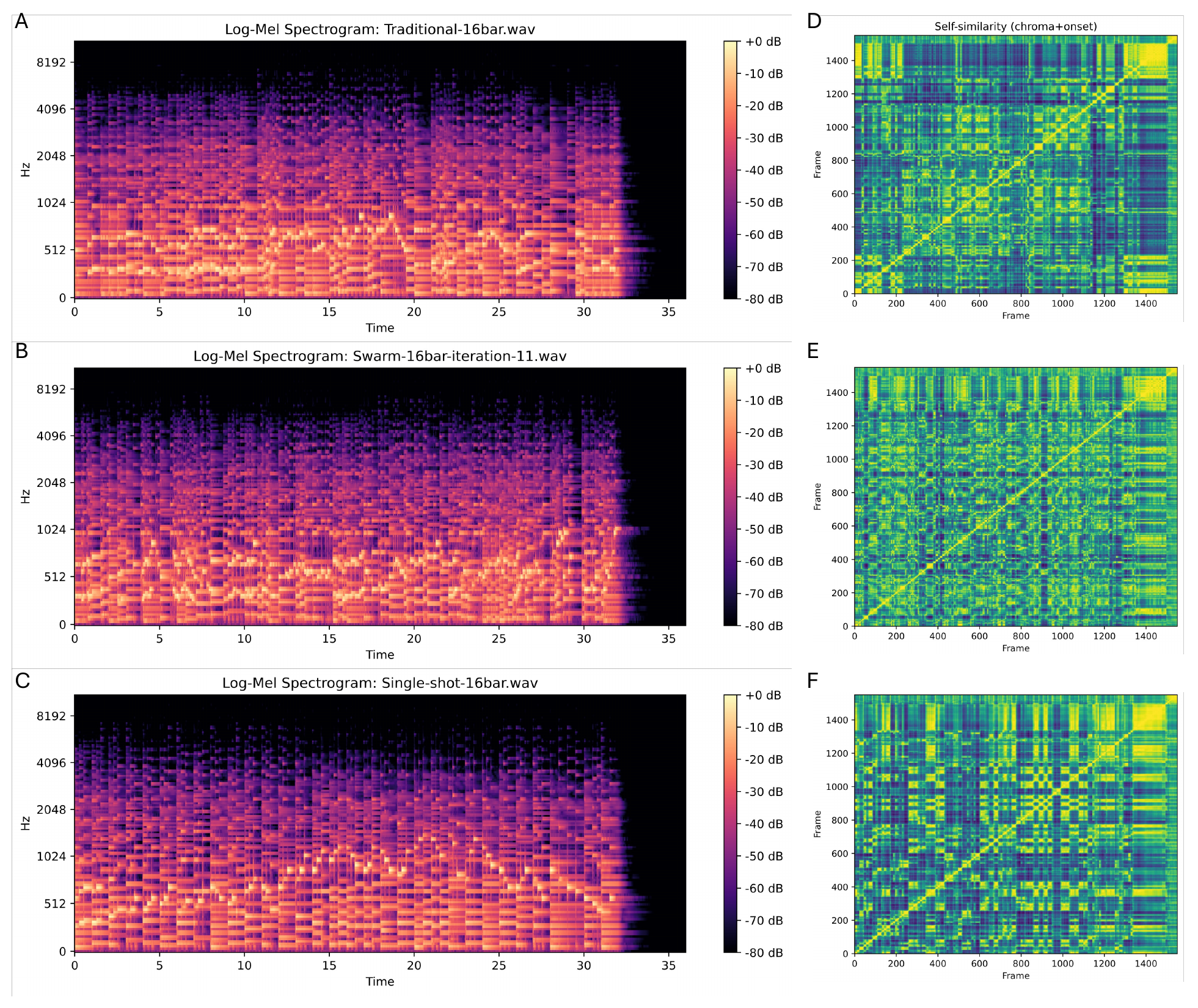}
    \caption{Detailed audio signal analysis using Log-Mel spectrograms and self-similarity matrices (SSMs). Log-Mel spectrograms are shown   
    of traditional (A), swarm (B), and single-shot (C) compositions. 
    The traditional output exhibits a diffuse, relatively uniform energy distribution across frequencies, reflecting its higher spectral entropy. 
    The swarm output shows alternating regions of concentrated and diffuse energy, producing a dynamic textural profile consistent with its elevated harmonic tension and frequent novelty events. 
    The single-shot output displays brighter spectra with clearer block-like segmentation, aligned with its higher spectral centroid and beat-synchronous novelty. 
    Together, these spectrograms illustrate that swarm compositions occupy a distinct timbral regime, balancing brightness, spread, and structural variability.
    Self-similarity matrices (SSMs) (chroma+onset) are shown for traditional (D), swarm (E), and single-shot (F) compositions. 
    Bright blocks along the diagonal represent regions of internal consistency, while off-diagonal streaks indicate recurring material across sections. 
    The traditional output shows clearer block segmentation with repeated sections, while the single-shot output exhibits highly regular, grid-like repetition. 
    In contrast, the swarm output displays a more irregular but richly interconnected pattern, suggesting emergent structural organization that is neither formulaic nor purely diffuse.
    }
    \label{fig:mel_spectrograms}
\end{figure}

\subsubsection*{Audio Analysis}

To enable audio-domain analysis, we rendered all symbolic scores into WAV files using Ableton Live using a consistent approach across all cases, ensuring consistent timbral quality across all systems (details see Materials and Methods). These audio files provided the basis for subsequent harmonic and timbral descriptor analyses.

As depicted in the spectrograms shown in Fig.~\ref{fig:mel_spectrograms}, the three approaches exhibit strikingly different textural signatures. The traditional composition shows a diffuse and relatively uniform spectral distribution, consistent with its higher entropy values. The swarm output instead alternates between concentrated and diffuse regions, producing a more dynamic spectral profile that mirrors its elevated harmonic tension and frequent novelty events. By contrast, the single-shot output displays brighter spectra with more pronounced block-like segmentation, aligning with its higher centroid and stronger beat-synchronous novelty. These contrasts reinforce that swarm compositions develop distinct timbral trajectories, balancing brightness, spread, and variability in ways not captured by either baseline.

The audio data is used to assess additional harmonic and timbral descriptors (Figures~\ref{fig:audio_creative_tension}). The harmonic analysis highlights that swarm outputs sustain the highest mean tension, reflecting a greater degree of harmonic instability and dissonant colorations, while the single-shot baseline remains most consonant and metrically predictable. The swarm pieces also exhibit the highest density of local Jensen–Shanno(JS)-novelty events, indicating more frequent micro-level harmonic recontextualizations. In contrast, single-shot outputs concentrate novelty at metrically aligned positions, suggesting a block-like structure with more conventional alignment to barlines. This pattern implies that swarm agents generate creative leaps that are not strictly bound to meter, a feature that distinguishes their emergent harmonic dynamics from baseline methods. Complementary timbral analyses reveal a consistent ``middle ground'' for swarm outputs: spectral bandwidths are broader than traditional pieces but narrower than single-shot, spectral entropy is higher than the single-shot but lower than traditional, and spectral centroids indicate brightness levels that lie between the two extremes. Taken together, these descriptors suggest that swarm compositions balance harmonic tension, novelty, and timbral richness in a way that is neither overly constrained nor diffuse, producing outputs that are both dynamically unstable and perceptually coherent—an emergent signature of distributed agentic creativity.
For more depth, PCA of global descriptors (Fig.~\ref{fig:pca_audio}) shows that swarm outputs form a distinct cluster, separating from both single-shot and traditional baselines, consistent with their intermediate timbral brightness and elevated harmonic tension.
Dynamic Time Warping (DTW) is a sequence alignment method that computes an optimal match between two temporal trajectories by allowing non-linear stretching and compression along the time axis, enabling comparison of patterns that may be similar in shape but misaligned in time. As shown in Fig.~\ref{fig:dtw_entropy}, such a DTW analysis of spectral entropy trajectories reveals that single-shot and traditional compositions follow closely aligned arcs of timbral complexity, while the swarm composition diverges markedly from both. This indicates that swarm outputs do not simply occupy an intermediate entropy level, but instead trace a qualitatively different time-course of ordered versus noisy spectral states, consistent with emergent dynamics.

To probe the higher-level organization of the generated pieces, we constructed self-similarity matrices (SSMs), a common tool in music information retrieval. An SSM encodes how similar each moment in a piece is to every other, creating a two-dimensional map of repetition, contrast, and structural development over time. This representation allows us to visualize large-scale form, identify recurring motifs, and compare the balance of novelty versus coherence across different generative paradigms.
As shown in Fig.~\ref{fig:mel_spectrograms}D-F, the SSMs reveal contrasting modes of structural organization. 
Therein, colors encode pairwise similarity between time frames, with brighter (yellow) regions indicating higher similarity in harmonic and onset features and darker (blue) regions indicating contrast. Off-diagonal bright patches mark recurring material across different sections, while darker regions reflect novel or contrasting passages.
The single-shot output exhibits grid-like repetition, while the traditional composition displays clear sectional blocks. The swarm output, however, produces a more irregular but interlinked pattern of similarities, suggesting emergent structure that avoids both rigid repetition and uniform diffusion.

We see that the combined analyses of SSMs and novelty curves reinforce the distinction between swarm and baseline outputs. As shown in Figure~\ref{fig:mel_spectrograms}, the single-shot composition exhibits highly regular, grid-like blocks, while the traditional piece presents clearer sectional segments with some repeated material. By contrast, the swarm output reveals a more irregular but densely interconnected pattern, with cross-similarities linking non-adjacent regions. This visual evidence aligns with the novelty analyses (Figure~\ref{fig:audio_creative_tension}), where swarm compositions show the highest density of framewise novelty peaks yet lower beat-synchronous alignment, suggesting that swarm agents generate frequent micro-level recontextualizations without adhering to rigid metric segmentation. Together, these analyses highlight that swarm outputs achieve structural coherence through distributed, emergent organization, distinct from both the block-like predictability of single-shot and the repetitive sectional form of traditional outputs. 
Notably, we find that the swarm self-similarity matrix exhibits more irregular, ``organic'' streaks and patchy structures, indicating fluid recombination of motifs across non-adjacent regions. This contrasts with the rigid, block-like diagonals of single-shot and the larger sectional blocks of traditional outputs, underscoring swarm’s emergent rather than formulaic organization.

\subsubsection*{Graph Analysis}

While the self-similarity matrices reveal how patterns of repetition and contrast manifest visually, they can also be interpreted formally as networks, where each time slice is represented as a node and edges encode similarity relationships. Casting the SSM into graph form enables the application of network science measures that quantify global organization, local clustering, and community structure. This perspective allows us to move beyond qualitative inspection of block-like or organic motifs and rigorously compare how swarm, single-shot, and traditional compositions differ in their connectivity patterns, and provide a deeper analysis of the organic patterns seen in the swarm case. In this way, emergent coherence in swarm compositions can be quantitatively captured through small-world properties, which reflect the balance between local clustering of motifs and global efficiency of connectivity across the piece.

\begin{figure}[h!]
    \centering
    \includegraphics[width=0.8\textwidth]{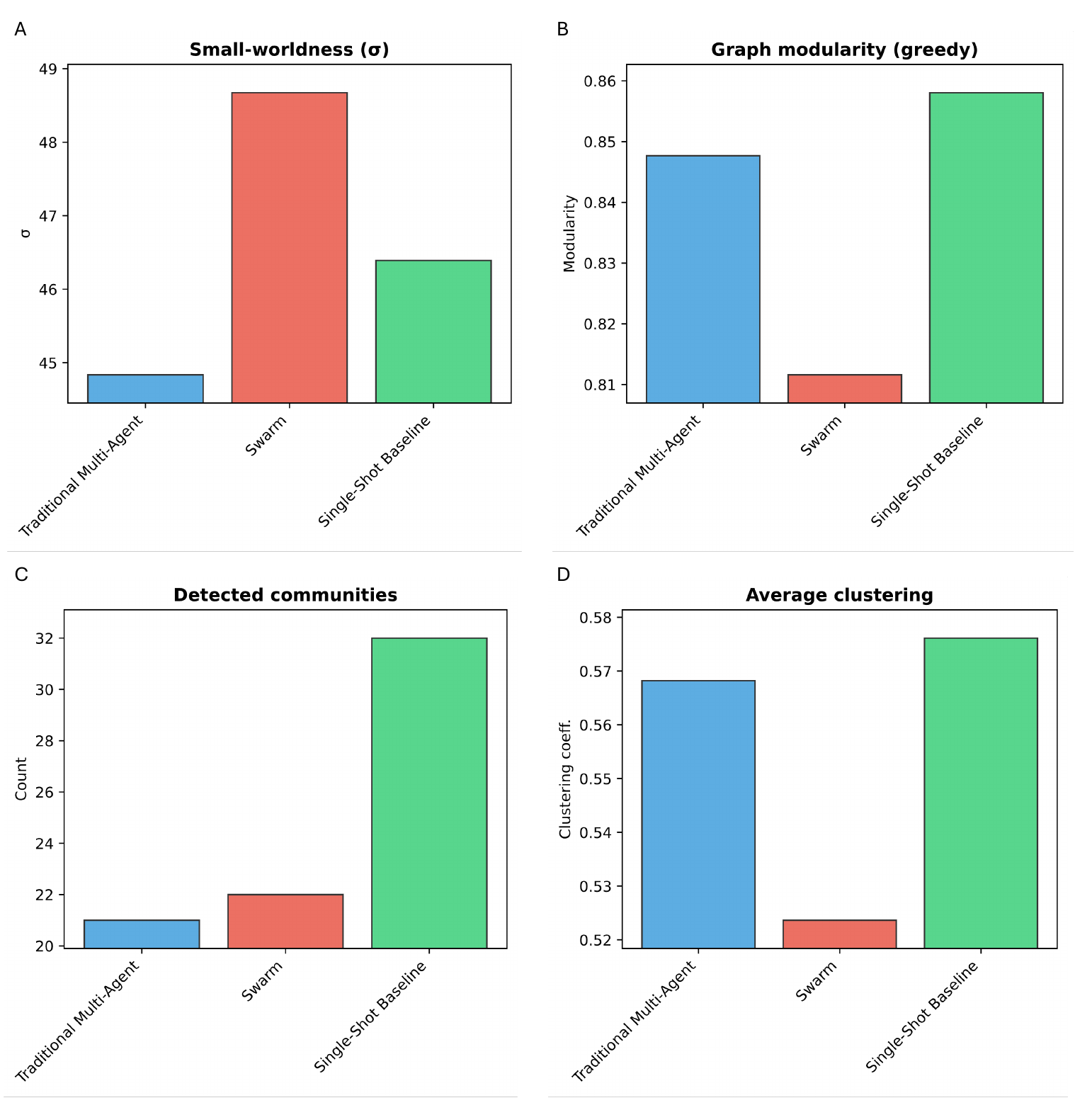}
    \caption{
    Graph-theoretic properties of self-similarity networks constructed from swarm, single-shot, and traditional compositions (see Table~\ref{table:audio_measures} for definitions of metrics using simple terms).  
    A: Small-worldness index $\sigma$ shows that all networks are highly small-world, with the swarm achieving the highest value, reflecting efficient global connectivity combined with local clustering. 
    B: Graph modularity reveals that single-shot and traditional graphs are more strongly partitioned into distinct communities, while the swarm network is less modular and more integrated. 
    C: Detected communities confirm this difference, with single-shot forming the largest number of separate clusters and swarm forming fewer. 
    D: Average clustering coefficient indicates that all graphs are locally cohesive, with single-shot highest, traditional slightly lower, and swarm reduced but still well above random baselines. 
    Together, these measures highlight that swarm compositions generate networks that are globally efficient yet less fragmented, contrasting with the more block-like structure of single-shot and traditional outputs.
    }
    \label{fig:graph_metrics}
\end{figure}

The graph-theoretic analyses highlight clear differences in how self-similarity networks organize under the three compositional approaches. As shown in Fig.~\ref{fig:graph_metrics}, all networks are strongly small-world, but with distinct emphases. The swarm graph achieved the highest small-worldness index ($\sigma \approx 48.7$) due to its combination of high clustering and the shortest path lengths, indicating efficient connectivity across the piece. In contrast, single-shot and traditional graphs were less efficient ($\sigma \approx 46.4$ and $44.8$, respectively), with longer path lengths despite comparable clustering levels. Modularity and community counts diverged: single-shot produced the most fragmented structure (32 detected communities, modularity 0.86), traditional slightly less (21 communities, modularity 0.85), while the swarm network was less modular (22 communities, modularity 0.81), reflecting a more integrated organization. 
We also note that consistent with prior work showing that human musical compositions exhibit small-world and scale-free network properties with coherent community structure~\cite{Liu2010MusicNetworks,Ferretti_2017}, our swarm-generated networks align more closely with these human-like traits than traditional or single-shot baselines. Notably, human‑like small‑world music networks arise from pheromone‑mediated consensus, not from a central critic.

These relationships are summarized in the small-world proxy plot (Fig.~\ref{fig:graph_smallworld_proxy}), where swarm occupies the upper-left quadrant, combining relatively high clustering with the greatest efficiency (1/ASPL) (where ASPL=Average Shortest Path Length, a network measure defined as the mean of the shortest path distances between all pairs of nodes in a graph; it quantifies the typical number of steps required to connect any two events, with lower ASPL indicating more efficient global connectivity). This balance suggests that swarm-generated music achieves a favorable trade-off between local cohesiveness and global reach, unlike single-shot, which favors strong modular clustering at the expense of longer global paths, and traditional, which yields more heterogeneous but less efficient structures.

\begin{figure}[ht]
  \centering
  \includegraphics[width=\textwidth]{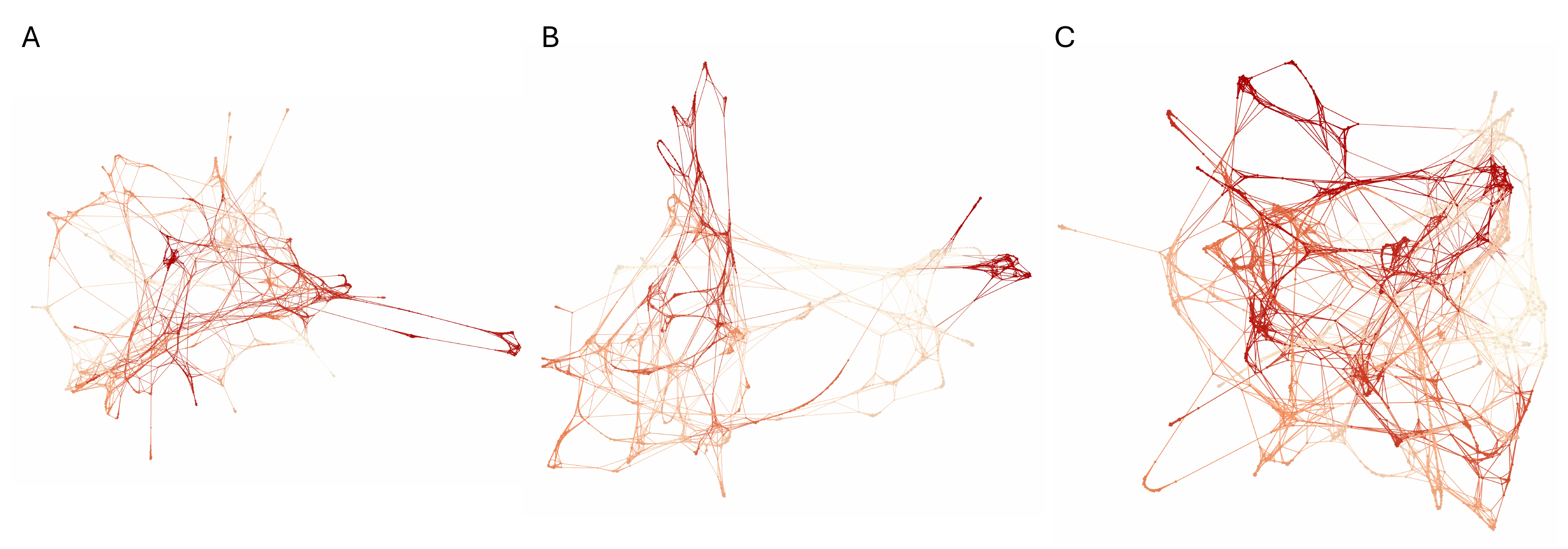}
  \caption{Graph representations of musical self-similarity for three generation paradigms:
  (A) traditional multi-agent baseline, (B) swarm (decentralized, pheromone-mediated),
  and (C) single-shot baseline. Node color denotes modularity class (community) detected
  by modularity optimization; edges encode pairwise similarity. All panels use identical
  layout and rendering parameters. The swarm condition exhibits multiple mid-sized communities with braided
  inter-community corridors and short-path redundancy, indicating stronger meso-scale integration and
  decentralized exchange than the traditional or single-shot baselines.}
  \label{fig:graph_communities}
\end{figure}

The qualitative network layouts reinforce these differences (Figure~\ref{fig:graph_communities}). The swarm graph is characterized by a smaller number of large, interlinked communities, producing a woven, integrated structure consistent with high efficiency and global coherence. By contrast, the single-shot graph shows the highest degree of fragmentation, with many smaller clusters corresponding to block-like motifs repeated in isolation. The traditional graph lies between these extremes, with numerous clusters connected by elongated branches, consistent with its higher degree entropy and heterogeneous local connectivity. Taken together, these graph analyses show that swarm compositions generate networks that are globally efficient, locally cohesive, and less partitioned, supporting the interpretation of emergent structural coherence distinct from both single-shot and traditional baselines.
The visual differences among the three network graphs seen in Figure~\ref{fig:graph_communities} align closely with the quantitative metrics. In the single-shot case, the network stretches into an elongated structure with many small, brightly colored clusters and weak cross-links. This matches its high modularity (0.86) and large number of communities (32), indicating fragmentation into tightly knit but poorly connected modules, further reflected in the longer average path length. The swarm graph, by contrast, appears more compact and rounded, with larger modules and braided seams where colors intermingle. These features echo its lower modularity (0.81, 22 communities), shorter path length, and highest small-worldness ($\sigma$= 48.7), pointing to stronger meso-scale integration and recurrent motif exchange. The traditional case falls in between: visually more cohesive than single-shot but less braided than swarm, consistent with its intermediate clustering and modularity values and its relatively long paths (around 9.5). This may imply that the swarm achieves a coherence–variety optimum, where we see music that is locally cohesive, globally efficient, and cross‑linked. This could serve as a powerful target criterion for future objective shaping in other contexts, such as scientific discovery.

Visual inspection of self-similarity graphs depicted in circular form reveals clear contrasts between systems (Fig.~\ref{fig:ssm_circle}) (detailed analysis and methods see Section~\ref{SI:deeper_structure}). In these circular layouts, nodes represent musical events arranged counterclockwise in time, with edges denoting high self-similarity and node size encoding betweenness centrality. The traditional multi-agent baseline produces fragmented long-range structure with several bottlenecks. The swarm output, in contrast, exhibits distributed long-range links without dominant hubs, reflecting balanced global coherence and shared structural roles. The single-shot baseline concentrates flow through a few large hubs, producing strong but motif-repetitive links. These qualitative differences suggest that swarm coordination enables broader coherence while maintaining structural variety.

To quantify these observations, we evaluated six graph-theoretic metrics of long-range coherence and variety (Fig.~\ref{fig:longrange_variety}). The swarm consistently achieved higher long-range edge fraction and long-range efficiency (panels A–B), indicating that distant sections are both more frequently and more efficiently bridged. Its communities spanned broader temporal arcs (panel C), showing greater persistence of thematic material. On measures of variety, swarm outputs displayed the highest edge-length entropy (panel D), mixing recurrence scales more evenly, and the highest participation coefficient (panel F), with motifs linking across sections rather than remaining siloed. The only exception was betweenness evenness (panel E), which was slightly lower for the swarm case. This reduction does not reflect a loss of coherence, but rather the emergence of specialized motifs that serve as structural bridges — analogous to connective passages in human-composed music. Such specialization reduces statistical evenness but provides musically meaningful scaffolding. 

Taken together, the six network measures (Fig.~\ref{fig:longrange_variety}) reveal a consistent pattern: swarm-based generation achieves both stronger long-range coherence and greater structural variety than baseline systems. Higher long-range edge fraction and efficiency demonstrate that distant events are connected more frequently and bridged more directly, while longer community spans show that thematic material persists across time. At the same time, swarm outputs diversify recurrence scales (higher edge-length entropy) and integrate sections more thoroughly (higher participation coefficient), producing variety without degenerating into fragmentation. The one exception, reduced betweenness evenness, reflects emergent specialization: a limited set of motifs assume bridging roles, functioning like cadences or transitions in composed music. This combination of global coherence, temporal richness, and role differentiation mirrors the dynamics of a musical ensemble, where different voices contribute complementary functions to create an integrated whole.
We note that the swarm paradigm produces music that is globally coherent and varied through emergent specialization, in contrast to the traditional multi-agent baseline, which yields darker but fragmented and sectionalized structures, and the single-shot baseline, which is bright but repetitive and hub-dominated.
Detailed definitions and further analysis are provided in the Supplementary Information, Section~\ref{SI:deeper_structure}.

For a cross-scale analysis of the graphs (details of the analysis, see Section~\ref{SI:multiscale_topological}), we vary an edge-weight threshold $\tau$ and recompute community partitions at each level; stability between successive levels is summarized by the mean best-match Jaccard index $J_{\ell\to \ell+1}$ (Fig.~\ref{fig:fig19924}A)~\cite{EdelsbrunnerHarer2010,Ollivier2009,Bauer2021Ripser,Sankey1898ThermalEfficiency,RosvallBergstrom2010Alluvial}. The swarm consolidates earliest (at the strictest filtration levels, that is, the largest $\tau$ for which $J\ge 0.90$), the traditional system consolidates more gradually, and the single-shot baseline locks in only at the loosest thresholds, indicating late formation of macro-sections. To relate local structure to global flow on the full graph, Fig.~\ref{fig:fig19924}B maps each system by triangle $Z$-score (excess of 3-cliques relative to a degree-preserving null; higher $\Rightarrow$ richer local motifs) versus spectral gap $1-|{\lambda}_2|$ (higher $\Rightarrow$ faster mixing and stronger long-range integration). Swarm occupies the high-integration, high-motif regime; Traditional is motif-rich but less integrated; Single-Shot is weak on both axes. The full consolidation pathways are shown in the Supplementary Sankey diagrams (Figs.~\ref{fig:sankey_traditional}–\ref{fig:sankey_singleshot}), whose columns are ordered by decreasing $\tau$ (left = strict, only top-weight edges kept; right = looser, weaker edges added). Intuitively, strict thresholds reveal many small motif “islands” that fuse into larger sections as $\tau$ is relaxed; the swarm fuses earlier and yields a more navigable global structure without sacrificing motif density.

These cross-scale analyses (Fig.~\ref{fig:fig19924}--\ref{fig:sankey_singleshot}) explain how the end-state structures quantified earlier in the paper arise and why the systems differ. First, the early rise of community stability across thresholds predicts the shorter diffusion return-times and higher long-range efficiency we measured: once sections stabilize, weak ties organize into reliable shortcuts, improving traversability without indiscriminate densification. Second, the motif–vs–integration placement (triangle $Z$-score versus spectral gap) rationalizes the small-world regime observed in our metrics suite: the swarm jointly maximizes local motif richness and global integration, whereas the traditional system skews toward motif density with weaker stitching, and the single-shot baseline is deficient on both axes. Third, the braided split–merge patterns visible along the edge-weight filtration provide a mechanism for two seemingly opposed observations—higher participation across communities and slightly lower betweenness evenness—by revealing a few specialized connectors that scaffold joins while most material remains richly interlinked within sections. Fourth, the way micro-motifs fuse as the threshold is relaxed aligns with broader temporal persistence of thematic material (longer community spans) and a more even mix of recurrence scales (higher edge-length entropy). Finally, the coexistence of a small set of connectors with densely knit neighborhoods is consistent with the heavier-tailed degree behavior seen in our degree-distribution fits and with the core/bridge/periphery roles extracted from node-feature clustering. In aggregate, the cross-scale view supplies process-level evidence that ties together the diffusion, variety, persistence, degree-fit, and role analyses, clarifying why the swarm achieves coherent, navigable form without sacrificing motif richness.

\section{Conclusion}

This work investigated whether distributed orchestration of a frozen foundation model can yield
domain expertise in music composition without additional weight updates, contrasting often-used methods in machine learning for composition~\ref{fig:key_hypothesis}. Other than other multi-agent systems with \textit{a priori} determined roles for agents, here we posit that we can achieve complex compositional reasoning with \textit{in situ} updating of policies via feedback, realized in a number of ways. Rather than defining agents by role, we assign a particular set of musical notes to each agents (here, one bar), and define certain local or long-range/global interactions. 
To test this we introduced agentic systems that adapt on the fly via policy-over-prompts, peer and environmental feedback,
and an external episodic memory substrate~\ref{fig:alg_overview}. Across symbolic and computational-musicology
analyses, the results support our central hypothesis: coherent and stylistically convincing
compositions can emerge from collective adaptation alone. The swarm framework, in particular,
exhibited the richest local novelty and rhythmic diversity, while the centralized multi-agent
framework delivered stronger global organization; both substantially outperformed a single-shot
baseline. Agent ``personality'' trajectories converged toward stable but differentiated roles,
and tension/stability curves showed that long-range musical arcs can arise without any
gradient-based fine-tuning of model parameters (see, Figures~\ref{fig:creative_surprise} and \ref{fig:rhythmic_analysis}). Crucially, these emergent dynamics are stigmergic in nature: agents coordinate by leaving and sensing traces in a shared medium, much like pheromone trails in biological swarms, enabling decentralized yet coherent organization.

Beyond the specific musical setting, these findings articulate a broader design principle for
agentic AI: reuse general-purpose models as static capability providers and obtain
task specialization through system-level feedback, memory, and role structure. In this view,
creativity is not embedded solely in weights but emerges from organization—how agents sense,
remember, critique, and respond to one another. This separation of concerns (frozen capacities
vs.\ adaptive coordination) offers a scalable pathway to deploy specialized behavior rapidly,
with reduced data, cost, and risk compared to conventional fine-tuning pipelines.

Our evaluation thus far focused on short, symbolic pieces and a restricted instrumentation; human
listening studies and expert adjudication will be required to calibrate and complement the
automatic metrics used here. The reward and critique channels, while effective, remain
hand-designed and may bias stylistic choices; more principled, learnable reward models and
richer environment signals (e.g., audio-domain MIR features, performance dynamics) are natural
extensions. Future work will expand to longer forms and larger ensembles, introduce explicit
form-level objectives (e.g., hierarchical sections and themes), and perform ablations to
quantify the contribution of memory, role diversity, and peer assessment. We also foresee
applications beyond music—arrangement/orchestration, collaborative writing, design, and
scientific planning—where distributed, feedback-driven orchestration of foundation models could
offer a compelling alternative to task-specific fine-tuning.

Interesting insights also arise from comparing results across different global objectives. For instance, under Global Objective~\#1 (Fig.~\ref{fig:creative_surprise}), the swarm and multi-agent systems displayed a complementary division of labor: the swarm excelled at local novelty, rule-breaking, and melodic surprise, whereas the multi-agent system outperformed in global unpredictability and aggregate risk-taking. However, under Global Objective~\#2 (Fig.~\ref{fig:creative_surprise_objective2}), this balance shifted dramatically, with the swarm dominating across all measured dimensions—expectation violations, melodic surprise density, unpredictability, and risk-taking. This contrast underscores the sensitivity of emergent creative dynamics to the formulation of system-level objectives: small changes in how global goals are posed can tilt the balance between locally disruptive versus globally coordinated strategies. More broadly, this illustrates that objective design is not merely a matter of scoring outputs, but a powerful lever for shaping how collective intelligence distributes creativity across scales. A direct comparison of the scores revealed significantly higher complexity achieved in the swarm case, especially under Global Objective~\#2, as seen when comparing Figures~\ref{fig:score_example_101} with \ref{fig:score_example_102}.

Beyond mean values of harmonic tension and timbral descriptors, our analyses of temporal dynamics, key evolution, and alignment of novelty events suggest that swarm-generated compositions embody a distinctive balance: they achieve diversity without degenerating into incoherence, and structure without rigid metric constraints. This emergent complexity separates them from both the predictably structured single-shot outputs and the diffuse traditional baseline (Figures~\ref{fig:audio_creative_tension} and \ref{fig:pca_audio}).
Spectrograms illustrate alternating regions of concentrated and diffuse energy, while self-similarity matrices confirm that swarm structures are irregular yet richly interconnected, contrasting with the block-like repetition of single-shot and the sectional segmentation of traditional compositions (Figure~\ref{fig:mel_spectrograms}). Trajectory-based comparisons using dynamic time warping reinforce this finding, showing that swarm follows a distinct temporal arc of spectral entropy and tension. Taken together, these multimodal results indicate that swarm agents generate structural coherence through distributed, emergent organization rather than formulaic repetition, producing music that is neither constrained to predictable templates nor diffused into noise, but instead balances novelty and coherence as an emergent property of collective interaction.
This shows that swarm-generated compositions differ fundamentally from both single-shot and traditional baselines. They sustain higher harmonic tension, introduce frequent micro-level novelty not bound to barlines, and balance timbral brightness and entropy between the extremes of the two baselines. These results highlight that structural coherence in swarm music arises through emergent organization, producing outputs that are neither rigidly repetitive nor diffusely unstructured.

Our quantitative graph analyses (Fig.~\ref{fig:longrange_variety} and Fig.~\ref{fig:fig19924}) confirm and extend these observations: swarm-based generation achieves stronger long-range coherence, richer recurrence variety, and emergent specialization of bridging motifs, yielding globally integrated yet flexibly organized musical structures, consistent with the qualitative contrasts shown in Fig.~\ref{fig:ssm_circle}.

Our approach builds on a lineage that reaches back to Minsky’s Society of Mind and early expert
systems such as MYCIN, where competence emerged from interacting, specialized processes~\cite{minsky1986society,shortliffe1976mycin}.
Those systems relied on hand-coded rules and centralized control; by contrast, we instantiate
multiple copies of a \emph{frozen} foundation model as role-conditioned agents whose behavior
adapts non-parametrically via policy-over-prompts, external episodic memory, and environment/peer
feedback. This connects to contemporary agentic LLM frameworks, for instance ReAct planning/acting, Reflexion, LLM-as-judge/debate, Sparks-style self-reflection and critique loops, or
tool-using agents such as AutoGPT—yet differs in where learning resides: not in weight
updates, but in the system organization (roles, communication channels, and shared memory)~\cite{yao2022react,shinn2023reflexion,zheng2023llmasjudge,autogpt2023,wang2023voyager,ghafarollahi2025sparksmultiagentartificialintelligence}.
Our swarm design operationalizes the Society-of-Mind intuition with modern foundation models,
replacing earlier brittle rule arbitration with bottom-up coordination (pheromone-like signals and consensus)
that yields coherent musical structure without fine-tuning. 

\subsection*{Complementary interpretations}

Another useful perspective on these results comes from game theory. The role differentiation observed in the swarm resembles the convergence to a Nash equilibrium, where each agent settles into a strategy that is optimal given the strategies of others \cite{nash1950equilibrium,nash1951noncooperative}. In our case, rhythmic, harmonic, and thematic traits evolve until agents find stable roles relative to their peers, yielding global coherence without centralized control. The central critic system parallels a coordinated equilibrium enforced by an external planner, while the single-shot baseline lacks the iterative dynamics needed to approach equilibrium at all. This framing suggests that creative coordination in multi-agent music systems can be interpreted through the same lens as strategic interaction in economics and social systems \cite{fudenberg1991game,osborne1994course}, applying Nash’s original insights into the domain of aesthetic collective intelligence.

Related to these ideas, our dynamical analysis provides direct evidence that the swarm approaches a locally stable behavioral fixed point consistent with a Nash–like equilibrium (Figure~\ref{fig:nash_analysis_delta}). From the trait trajectories we fit a per–trait best–response map on the true line topology and find it contractive (spectral radii $\rho(J_k)<1$), implying convergence under iteration. Empirically, the observed mean step change $\overline{\Delta}_t$ descends into low–motion plateaus, and a simple affine calibration of the learned dynamics closely overlays the data, indicating that the remaining motion is well explained by a small noise floor rather than lack of equilibrium. Fixed–point residuals $\lvert f(x)-x\rvert$ at the terminal iterate are uniformly small across agents and traits, yielding a behavioral $\varepsilon$–Nash certificate: holding others fixed, no agent would change much next step. Notably, two basins appear; the later, deeper basin coincides with the higher score peak, suggesting convergence to a higher–quality configuration.

Our results can also be interpreted in light of Gödel’s incompleteness theorems \cite{godel1931formal}. In a similar way as Gödel proved that no sufficiently rich formal system can be both consistent and complete, a single monolithic model (single-shot baseline) is limited by the bounds of its learned parameters and cannot generate novelty outside its own formalism. By contrast, swarm and multi-agent orchestration loops effectively transcend these limits: agents critique one another and update their roles, thereby extending the system’s reach beyond what any single component can prove or generate. In this sense, the higher levels of creative surprise evidenced in both traditional and swarm models echoes Gödelian incompleteness: the inevitability that new truths (or musical ideas) emerge that cannot be foreseen or fully captured by a closed set of rules.

Although each agent in our swarm is itself a closed system - analogous to a formal calculus bounded by its own rules or, in our case, by frozen model weights - the collective transcends this limitation through interaction and emerging phase transitions to appropriate solutions. Gödel’s incompleteness theorems show that no single closed formal system can be both complete and consistent \cite{godel1931formal}, but multiple identical systems, when run in parallel with different contexts and allowed to interact, can extend one another’s reach. In logic, this principle was demonstrated by Turing in his work on ordinal logics \cite{turing1939} and by Feferman in his theory of iterated reflection principles \cite{feferman1962}, where successive applications of identical calculi, each importing statements undecidable in the previous, generate strictly stronger systems. In complexity theory, the leap from single to multiple identical provers provides a rigorous analogue: Shamir’s celebrated result \cite{shamir1992} showed that interaction between a weak verifier and a prover extends power from polynomial-time checking to all of PSPACE, while multi-prover interactive proofs demonstrated even greater leaps (MIP = NEXP \cite{babai1991}, MIP* = RE \cite{ji2021}). In automated reasoning, the Nelson–Oppen framework \cite{nelsonoppen1979} shows how multiple identical decision procedures, when combined, can decide statements no single procedure could handle alone. Outside mathematics and computer science, biology offers a natural instantiation: genetically identical cells differentiate into specialized tissues and organs through interaction, achieving organismal intelligence \cite{gilbert2010developmental}, while clonal immune cells diversify into effector and memory lineages through environmental cues \cite{murphy2012immunobiology}. Analogous principles are found in other areas of biology: complex organisms and ecosystems are composed of locally constrained building blocks (cells, molecules, tissues), yet global novelty and adaptability emerge through communication, feedback, and diversity of roles. In a similar way by which incompleteness is addressed in mathematics by constructing interacting hierarchies of closed theories, nature addresses the limits of uniform design by cultivating heterogeneity and diversity—principles that lead to resilience, adaptability, and emergent functionality via the Universality-Diversity-Principle (UDP) \cite{Giesa2011ReoccurringAnalogies,Ackbarow2008HierarchicalMaterials,Cranford2012Biomateriomics}. Finally, human collectives exhibit the same principle: groups of individuals with similar cognitive capacity display a higher-level collective intelligence factor $c$ that exceeds individual IQ, driven by interactional diversity \cite{woolley2010collective}. In all these cases, incompleteness at the local level is preserved, but interaction opens an indefinitely extensible horizon of capability (Table~\ref{table:godel_extension}). This work shows that the swarm framework mirrors these dynamics: although each agent is a closed instantiation of the same foundation model, the ensemble’s iterative feedback, memory, and role negotiation generate emergent behaviors unreachable by any single instance.  At its core, this work extends a generative agenda that has defined emerging research across materials, AI, and music, revealing how simple rules, when orchestrated collectively, can generate complexity, novelty, and beauty across scales.  

\subsection*{Bringing it all together}

The results reported in this paper lay out a possible strategy for the development of creative intelligence. Rather than pursuing ever-smarter monolithic models to designing ``smarter'' systems, we believe creativity and domain expertise emerges not from baked-in weight updates within a single network (at pre-training or post-training) but from the organization, coordination, and consensus-building of multiple general-purpose agents as the system solves problems and adapts. Our findings in the musical domain illustrate how decentralized orchestration can yield both novelty and coherence, producing results that single models cannot easily achieve.  This underscores the principle that adaptive expertise arises less from individual capacity than from patterns of coordination and interaction. Extending this to artificial intelligence suggests that the future of creative and problem-solving systems may lie less in scaling parameters and more in engineering the dynamics of collaboration—mechanisms that enable generalist models to self-specialize, negotiate roles, and collectively discover new solutions.

This research also echoes a range of earlier questions, including whether the deep beauty of music lies in its ability to reflect universal patterns of organization that span from molecules to human culture. In biology, proteins fold by balancing stability and flexibility~\cite{Yu2019AAI,Franjou2019SoundsProteins}, and hierarchical biological materials achieve strength through the orchestration of diverse motifs~\cite{Giesa2011ReoccurringAnalogies,buehler2023biomateriomics}. In a similar way, music balances novelty with coherence, and tension with resolution. This resonance is not merely metaphorical: prior work has shown that molecular vibrations can be translated directly into musical structures, revealing striking correspondences between the architectures of life and the architectures of sound~\cite{Milazzo0BioinspiredModeling,Buehler2023UnsupervisedDesigns,helmholtz1863sensations}. In this sense, the human perception of structure in music may emerge because the same principles that shape the universe—repetition, variation, emergence, and transformation—also shape the dynamics of our inner lives. Music thus becomes not only a reflection of the human soul, but also a mirror of the hidden order of nature itself.

We conclude by summarizing our central insights:  

\begin{itemize}
    \item Emergent creativity can arise without fine-tuning, through distributed orchestration of frozen foundation models.  
    \item Swarm intelligence emphasizes local novelty and rhythmic diversity, while centralized multi-agent coordination yields stronger global structure; both clearly outperform single-shot baselines.  
    \item The swarm’s advantage grows as the task gets longer and more open‑ended, suggesting that decentralized stigmergy scales.
    \item Creativity is shaped less by parameter updates and more by organization: how agents sense, remember, critique, and interact, positing creativity as an emergent phase transition. 
    \item The distribution of creativity across local and global scales is sensitive to objective design, underscoring the importance of carefully formulating system-level goals.  
    \item These principles extend beyond music, offering a generalizable pathway for collaborative AI systems in writing, design, and scientific discovery.  
\end{itemize}

\section{Materials and Methods}

We summarize key materials, methods, and algorithmic details here. Unless indicated otherwise, the frozen foundation model used for the experiments is \texttt{GPT-5-mini}~\cite{OpenAI2025GPT5SystemCard}.

\subsection{Multi-Agent Swarm Composition System with Critic}

We implement composition as a reinforcement-style, centrally evaluated multi-agent loop with two passes per iteration. An \(N\)-bar piece is decomposed into agents \(\{A_i\}_{i=1}^{N}\) (one per bar). The system exposes a fixed instrument set \(\mathcal{V}\) (exact voice names) and a context window \(k\) (parameter \texttt{neighbor\_bars}; \(k=-1\) shows the full draft, otherwise each agent sees bars \(|j-i|\le k\)). We refer to this variant as a ``traditional'' multi-agent swarm due to the presence of a central evaluator that has access to a global perspective. 

\paragraph{Data structures and constraints.}
Each agent maintains an episodic state
\[
\mathcal{M}_i=\{\texttt{agent\_id},\ \texttt{local\_objective},\ \texttt{past\_actions},\ \texttt{past\_feedback},\ \texttt{past\_objectives}\}.
\]
Bars are exchanged in a strict JSON schema:
\texttt{BarProposal}~$\to$~{\small\{\texttt{rationale}, \texttt{voices[ ]}\}} with \texttt{VoiceLine} entries {\small\{\texttt{instrument}, \texttt{notes[ ]}, \texttt{durations[ ]}\}}.
Instrument names must match \(\mathcal{V}\) exactly; per-voice durations must sum to one bar (e.g., 4.0 beats in common time). Proposals are parsed into \texttt{BarOutput} {\small\{\texttt{bar\_number}, \texttt{voices}, \texttt{rationale}, \texttt{feedback}\}} and later assembled with \texttt{music21}~\cite{cuthbert2010music21}.

\paragraph{System signals.}
At iteration \(t\) the system holds the current draft \(\mathcal{P}^{(t-1)}\), the global objective \(\Omega\), and the critic’s prior justification \(\psi^{(t-1)}\). A central critic LLM evaluates full drafts and returns a scalar score \(\sigma^{(t)}\in[0,1]\) and natural-language justification \(\psi^{(t)}\) (covering harmony, voice-leading, rhythm, form). By default the critic returns a single shared justification; agents self-extract bar-specific cues from \(\psi^{(t)}\).

\paragraph{Iteration \(t=1..T\) (two passes).}
\emph{Pass A — Reflection / objective update.}
Each agent reads \(\psi^{(t-1)}\) and \(\Omega\) and emits a one-sentence local objective \(o_i^{(t)}\) (JSON \texttt{\{ "new\_objective": ... \}}). This updates \(\texttt{local\_objective}\) and appends to \(\texttt{past\_objectives}\).

\emph{Pass B — Local composition.}
Each agent observes its visible context
\[
\mathcal{K}_i^{(t)}=\begin{cases}
\mathcal{P}^{(t-1)} & k=-1,\\
\{\,b_j^{(t-1)}:|j-i|\le k\,\} & k\ge 0,
\end{cases}
\]
and, conditioned on \(\Omega\) and \(o_i^{(t)}\), produces a strict-JSON bar \(b_i^{(t)}\) over the fixed voice set \(\mathcal{V}\). Bars are assembled to form \(\mathcal{P}^{(t)}\). Malformed tokens are sanitized (e.g., \texttt{clean\_note\_name}), and \texttt{rest} is permitted.

\emph{Central critique and selection.}
The critic evaluates \(\mathcal{P}^{(t)}\) to return \((\sigma^{(t)},\psi^{(t)})\). We track the best-so-far draft \((\mathcal{P}_\star,\sigma_\star)\) across iterations. Iteration artefacts are persisted: \texttt{iter\_\{t\}/bars.json}, \texttt{agent\_states.json}, \texttt{critic\_feedback.json}, and renderings (\texttt{piece.mid}, \texttt{piece.musicxml}, piano-only variants, and WAV via MusicXML\(\to\)synthesis).

\paragraph{Reinforcement learning view and shaping signal.}
We view each agent as sampling actions \(a_i^{(t)}=b_i^{(t)}\sim \pi_i(\cdot\mid s_i^{(t)})\) where
\[
s_i^{(t)}=(\mathcal{K}_i^{(t)}, \Omega, \psi^{(t-1)}, \mathcal{M}_i^{(t-1)}, o_i^{(t)}).
\]
A global improvement signal \(\Delta\sigma^{(t)}=\sigma^{(t)}-\sigma^{(t-1)}\) is used as a shaped reward for analysis, selection, and objective refinement. In this implementation we do not update model weights online; learning occurs as \emph{policy-over-prompts}—agents adapt \(o_i^{(t+1)}\), sampling controls, and retrieval cues from \(\psi^{(t)}\) and memory, while the critic provides a stable scalar gradient for convergence diagnostics.

\paragraph{Initialization and termination.}
At \(t=0\) we seed each bar with C-major quarter notes per voice and initialize all agents with an \emph{initial objective}. We iterate for \(T\) steps or until a plateau in \(\sigma^{(t)}\). Final artefacts include \texttt{score\_history.svg}, \texttt{final\_piece.mid/.musicxml/.wav}, and \texttt{best\_composition.json} (best score, iteration index, and composition snapshot). 

\begin{algorithm}[h]
  \caption{{Multi-Agent Composition System with Critic}}
  \label{alg:central_critic}
  \begin{algorithmic}[1]
    \Require Bars $N$; voices $\mathcal{V}$; iterations $T$; context window $w$ ($w=-1$ for global); objective $\Omega$
    \Ensure Best composition $\mathcal{P}_\star$ with score $\sigma_\star$
    \State \textbf{Initialize:} empty/seed piece $\mathcal{P}^{(0)}$; per-bar agent states $\{\mathcal{M}_i\}_{i=1}^N$; score trace $S \gets [\,]$; $(\mathcal{P}_\star,\sigma_\star)\gets(\mathcal{P}^{(0)},-\infty)$
    \For{$t = 1$ to $T$}
      \State \textbf{Pass A — Reflection / Objective Update}
      \For{$i = 1$ to $N$}
        \State $\phi_i^{(t)} \gets \textsc{SummarizeCriticFeedbackForBar}(i,\ \mathcal{P}^{(t-1)},\ \Omega)$
        \State $o_i^{(t)} \gets \textsc{ProposeLocalObjective}(\mathcal{M}_i,\ \phi_i^{(t)},\ \Omega)$
      \EndFor
      \State \textbf{Pass B — Local Composition Under New Objectives}
      \State $\mathcal{P}^{(t)} \gets [\,]$
      \For{$i = 1$ to $N$}
        \State $\mathcal{K}_i^{(t)} \gets \textsc{ExtractContext}(\mathcal{P}^{(t-1)},\ i,\ w,\ \mathcal{V})$
        \State $b_i^{(t)} \gets \textsc{GenerateBar}(\mathcal{M}_i,\ o_i^{(t)},\ \mathcal{K}_i^{(t)},\ \Omega,\ \mathcal{V})$
        \State Append $b_i^{(t)}$ to $\mathcal{P}^{(t)}$
      \EndFor
      \State \textbf{Global Critique and Memory Update}
      \State $(\sigma^{(t)},\ \psi^{(t)}) \gets \textsc{AssessPiece}(\mathcal{P}^{(t)},\ \Omega)$ \Comment{$\psi^{(t)}$: structured NL justification, per-metric scores}
      \State $S \gets S \cup \{\sigma^{(t)}\}$;\quad $\{\mathcal{M}_i\}\gets\textsc{UpdateAgentMemories}(\{\mathcal{M}_i\},\ \psi^{(t)},\ \mathcal{P}^{(t)})$
      \If{$\sigma^{(t)} \ge \sigma_\star$} \quad $(\mathcal{P}_\star,\sigma_\star)\gets(\mathcal{P}^{(t)},\sigma^{(t)})$ \EndIf
      \State \textbf{Persistence (optional):} \textsc{SaveArtifacts}$(\mathcal{P}^{(t)},\psi^{(t)},S)$
      \If{\textsc{Converged}$(S)$} \textbf{break} \EndIf
    \EndFor
    \State \Return $(\mathcal{P}_\star,\sigma_\star)$
  \end{algorithmic}
\end{algorithm}

\subsection{Biologically-Inspired Decentralized Swarm Composition System}

We implement composition as a decentralized, reinforcement-style swarm process with \emph{no} central planner. An \(N\)-bar piece is realized by agents \(\{A_i\}_{i=1}^{N}\) (one per bar) that interact only through (i) local musical context, (ii) peer assessment, and (iii) shared environmental signals (“musical pheromones”). Policies are adapted \emph{without online weight updates} via a “policy-over-prompts” mechanism that evolves each agent’s \emph{personality vector}.

\paragraph{Environment and signals.}
A shared environment \(\mathcal{E}^{(t)}\) maintains: a pheromone map \(\Phi^{(t)}=\{\texttt{melodic\_motif},\texttt{rhythm},\ldots\}\), a global energy level \(g^{(t)}\in[0,1]\) (normalized from aggregate pheromone activity), and a list of emergent themes detected when strong, repeated patterns occur in multiple locations. Each pheromone \(p\in\Phi^{(t)}\) stores \((\texttt{pattern\_type}, \texttt{pattern\_data}, \texttt{strength}\in[0,1], \texttt{source\_bar}, \texttt{success\_score}, \texttt{timestamp})\) and undergoes decay; successful patterns are reinforced. Each iteration: decay weak pheromones, update \(g^{(t)}\), and detect themes.

\paragraph{Agents, memory, personality.}
Agent \(A_i\) (bar \(i\)) holds (1) a bounded episodic memory \(\mathcal{M}_i\) (recent compositions and received peer feedback), and (2) a personality vector
\[
\theta_i=\{\texttt{risk\_taking},\ \texttt{harmonic\_sensitivity},\ \texttt{rhythmic\_drive},\ \texttt{theme\_loyalty},\ \\ \texttt{neighbor\_influence}\}\in[0.1,0.9]^5,
\]
initialized by mode \(\in\{\texttt{uniform},\texttt{random}\}\). Personality governs exploration vs.\ conservatism, harmonic strictness, rhythmic emphasis, motif repetition, and susceptibility to neighbor influence.

\paragraph{Observation and action.}
At iteration \(t\), agent \(A_i\) observes
\[
o_i^{(t)}=\big(\mathcal{K}_i^{(t)},\ \Phi_i^{(t)}(r),\ \Omega,\ \mathcal{M}_i\big),
\]
where \(\mathcal{K}_i^{(t)}\) is the local score context (neighbor bars within sensing radius \(r\); if unavailable, falls back to global view), \(\Phi_i^{(t)}(r)\) are pheromones sensed within range \(r\), and \(\Omega\) is the global objective. The action \(a_i^{(t)}\) is an \emph{EnhancedBarProposal} \(b_i^{(t)}\) containing strict-JSON \texttt{voices} (per-voice \((\text{note},\text{duration})\) pairs summing to 4.0 beats) plus rich natural-language fields: \texttt{rationale}, \texttt{detailed\_reasoning}, \texttt{personality\_reflection}, and \texttt{pheromone\_interpretation}. After emitting \(b_i^{(t)}\), the agent deposits pheromones derived from motifs and rhythms detected in its bar; \(\Phi^{(t)}\) is updated accordingly.

\paragraph{Distributed peer assessment and consensus.}
After assembling \(\mathcal{P}^{(t)}=\{b_i^{(t)}\}_{i=1}^N\), agents perform \emph{local} peer assessments. For each bar \(j\), evaluators are agents in a neighborhood \(|i-j|\le k\) (peer-assessment range \(k\)). Each evaluator returns a \emph{DetailedPeerAssessment} with scalar ratings in \([0,1]\) for \(\texttt{musical\_quality}\), \(\texttt{objective\_alignment}\), \(\texttt{swarm\_cooperation}\), \(\texttt{innovation\_value}\), plus natural-language feedback fields (e.g., \texttt{musical\_feedback}, \texttt{cooperation\_feedback}, \texttt{innovation\_commentary}, \texttt{suggestions}). A consensus module aggregates per-bar metrics by averaging and estimates agreement as \(1-\)std.\ deviation across raters. The \emph{overall swarm satisfaction} \(\sigma^{(t)}\in[0,1]\) is defined as the mean of all aggregated metric values.

\paragraph{Shaped rewards and adaptation.}
For analysis and adaptation, each bar receives a shaped signal \(r_j^{(t)}\) from its consensus metrics (e.g., weighted combination of \(\texttt{musical\_quality}, \texttt{alignment}, \texttt{cooperation}, \texttt{innovation}\)). Agents append received peer comments for their own bar to \(\mathcal{M}_i\). Personality evolution is performed by an LLM-guided \emph{policy-over-prompts} step: given (i) recent peer feedback for bar \(i\), (ii) \(\Omega\), and (iii) neighbor context, the agent proposes a trait update \(\theta_i\!\leftarrow\!\theta_i+\Delta\theta_i\) with bounded adjustments (typically \(\pm 0.05\) to \(\pm 0.20\) per trait, clamped to \([0.1,0.9]\)). If the LLM step fails, a tiny random drift (\(\pm 0.01\)) is applied as a fallback. Environment signals are also updated from consensus (reinforce pheromones associated with well-rated patterns, attenuate weak ones), coupling social judgment to the shared medium.

\paragraph{Loop structure.}
Each iteration executes:
\begin{enumerate}[leftmargin=1.2em, itemsep=2pt, topsep=2pt]
    \item \textbf{Environment update:} decay pheromones, update global energy, detect emergent themes.
    \item \textbf{Swarm composition (parallel):} for each \(i\): sense \((\mathcal{K}_i^{(t)},\Phi_i^{(t)}(r)) \rightarrow b_i^{(t)} \rightarrow\) deposit pheromones from motifs/rhythms.
    \item \textbf{Consensus:} neighborhood peer assessments \(\rightarrow\) aggregate metrics, agreement, and \(\sigma^{(t)}\).
    \item \textbf{Adaptation:} write peer feedback to \(\mathcal{M}_i\); LLM-guided evolution of \(\theta_i\); update environment from consensus.
    \item \textbf{Selection/persistence:} track best \((\mathcal{P}_\star,\sigma_\star)\); save JSON/MIDI/MusicXML/WAV and analysis plots (score trajectory, pheromone stats, personality radar charts, evolution traces).
\end{enumerate}

\paragraph{Initialization, translation, termination.}
Agents are created for bars \(1..N\) with \(\theta_i\) set by \texttt{personality\_init\_mode}\(\in\)\{\texttt{uniform},\texttt{random}\}. If enabled, the global objective \(\Omega\) is pre-assessed/translated to enforce musicality and instrument constraints before the loop. We iterate for a fixed budget or until satisfaction plateaus; outputs include \(\mathcal{P}^{(t)}\), \((\mathcal{P}_\star,\sigma_\star)\), swarm diagnostics, and environment statistics.

\begin{algorithm}[h]
  \caption{Biologically-Inspired Decentralized Swarm Composition System}
  \label{alg:swarm}
  \begin{algorithmic}[1]
    \Require Bars $N$; voices $\mathcal{V}$; iterations $T$; objective $\Omega$; neighbor range $k$; sensing radius $r$
    \Ensure Best swarm composition $\mathcal{P}_\star$ with consensus score $\sigma_\star$
    \State \textbf{Initialize:} environment $E$ (musical pheromones, global energy, theme memory); agents $\{a_i\}_{i=1}^N$ with local states $\{\mathcal{M}_i\}$ and personalities; seed $\mathcal{P}^{(0)}$; $(\mathcal{P}_\star,\sigma_\star)\gets(\mathcal{P}^{(0)},-\infty)$
    \For{$t = 1$ to $T$}
      \State \textbf{Environment Update} $E \gets \textsc{UpdateEnvironment}(E)$ \Comment{pheromone decay/reinforce, energy, emergent themes}
      \State \textbf{Local Sensing \& Proposal (parallel for all $i$)}
      \ForAll{$i \in \{1,\dots,N\}$ \textbf{in parallel}}
        \State $\text{obs}_i \gets \textsc{Sense}(a_i,\ E,\ r)$
        \State $\mathcal{K}_i^{(t)} \gets \textsc{BuildLocalContext}(\mathcal{P}^{(t-1)},\ i,\ r,\ \mathcal{V})$
        \State $(\hat{b}_i^{(t)},\ \Delta \Phi_i) \gets \textsc{ComposeLocally}(\mathcal{M}_i,\ \text{obs}_i,\ \mathcal{K}_i^{(t)},\ \Omega)$
        \State \textsc{DepositSignal}$(E,\ \Delta \Phi_i,\ i,\ t)$
      \EndFor
      \State $\mathcal{P}^{(t)} \gets \textsc{AssembleBars}(\{\hat{b}_i^{(t)}\}_{i=1}^N)$
      \State \textbf{Distributed Peer Assessment \& Consensus}
      \ForAll{$i$ \textbf{in parallel}}
        \State $\Pi_i \gets \textsc{PeerAssessLocalNeighborhood}(\mathcal{P}^{(t)},\ i,\ k,\ \Omega)$ \Comment{NL feedback + per-metric scores}
      \EndFor
      \State $\mathcal{C}^{(t)} \gets \textsc{AggregateConsensus}(\{\Pi_i\})$ \Comment{agreement levels, consensus per bar/metric}
      \State $E \gets \textsc{UpdateSignalsFromConsensus}(E,\ \mathcal{C}^{(t)})$; \quad $\{\mathcal{M}_i\}\gets\textsc{EvolvePersonalities}(\{\mathcal{M}_i\},\ \mathcal{C}^{(t)},\ \Omega)$
      \State $\sigma^{(t)} \gets \textsc{ConsensusScore}(\mathcal{C}^{(t)})$ \Comment{or optional external global critic}
      \If{$\sigma^{(t)} \ge \sigma_\star$} \quad $(\mathcal{P}_\star,\sigma_\star)\gets(\mathcal{P}^{(t)},\sigma^{(t)})$ \EndIf
      \State \textbf{Persistence (optional):} \textsc{SaveArtifacts}$(\mathcal{P}^{(t)},\mathcal{C}^{(t)},E)$
    \EndFor
    \State \Return $(\mathcal{P}_\star,\sigma_\star)$
  \end{algorithmic}
\end{algorithm}

\subsection{Single-shot approach}

We provide a one-pass, non-iterative, baseline in which the model generates the entire \(N\)-bar composition in a single call, without reflection, peer critique, or central evaluation updates. The approach fixes the instrument set \(\mathcal{V}\) (exact voice names), tempo \(q\), time signature, and global objective \(\Omega\), and returns a machine-parsable score.

The system takes \((N,\mathcal{V},q,\Omega)\) and constructs a strict prompt that specifies:
(i) the exact, allowed instrument names \(\mathcal{V}\);
(ii) a bar count of \(N\);
(iii) per-voice sequences of \((\texttt{note},\texttt{duration})\) tokens whose durations sum to one bar (e.g., 4.0 beats in common time);
(iv) a JSON contract with top-level fields for metadata (key, time signature, tempo), per-bar content, and a brief natural-language \texttt{rationale}.
The prompt explicitly forbids extra fields, alternative instrument labels, or free-form text outside the JSON.

We assemble a comprehensive instruction \(\mathcal{U}\) from the above constraints and call the LLM once. No iterative refinement or external scoring is performed.

The response \(y\) is parsed against a strict schema (e.g., \texttt{Pydantic}) into an internal score representation. Validation enforces:
(1) instrument names match \(\mathcal{V}\) exactly;
(2) per-voice durations sum to one bar;
(3) note tokens are valid (with a minimal sanitizer that maps malformed or out-of-range tokens to safe defaults, and accepts \texttt{rest});
(4) bar indices cover \(1{:}N\) without duplication.
If any hard constraint fails, the run is flagged; as a baseline we do not request a repair pass.

Validated bars are assembled into a score \(\mathcal{P}\) (one staff per \(v\in\mathcal{V}\)) and rendered to MIDI and MusicXML; optional audio (WAV) can be produced from MusicXML via a software synthesizer. We persist the raw JSON response, the parsed score object, and the rendered files for downstream analysis.

This approach removes reinforcement-style credit assignment and social/central critique, serving as a lower-bound baseline for quality and a control for dataflow (prompt\(\rightarrow\)parse\(\rightarrow\)render) without iterative learning signals. Any improvements in multi-agent or swarm settings can thus be attributed to reflection, critique, consensus, and/or policy-over-prompts adaptations absent here.
 Outputs include the raw JSON composition, \texttt{.mid}, \texttt{.musicxml}, optional \texttt{.wav}, and a minimal run manifest (inputs, schema version, timings).

\begin{algorithm}[h]
  \caption{Single-Shot Composition}
  \label{alg:single_shot}
  \begin{algorithmic}[1]
    \Require Bars $N$; voices $\mathcal{V}$; tempo $q$; objective $\Omega$
    \Ensure One-pass composition $\mathcal{P}$ and rendered score
    \State $\mathcal{U} \gets \textsc{BuildSingleShotPrompt}(N,\ \mathcal{V},\ q,\ \Omega)$ \Comment{includes strict JSON schema + musical constraints}
    \State $y \gets \textsc{OnePassModelCall}(\mathcal{U})$ \Comment{no iterative reflection or critique}
    \State $(\mathcal{P},\ \mu) \gets \textsc{ParseAndConstructScore}(y,\ \mathcal{V},\ N,\ q)$ \Comment{$\mu$: parsed metadata/analyses}
    \State \textsc{SaveArtifacts}$(\mathcal{P},\ \mu)$ \Comment{JSON, MIDI, MusicXML; WAV}
    \State \Return $\mathcal{P}$
  \end{algorithmic}
\end{algorithm}

\subsection{Musical Analysis and Comparison}
Musical analyses were performed using the \texttt{Music21} library \cite{cuthbert2010music21} with custom algorithms for advanced metrics. MusicXML compositions were preprocessed to remove empty bars for fair comparison across generation systems.

\textbf{Expectation Violations:} Melodic expectation analysis identified violations where stepwise motion (less or equal than 2 semitones) was followed by
   large leaps (>7 semitones). Surprise levels were calculated as $\min(1.0, \text{interval}|/12)$.

  \textbf{Melodic Surprise Density:} Large intervallic leaps (>7 semitones) were counted and normalized by total note count as surprises notes.

  \textbf{Musical Unpredictability:} Predictability scores were computed through interval sequence pattern analysis, with unpredictability
   defined as $1 - \text{predictability\_score}$.

  \textbf{Creative Risk Taking:} A weighted composite metric combining large leap ratio, expectation violation density, and
  musical unpredictability at equal weight.

  Score rendering utilized musicxml2ly and LilyPond for PDF generation with customized metadata (author: "AI Generated", title: left blank).

  \textbf{Tonal Tension Analysis:} Tonal stability was computed using \texttt{Music21}'s key detection algorithms combined with chord-to-key
  relationship analysis. Tension curves were calculated by measuring harmonic dissonance levels and key distance metrics across musical
  measures. Tension peaks and valleys were identified using local maxima/minima detection with adaptive thresholds based on the
  composition's dynamic range. We do this by collecting all the
  notes sounding at each time point across all voices, which gives the complete harmonic picture needed for accurate tonal tension
  calculation.

  \textbf{Rhythmic Distribution Analysis:} Note durations were extracted using \texttt{Music21}'s quarterLength attribute and systematically
  categorized across all composition systems. Duration values were sorted from shortest to longest (e.g., 0.0625 = 64th note, 1.0 =
  quarter note, 4.0 = whole note) for consistent comparison. Side-by-side visualization enabled direct comparison of rhythmic vocabulary
  diversity across Single-Shot LLM, Multi-Agent, and Swarm Intelligence approaches. Each panel employed a plasma color gradient to
  distinguish duration categories while maintaining visual consistency with individual analysis plots.

\subsection{Audio Rendering}

We render MIDI files using Ableton Live 12.2.5, using the Roland SRX Piano II Superb Grand patch. Audio files are saved as WAV and MP3; WAV files are used for audio analysis (see, Figure~\ref{fig:roland}).

\subsection{Audio analysis}

We analyzed all compositions using the \texttt{librosa} Python library~\cite{mcfee2015librosa}, following standard workflows in music information retrieval (MIR). Each audio file was down-mixed to mono, resampled to 22.05~kHz, and transformed via short-time Fourier transforms (STFT; 2048-sample windows, 512-sample hop). From these representations, we extracted both harmonic and timbral descriptors. Harmonic tension was quantified as the complement of triad-template correlations, following cognitive models of consonance and dissonance~\cite{lerdahl2001tonal,chew2002spiral}. 

Structural organization was examined through chroma-based self-similarity matrices and Jensen–Shannon (JS) divergence~\cite{lin1991divergence} between adjacent chroma windows, computed both framewise and beat-synchronously to capture local recontextualizations versus meter-aligned changes. JS divergence, a symmetric and bounded measure of distributional dissimilarity derived from Kullback–Leibler divergence, was used to quantify novelty between consecutive chroma windows, with higher values indicating greater harmonic change.

Timbral properties were characterized using spectral centroid, bandwidth, and entropy, widely employed descriptors of brightness, spectral spread, and flatness~\cite{peeters2004large}. Onset strength was used as a proxy for rhythmic density, and log-mel spectrograms provided complementary visualization of timbral texture. Temporal trajectories of descriptors were compared across pieces using dynamic time warping (DTW) with derivative preprocessing (DDTW), normalized by path length and constrained by a Sakoe–Chiba band~\cite{sakoe1978dynamic}. Finally, global descriptors and novelty densities were contextualized via principal component analysis (PCA)~\cite{jolliffe2016principal}, enabling visualization of method-specific clustering in reduced feature space.

\subsection{Graph-theoretic analysis of self-similarity based on audio analysis}
For each audio file we constructed a self-similarity matrix (SSM) from concatenated chroma and onset-strength features, using cosine similarity between frame-wise vectors. From each SSM we derived a weighted, undirected similarity graph by retaining $k$ nearest neighbors per node. We then computed a range of global and local graph measures: number of nodes and edges, graph density, connected components, average clustering coefficient and transitivity~\cite{watts1998collective}, degree assortativity~\cite{newman2002assortative}, modularity (via greedy community detection)~\cite{newman2004fast}, betweenness centrality~\cite{freeman1977set}, and average shortest path length on the giant connected component (weighted by inverse similarity). To assess small-world structure, we calculated the Watts–Strogatz small-worldness index $\sigma = (C/C_{\mathrm{rand}})/(L/L_{\mathrm{rand}})$~\cite{humphries2008network}, where $C$ and $L$ are clustering and path length of the empirical graph, and $C_{\mathrm{rand}},L_{\mathrm{rand}}$ are the corresponding averages from Erdős–Rényi graphs with the same $n$ and $m$~\cite{erdos1959random}. We also characterized connectivity diversity using normalized degree entropy. Finally, empirical degree distributions were compared to candidate generative models (Poisson, lognormal, and power-law tails) by maximum-likelihood fitting~\cite{clauset2009power}; for each model we computed log-likelihood, Akaike information criterion (AIC)~\cite{akaike1974new}, Kolmogorov–Smirnov distance, and the best-fitting parameters.

\subsection{Audio Summary}
Supplementary Audio A1 presents an audio summary of this paper in the style of a podcast, created using PDF2Audio (\url{https://huggingface.co/spaces/lamm-mit/PDF2Audio}~\cite{ghafarollahi2024sciagentsautomatingscientificdiscovery}). The audio format in the form a conversation enables reader to gain a broader understanding of the results of this paper, including expanding the broader impact of the work. The transcript was generated using the \texttt{GPT-5} model~\cite{OpenAI2025GPT5SystemCard} from the final draft of the paper.

\section*{Data availability}

Codes and additional materials are available at \url{https://github.com/lamm-mit/MusicSwarm}.  


\section*{Conflicts of Interest}

The author declares no conflicts of interest of any kind.

\bibliographystyle{naturemag}

\bibliography{references,references-Mendeley}


\clearpage
\appendix
\renewcommand{\thesection}{S\arabic{section}}
\renewcommand{\thefigure}{S\arabic{figure}}
\renewcommand{\thetable}{S\arabic{table}}
\renewcommand{\thetextbox}{S\arabic{textbox}}

\setcounter{section}{0}
\setcounter{figure}{0}
\setcounter{table}{0}
\setcounter{textbox}{0}

\newpage

\section*{\vspace{1cm} \centering \Large  \sffamily \\ \textbf{Supplementary Information}}
\vspace{4cm}

\begin{center}
    {\LARGE\sffamily  \textbf{MusicSwarm: Biologically Inspired Intelligence for Music Composition }} \\[1em] 
    
\end{center}

\vspace{3cm}

\begin{center}
    {\large Markus J. Buehler} \\[1.em]  

    Laboratory for Atomistic and Molecular Mechanics\\Center for Computational Science and Engineering \\
        Schwarzman College of Computing \\
	Massachusetts Institute of Technology\\
	Cambridge, MA 02139, USA \\
[0.5em]
    
    {\normalsize mbuehler@MIT.EDU}
\end{center}

\newpage
\section{Analogical Particle System}

\begin{figure}[h!]
    \centering
    \includegraphics[width=1.\textwidth]{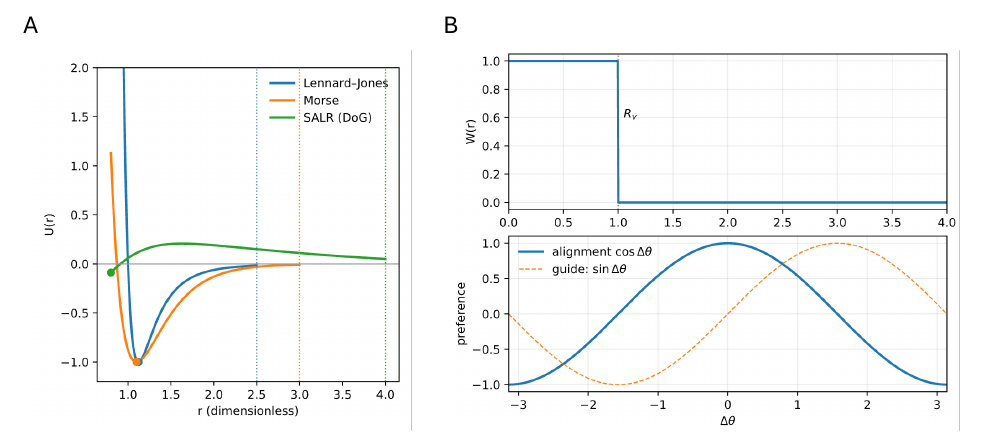}
   \caption{Local interaction rules used in the analogy.
    A: Pair potentials for equilibrium systems: Lennard--Jones (short-range repulsion with a single cohesive well) \cite{LennardJones1924}, Morse (tunable well depth and range, controlling effective bond elasticity) \cite{Morse1929}, and SALR (short-range attraction plus long-range repulsion, leading to microphase clustering) \cite{SeulAndelman1995,Stradner2004}. 
    B: Alignment interactions in the Vicsek active-matter model \cite{Vicsek1995}: top, neighborhood weighting function with cutoff radius \(R_v\); bottom, angular alignment preference modeled by a cosine kernel. 
    Together, these four interaction laws span canonical cases of conservative equilibrium potentials and nonequilibrium alignment rules, forming the basis of our particle-system analogy for multi-agent interactions.}
    
    \label{fig:interactions_analogy}
\end{figure}

We simulated \(N=1024\) point particles in a two–dimensional periodic square of side \(L=\sqrt{N/\rho}\) at number density \(\rho=0.8\) (dimensionless units). For equilibrium assemblies (Lennard–Jones, Morse, SALR) we integrated overdamped Langevin dynamics with Euler–Maruyama,
\[
\mathrm{d}\mathbf{r}_i=\mu\!\sum_{j\neq i}\mathbf{F}_{ij}\,\mathrm{d}t+\sqrt{2\,T(t)}\,\mathrm{d}\mathbf{W}_i,
\qquad 
\mathbf{F}_{ij}=-\nabla U(r_{ij}),
\]
using the minimum–image convention and a cell–linked neighbor list (rebuilt every 10 steps). We used \(\mu=1\), time step \(\Delta t=10^{-3}\), and an annealing schedule
\[
T(t)=T_{\min}+\bigl(T_0-T_{\min}\bigr)\!\left(1-\tfrac{t}{S-1}\right)^{p},
\quad (T_0=0.4,\;T_{\min}=0.02,\;p=1.5,\;S=3000).
\]
Potentials (truncated at \(r_\mathrm{c}\) without shifting) were
\[
U_{\mathrm{LJ}}(r)=4\varepsilon\!\left[\left(\tfrac{\sigma}{r}\right)^{12}-\left(\tfrac{\sigma}{r}\right)^6\right],\quad
\sigma=1,\;\varepsilon=1,\;r_\mathrm{c}=2.5\sigma,
\]
\[
U_{\mathrm{Morse}}(r)=D_e\!\left(1-e^{-\alpha(r-r_e)}\right)^2-D_e,\quad
D_e=1,\;\alpha=3,\;r_e=1.1,\;r_\mathrm{c}=3.0,
\]
\[
U_{\mathrm{SALR}}(r)=-A\,e^{-(r/\sigma_a)^2}+B\,e^{-(r/\sigma_r)^2},\quad
A=1,\;\sigma_a=0.8,\;B=0.30,\;\sigma_r=3.0,\;r_\mathrm{c}=4.0.
\]
Active flocking (Vicsek) used headings \(\theta_i\) updated by the mean neighbor orientation within radius \(R_v=1.0\) plus uniform angular noise \(\xi_i\!\sim\!\mathcal{U}[-\eta/2,\eta/2]\) with \(\eta=0.25\), then self–propelled motion \(\mathbf{r}_i\leftarrow \mathbf{r}_i+v_0(\cos\theta_i,\sin\theta_i)\Delta t\) with \(v_0=0.03\). We ran \(S=2000\) steps for the Vicsek model and enforced periodic boundaries. See Figure~\ref{fig:interactions_analogy} for a visualization of the interactions.

Visualizations show final configurations colored by local order: for LJ and Morse we computed the hexatic parameter
\[
\psi_6(i)=\frac{1}{n_i}\sum_{j\in\mathcal{N}_i} e^{\,6\mathrm{i}\theta_{ij}},\qquad |\psi_6|\in[0,1],
\]
using neighbors within \(r=1.5\sigma\); for SALR we used normalized neighbor counts within \(r=1.25\sigma\) as a local–density proxy; for Vicsek we colored by normalized heading \(\theta/(2\pi)\). The radial distribution function \(g(r)\) was computed from pair distances (to \(r_{\max}=L/2\)) via a histogram with bin width \(\Delta r\) and normalized as
\[
g(r)=\frac{H(r)}{N\,2\pi r\,\Delta r\,\rho},
\]
where \(H(r)\) is the number of pairs in the shell \([r,r+\Delta r)\). Random initial positions were uniform in the box (seed \(=42\)). Figures were rendered with scatter markers and per–panel colorbars; \(g(r)=1\) is indicated by a dashed baseline.

This experiment was designed to isolate the role of local interactions in producing qualitatively different global organizations under controlled conditions. We fixed particle number (\(N=1024\)), density, integration scheme, and annealing schedule, so that the only salient difference across panels is the interaction law. The three equilibrium pair potentials span canonical cases: Lennard–Jones (short-range repulsion with a single cohesive well) yields dense crystals and serves as a baseline for close-packed order \cite{LennardJones1924}; Morse provides a tunable well depth and range, capturing stiffer/softer ``bond elasticity'' and shifting crystallinity vs. fluidity \cite{Morse1929}; and SALR (short-range attraction, long-range repulsion) represents *competing* interactions that generically drive microphase separation (clusters/stripes) rather than bulk phase separation \cite{SeulAndelman1995,Stradner2004}. To contrast equilibrium self-assembly with nonequilibrium patterning, we also include the Vicsek active-matter model, where alignment interactions plus noise generate flocking bands and polar order that are not captured by isotropic conservative forces \cite{Vicsek1995}. Readouts are chosen to be interaction-agnostic: configurations (geometry), and \(g(r)\) (two-point structure), so differences in the figure can be directly attributed to the potentials rather than to protocol or visualization.

\begin{figure}[h!]
    \centering
    \includegraphics[width=.8\textwidth]{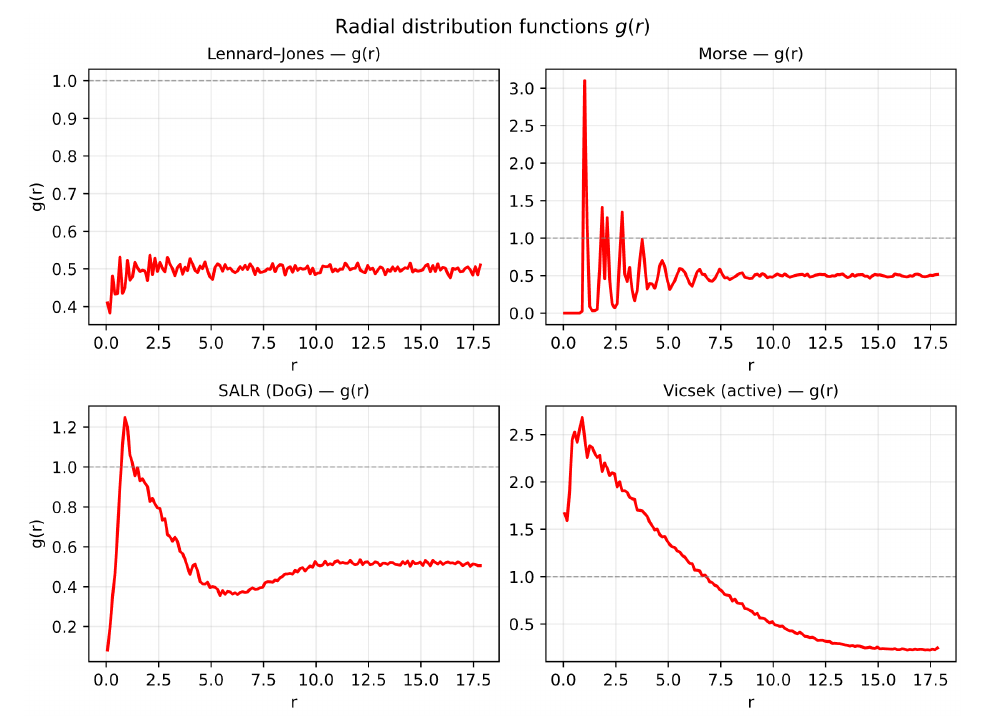}
    \caption{Radial distribution functions (RDFs) \(g(r)\) for \(N=1024\) under different local interaction rules.
    Top left (Lennard--Jones): nearly flat profile close to unity, indicating the absence of long-range crystalline order \cite{LennardJones1924}. 
    Top right (Morse): sharp peaks at regular intervals reflecting strong crystallinity and long-range correlations \cite{Morse1929}. 
    Bottom left (SALR): a pronounced first peak followed by damped oscillations and a split secondary maximum, characteristic of cluster-forming microphase order \cite{SeulAndelman1995,Stradner2004}. 
    Bottom right (Vicsek): broad decaying correlations without oscillatory structure, consistent with transient flocking clusters and the lack of equilibrium crystalline order \cite{Vicsek1995}.}
    \label{fig:g_r_result}
\end{figure}

\newpage
\section{Equilibrium analysis}

\subsection{Mean step-to-step trait change}

Let $i\in\{1,\dots,N\}$ index agents, $k\in\{1,\dots,K\}$ index traits, and $t=0,\dots,T$ index iterations.
Denote the (normalized) trait value of agent $i$ on trait $k$ at iteration $t$ by $x_{i,k}^{(t)}\in[0,1]$.
For each step $t\to t{+}1$, we compute the \emph{mean absolute change} across all agents and traits:
\[
\overline{\Delta}_t
~=~
\frac{1}{M_t}\sum_{(i,k)\in\Omega_t}\bigl|\,x_{i,k}^{(t+1)}-x_{i,k}^{(t)}\,\bigr|,
\qquad
\Omega_t=\{(i,k): x_{i,k}^{(t)} \text{ and } x_{i,k}^{(t+1)} \text{ observed}\},
\]
where $M_t=|\Omega_t|$. (Entries missing at either step are omitted.)

We plot $\overline{\Delta}_t$ against $t{+}1$ (the ``next'' iteration). Small values indicate that, on average,
agents would \emph{barely change} their traits on the next round; sustained lows (plateaus) and local minima of
$\overline{\Delta}_t$ therefore mark \emph{low-motion basins} or \emph{local rest points} in the dynamics.
In our experiments, early dips (e.g., $t{+}1\approx 4\text{–}5$) show transient stabilization, while later,
deeper dips (e.g., $t{+}1\approx 9\text{–}11$) indicate convergence to a more stable configuration.
This is consistent with a behavioral fixed-point interpretation (\emph{no unilateral change} at the profile),
though a payoff-based Nash certificate requires explicit utilities.

\subsection{Learned dynamics}

We analyze a swarm of $N$ agents with $K$ normalized traits. Let $x^{(t)}_{i,k}\in[0,1]$ denote trait $k$ of agent $i$ at iteration $t$. The \emph{observed mean step change} is
\[
\overline{\Delta}_t \;=\; \frac{1}{M_t}\sum_{(i,k)\in\Omega_t}\left|x^{(t+1)}_{i,k}-x^{(t)}_{i,k}\right|,
\qquad 
\Omega_t=\{(i,k): x^{(t)}_{i,k},x^{(t+1)}_{i,k}\ \text{observed}\},
\]
which we plot against $t{+}1$ (missing entries are omitted).

\begin{figure}[h!]
    \centering
    \includegraphics[width=.8\textwidth]{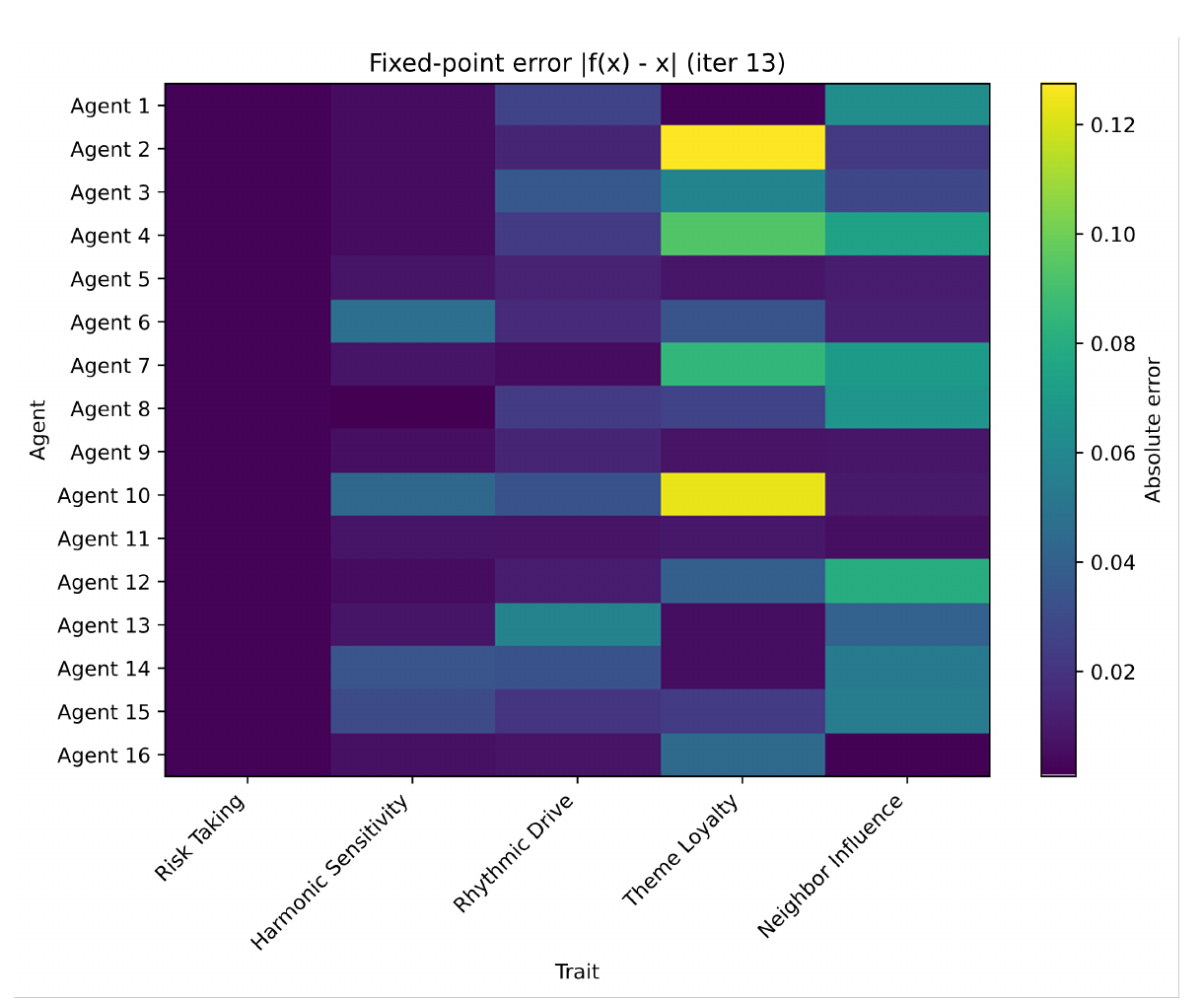}
    \caption{Fixed-point residual heatmap at the terminal iteration.
Each cell shows $|f(x) - x|$ for one agent–trait pair under the learned line-topology
best-response map. The matrix is predominantly dark, indicating agents would make only
minor adjustments if others stayed fixed; two localized spikes in Theme Loyalty
set the global $\varepsilon \approx 0.13$. Together with the calibrated decay of
mean $|\Delta|$, this supports a locally stable behavioral equilibrium.}
    \label{fig:heatmap_trait}
\end{figure}

To obtain a simple \emph{learned best–response} dynamics, for each trait $k$ we pool all agents and all consecutive saved iterations $t\!\to\!t{+}1$ and fit the linear regression
\[
x^{(t+1)}_{i,k} ~=~ \alpha_k\,x^{(t)}_{i,k} \;+\; \beta_k\,\overline{x}^{(t)}_{N(i),k} \;+\; \gamma_k \;+\; \varepsilon,
\]
where the interaction graph is a \textbf{line (path)}: interior agents have $N(i)=\{i{-}1,i{+}1\}$ and the two ends have a single neighbor. Here $\overline{x}^{(t)}_{N(i),k}$ is the average over $N(i)$. Ordinary least squares yields $(\alpha_k,\beta_k,\gamma_k)$ and induces, for each trait, a linear map on the full agent vector
\[
x^{(t+1)}_{:,k} ~=~ M_k\,x^{(t)}_{:,k} + c_k,\qquad 
M_k=\alpha_k I + \beta_k P,\ \ c_k=\gamma_k\mathbf{1},
\]
with $A$ the path (line) adjacency, $D=\mathrm{diag}(A\mathbf{1})$, and $P=D^{-1}A$ the row–normalized adjacency (ends have degree~1, interiors degree~2). Iterating this map from the first observed profile produces a \emph{model–predicted} step–change curve
\[
\overline{\Delta}^{\text{model}}_t \;=\; \frac{1}{NK}\sum_{i,k}\left|\,\bigl(M_k x^{(t)}_{:,k}+c_k\bigr)_i - x^{(t)}_{i,k}\,\right|.
\]
To compare fairly with the empirical process (which includes exploration noise, clipping, and asynchrony), we align the model to data via an affine calibration that estimates a timescale factor $\lambda$ and a noise floor $\delta$:
\[
(\lambda,\delta)\;=\;\arg\min_{\lambda,\delta}\sum_t\!\Bigl(\overline{\Delta}_t - \bigl(\delta + \lambda\,\overline{\Delta}^{\text{model}}_t\bigr)\Bigr)^2,
\]
and we plot the calibrated overlay $\delta+\lambda\,\overline{\Delta}^{\text{model}}_t$.
Empirically, the calibrated model tracks the decay and plateau in $\overline{\Delta}_t$, with $\delta$ capturing the persistent motion floor, supporting practical convergence to a stable behavioral fixed point.
Parameters are listed in Table~\ref{tab:br_fits} and \ref{tab:calibration}. Figure~\ref{fig:heatmap_trait} shows the fixed-point residual heatmap at the terminal iteration.

\begin{table}[h]
\centering
\caption{Per–trait best–response coefficients fitted by pooled OLS on the line topology. $R^2$ is the in–sample coefficient of determination; $n$ is the number of transitions used.}
\label{tab:br_fits}
\begin{tabular}{lrrrrr}
\toprule
Trait & $\alpha$ & $\beta$ & $\gamma$ & $R^2$ & $n$ \\
\midrule
risk\_taking          & 0.436 & 0.104 & 0.411 & 0.751 & 192 \\
harmonic\_sensitivity & 0.719 & 0.173 & 0.102 & 0.688 & 192 \\
rhythmic\_drive       & 0.628 & 0.128 & 0.206 & 0.691 & 192 \\
theme\_loyalty        & 0.573 & 0.161 & 0.129 & 0.359 & 192 \\
neighbor\_influence   & 0.551 & 0.245 & 0.146 & 0.439 & 192 \\
\bottomrule
\end{tabular}
\end{table}

\begin{table}[h]
\centering
\caption{Affine calibration parameters for the model overlay $\delta+\lambda\,\overline{\Delta}^{\text{model}}_t$.}
\label{tab:calibration}
\begin{tabular}{lrr}
\toprule
 & $\lambda$ & $\delta$ \\
\midrule
Calibrated overlay & 0.130 & 0.073 \\
\bottomrule
\end{tabular}
\end{table}

\clearpage
\newpage

\begin{figure}[h!]
    \centering
    \includegraphics[width=1.\textwidth]{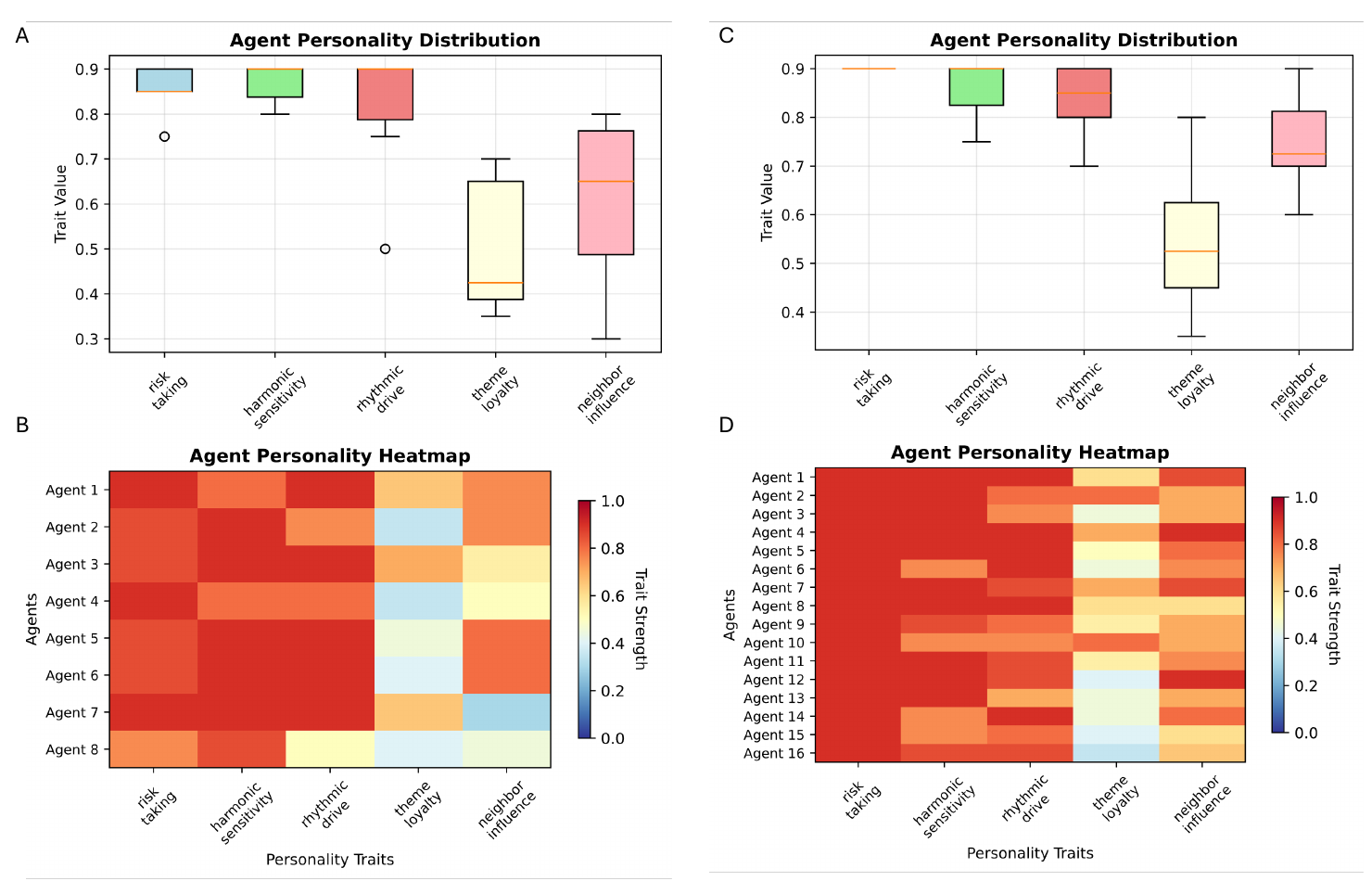}
    \caption{Agent personality landscape at convergence, for Global Objective \#1 (A-B) and Global Objective \#2 (C-D). A, Distribution of five traits across eight agents at the final iteration: risk taking, harmonic sensitivity, rhythmic drive, theme loyalty, and neighbor influence. Boxes show IQR with median line; whiskers denote range; circles mark outliers. Risk taking, harmonic sensitivity, and rhythmic drive are high and tightly clustered; theme loyalty is lower with broader spread; neighbor influence shows the largest variance, indicating heterogeneous coupling strengths. B, Heatmap of per-agent profiles (values normalized to 
$[0,1]$). Most agents share a high-risk/high-harmonic signature; theme loyalty remains modest; coupling varies by agent (e.g., Agent 8 shows reduced rhythmic drive but elevated neighbor influence). Together these statistics indicate confident exploration, stable harmonic modeling, and differentiated coordination roles without a central planner. Panels C and D show the results for Global Objective \#2. The results also imply that the system needs uneven roles to sustain global coherence and novelty. Over‑regularizing personalities would likely harm structure and  heterogeneity is hence an emergent design feature, not a variance to minimize.
We also note that capturing traits in different iterations or at local minima means we could potentially snapshot personalities and environment and re‑instantiate them as a portable style prior—a zero‑weight ``checkpoint'' for fast reuse/transfer.}
    \label{fig:trait_stats}
\end{figure}

\begin{figure}[h!]
    \centering
    \includegraphics[width=.8\textwidth]{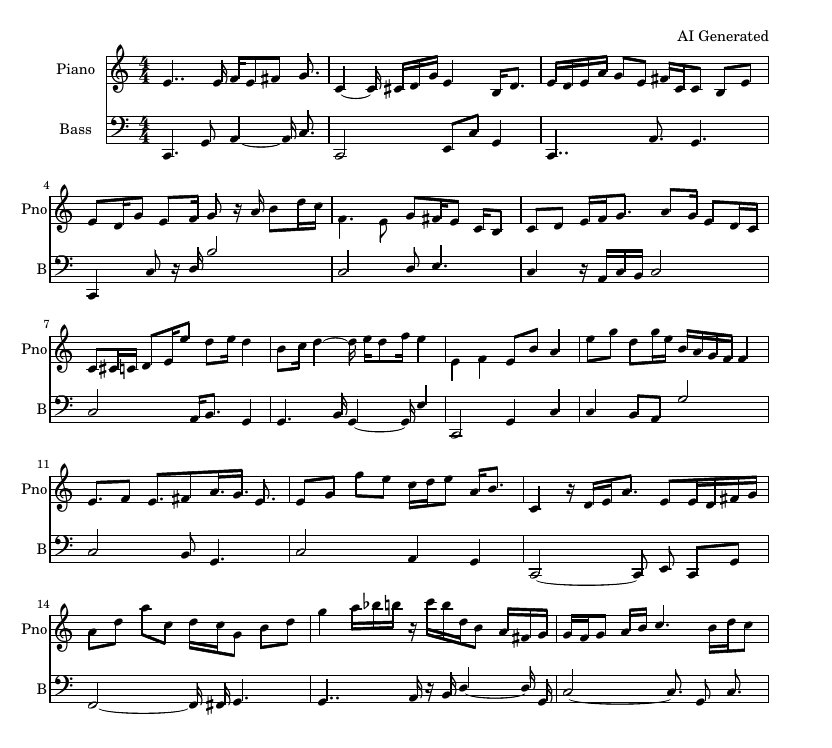}
    \caption{Resulting musical score at the first local maximum-scoring iteration, for the swarm case, for Global objective \#2. While simpler, it still features notable complexity, far exceeding the results for the other methods shown in Figure~\ref{fig:score_example_101}}
    \label{fig:score_example_101}
\end{figure}

\begin{figure}[h!]
    \centering
    \includegraphics[width=.67\textwidth]{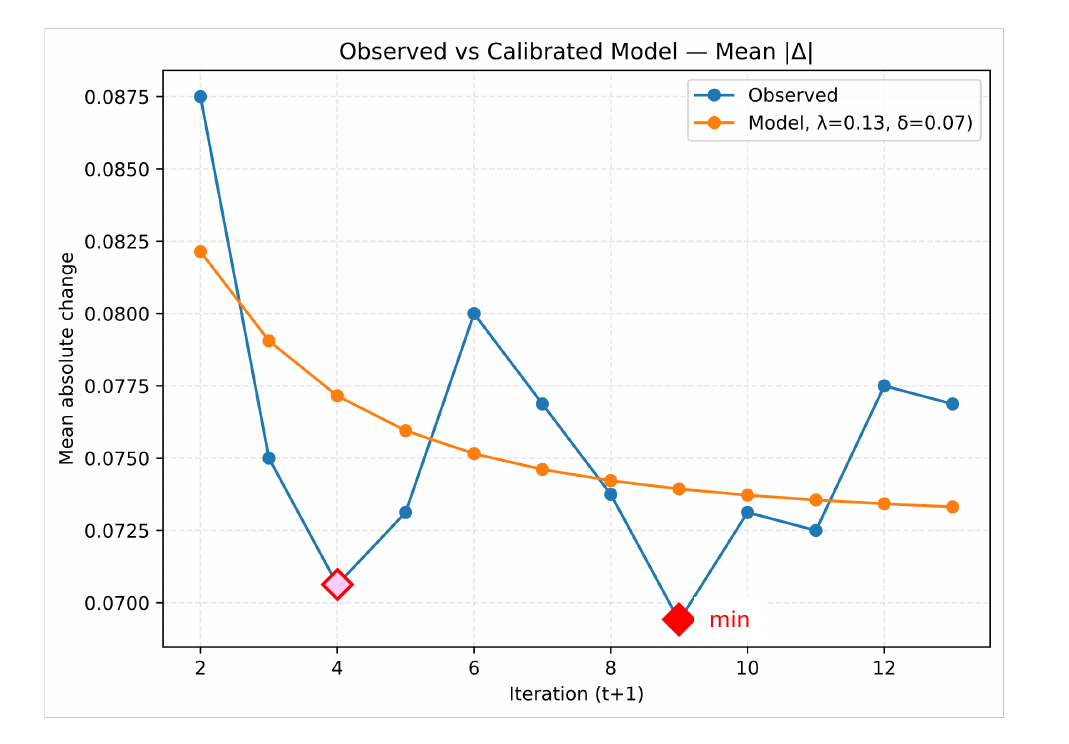}
    \caption{Analysis of the dynamics in the swarm system, for Global Objective \#2. The plot depicts mean step-to-step absolute trait change  $\bar{\Delta}_t=\frac{1}{M_t}\sum_{(i,k)\in\Omega_t}\lvert x^{(t+1)}_{i,k}-x^{(t)}_{i,k}\rvert$
across 16 agents and 5 traits. Lower values indicate less adjustment and thus proximity to a low-motion rest point. The curve shows an early stabilization around iterations $3$–$5$ and a deeper basin with the global minimum at $t{+}1\!\approx\!9$ (red diamond) towards the end, after which motion remains low. The orange curve shows learned–dynamics prediction of the step–to–step mean absolute trait change $\overline{\Delta}_t$ across all agents and traits. The model curve is obtained by iterating the per–trait linear best–response fitted on the line topology and then applying an affine calibration $\delta+\lambda\,\overline{\Delta}^{\mathrm{model}}_t$ (legend reports $\lambda,\delta$). The calibrated overlay captures the empirical decay and low–motion plateau, consistent with a contractive dynamic converging to a stable behavioral fixed point; the residual floor reflects persistent exploration/noise.}
\label{fig:nash_analysis_delta}
\end{figure}

\begin{figure}[h!]
    \centering
    \includegraphics[width=.8\textwidth]{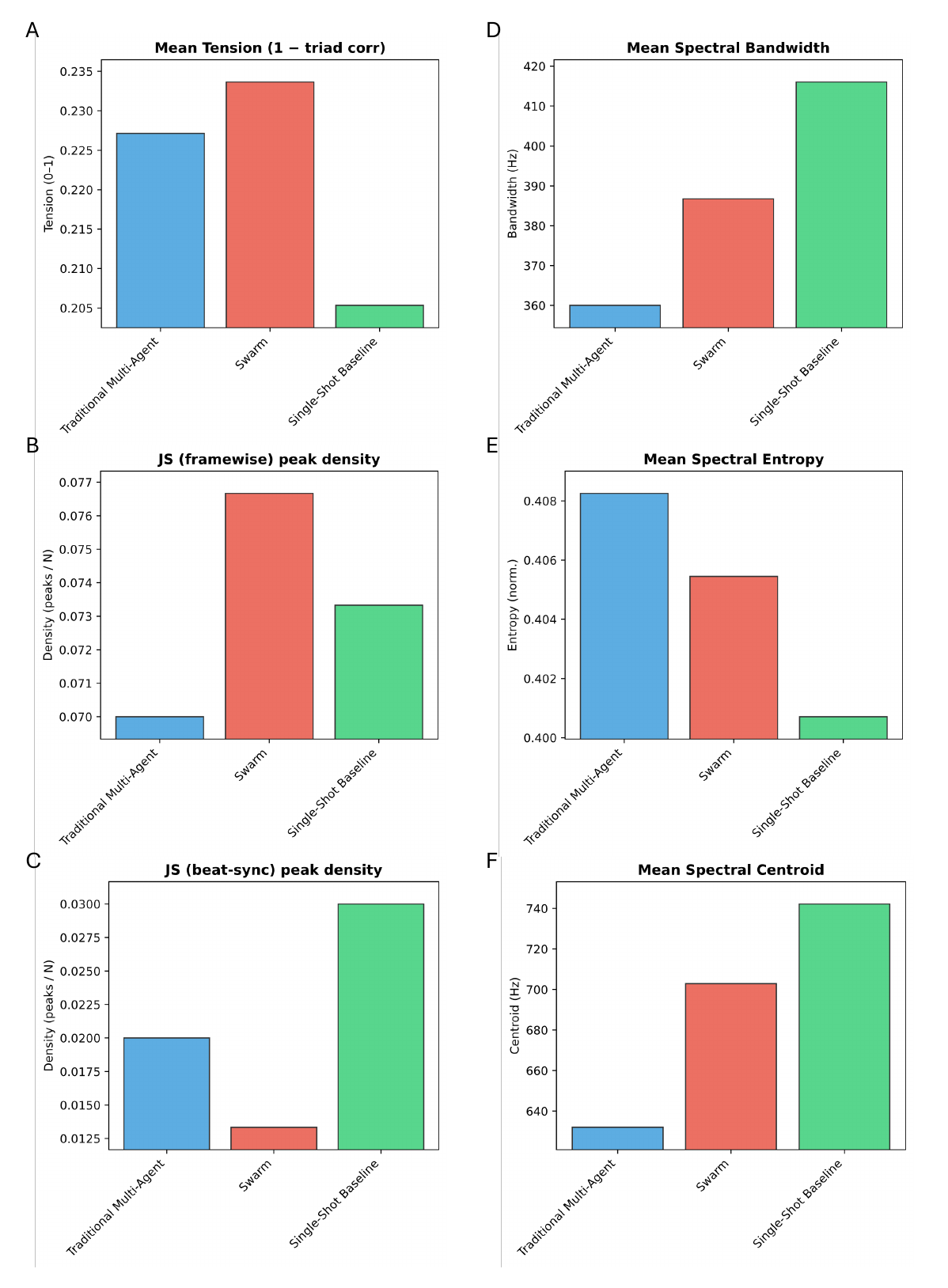}
    \caption{Comparative metrics of swarm, single-shot, and traditional compositions (A-C) and comparative timbre metrics of swarm, single-shot, and traditional compositions (D-F) (see Table~\ref{table:audio_measures} for definitions of metrics using simple terms).
    A: Mean harmonic tension (1 -- triad correlation) shows that the swarm output sustains the highest average tension, indicating greater harmonic instability, while the single-shot baseline remains most consonant. 
    B: Framewise Jensen–Shannon (JS)-novelty peak density reveals that swarm compositions generate the greatest frequency of local harmonic recontextualizations, reflecting frequent micro-level novelty. 
    C: Beat-synchronous JS-novelty peak density highlights that single-shot outputs concentrate more novelty at metrically aligned positions, whereas swarm novelty is less tied to the bar structure, suggesting off-beat or irregular creative leaps.
    D: Mean spectral bandwidth quantifies the spread of frequency energy around the spectral centroid. Single-shot outputs exhibit the broadest bandwidth, swarm results are intermediate, and traditional outputs are narrowest, indicating more compact timbre. 
    E: Mean spectral entropy captures how evenly energy is distributed across frequencies. Traditional compositions show the highest entropy, suggesting a flatter, more diffuse spectral profile, while single-shot outputs are the most ordered; swarm again occupies the middle ground. 
    F: Mean spectral centroid reflects the brightness of the sound. Single-shot outputs are brightest, traditional the darkest, and swarm lies between the two. 
    Together, these timbre measures show that swarm compositions balance spectral richness and brightness between the extremes of single-shot and traditional baselines.\textbf{}
    }
    \label{fig:audio_creative_tension}
\end{figure}

\begin{figure}[h!]
    \centering
    \includegraphics[width=0.55\textwidth]{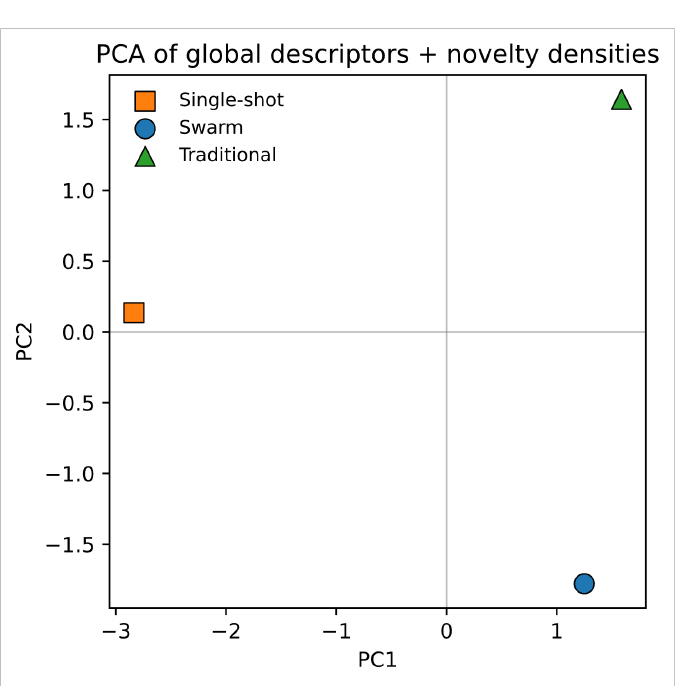}
    \caption{
    Principal component analysis (PCA) of global audio descriptors and novelty densities across swarm, single-shot, and traditional compositions. 
    The three approaches occupy clearly separated regions in the reduced feature space, indicating that their timbral and harmonic signatures are systematically distinct. 
    The single-shot output clusters on the negative side of PC1, reflecting its narrow spectral profile and lower tension; 
    the traditional output projects strongly along PC2, consistent with its higher entropy and broader timbral diffusion; 
    and the swarm output lies apart along positive PC1 and negative PC2, capturing its elevated harmonic tension and frequent novelty events. 
    Together, this separation highlights that swarm-generated music embodies a distinct balance of harmonic and timbral properties not reducible to either baseline.
    }
    \label{fig:pca_audio}
\end{figure}

\begin{figure}[h!]
    \centering
    \includegraphics[width=0.6\textwidth]{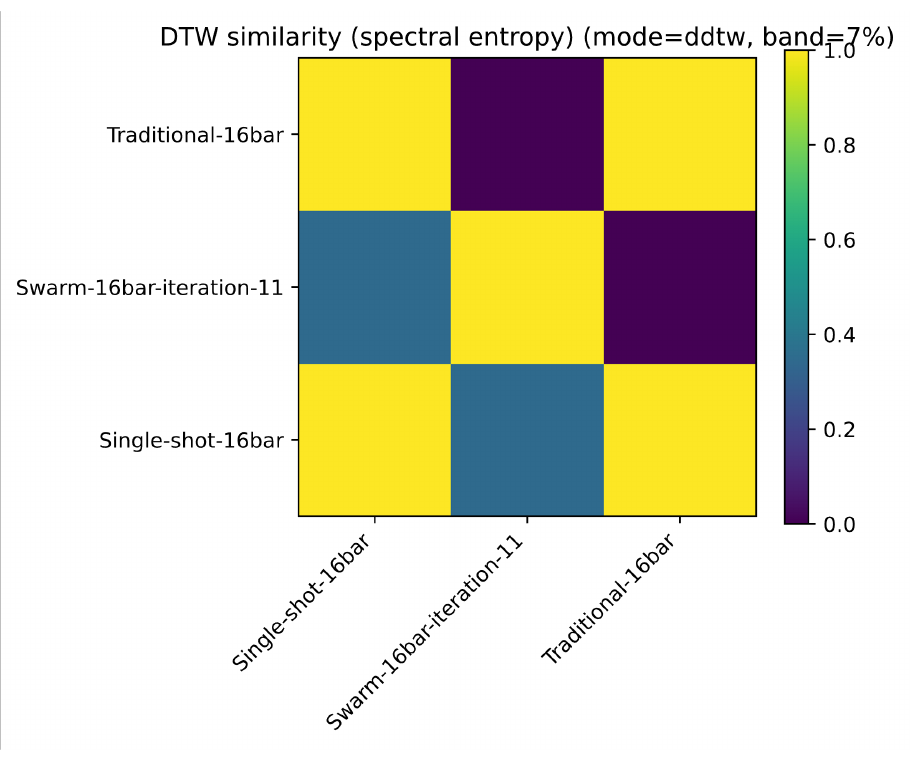}
    \caption{
    Dynamic time warping (DTW) similarity of spectral entropy trajectories across swarm, single-shot, and traditional compositions. 
    Spectral entropy reflects how ordered versus diffuse the spectral profile is, with lower values indicating tonal focus and higher values indicating noisy or flattened spectra. 
    The similarity matrix shows that single-shot and traditional outputs share highly similar entropy trajectories (bright yellow, high similarity), 
    whereas the swarm output diverges from both baselines (darker off-diagonal entries), following a distinct temporal evolution of spectral complexity. 
    This separation highlights that swarm-generated music does not merely interpolate between baselines in timbral entropy but instead develops a unique trajectory of ordered–disordered balance over time.
    We also conclude that the swarm naturally produces cross‑metric phrasing, a hallmark of expressive timing/syncopation, but without an explicit performance model.
    }
    \label{fig:dtw_entropy}
\end{figure}

\clearpage
\newpage

\section{Overcoming Closure through Interaction: Gödelian Limits and Swarm Emergence}

Table~\ref{table:godel_extension} shows examples from the literature where many identical closed systems, through interaction, transcended the limitations of any one system. These underscore the observations made in the swarm approach presented in this paper. 
 
\begin{table}[h!]
\centering
\caption{Examples where many identical closed systems, through interaction, transcend the limitations of any one system.}
\label{table:godel_extension}
\begin{tabular}{p{3cm} p{3.8cm} p{3.8cm} p{3.8cm}}
\hline
\textbf{Domain} & \textbf{Closed System} & \textbf{Mechanism of Transcendence} & \textbf{Outcome} \\
\hline
Logic & Peano Arithmetic (PA) or similar & Iterated reflection \cite{feferman1962}, ordinal logics \cite{turing1939} & Strictly stronger progressions of theories \\
Complexity Theory & Polynomial-time verifier, identical provers & Interactive proofs (IP=PSPACE) \cite{shamir1992}, multi-prover (MIP, MIP*) \cite{babai1991,ji2021} & Verification power leaps from P to PSPACE, NEXP, and RE \\
Automated Reasoning & Identical decision procedures & Nelson–Oppen combination \cite{nelsonoppen1979} & Completeness for theory unions unreachable by any single solver \\
Biology & Genetically identical cells & Differentiation via signaling, feedback \cite{gilbert2010developmental} & Emergence of multicellular intelligence and specialized tissues \\
Immunology & Clonal immune cells & Environmental cues, interaction with antigens \cite{murphy2012immunobiology} & Specialization into effector and memory lineages \\
Biological Materials & Protein Building Blocks, e.g. Alpha-helices, beta-sheets & Hierarchical structures \cite{Giesa2011ReoccurringAnalogies,Ackbarow2008HierarchicalMaterials,Cranford2012Biomateriomics}
 & Specialization into distinct material properties \\
Human Collectives & Individuals with similar IQ & Group interaction, turn-taking, diversity of participation \cite{woolley2010collective} & Emergent collective intelligence factor $c$ \\
\hline
\end{tabular}
\end{table}

As another note, one might object that a very large language model is not in fact a ``closed" system'', since it exhibits stochasticity, vast parametric complexity, and can be conditioned by diverse prompts. However, in the Gödelian sense of closure, what matters is not surface variability but the existence of a fixed, finite formal calculus. Once the weights are frozen, the model is a deterministic function from inputs and random seeds to outputs. Stochastic sampling merely explores distributions already encoded; prompting conditions the search but does not extend the underlying proof power. In this respect, a single LLM, however large, is akin to a closed formal system: powerful, but bounded, and inevitably incomplete. By contrast, our swarm framework operationalizes a mechanism for moving between such closed instances: outputs from one agent become new inputs or "axioms" for others, roles evolve under peer feedback, and collective memory reshapes the effective rule set. This continual re-organization transforms a collection of closed models into an open-ended meta-system, echoing Gödel’s insight that closure can only be transcended by extension.

\clearpage

\newpage

\begin{tcolorbox}[
  colback=gray!2,
  colframe=black!20,
  boxrule=0.5pt,
  enhanced, breakable,
  title={Text Box~S1: Swarm with central control, objectives of agent 1 over 6 iterations.}
]
\begin{minted}[fontsize=\footnotesize, breaklines, linenos, numbersep=5pt]{json}
"past_objectives": [
  "Establish A harmonic minor and introduce the motif's opening fragment (A4–C5–E5) in the upper voice with a lyrical, expressive rhythm while the lower voice provides steady rhythmic grounding and harmonic support (A/E and hinted G#) to set up a smooth transition into bar 2.",
  "Establish A harmonic minor by presenting the A4–C5–E5 motif lyrically with a varied, expressive rhythm (not a static arpeggiation), keep the bass in a low register (A2/E2) and use stepwise voice-leading toward G# to create forward momentum into bar 2 while preserving clear two-voice spacing.",
  "Present the A4–C5–E5 motif lyrically with varied, expressive rhythm and clear two-voice spacing (bass on A2/E2), use mostly stepwise/contrary motion to hint toward G# for forward momentum into bar 2, avoid extreme leaps or ambiguous bass notes (e.g., C3), and shape the phrase so the motif is easily transformable later.",
  "Present the A4–C5–E5 motif as a smooth, lyrical, mostly stepwise line with varied, legato rhythms while the bass clearly outlines the i-harmony with a steady quarter-note pulse (A2 with supportive E2), keep clear octave-plus spacing, avoid static repeated bass tones, and shape the bar’s closing note(s) to voice-lead naturally toward a G# entry in bar 2 so the motif is easily transformable.",
  "Present the A4–C5–E5 motif as a smooth, lyrical, mostly stepwise line while adding a small expressive twist (a passing F as a minor-sixth neighbor or a gentle embellishment), give the bass more independence with slight rhythmic variation (avoid static repeated tonic and long octave parallels), and shape the bar’s closing voices so the bass/inner harmony moves clearly toward G# to prepare Bar 2.",
  "Open with a smooth, lyrical statement of the A4–C5–E5 motif (with a gentle F passing neighbor) while the bass establishes a clear tonic foundation (sustained A2 with subtle rhythmic variation rather than large leaps) and moves purposefully toward G#2 at the bar’s close to prepare Bar 2’s dominant."
]
\end{minted}
\end{tcolorbox}

\newpage
\begin{tcolorbox}[
  colback=gray!2,
  colframe=black!20,
  boxrule=0.5pt,
  enhanced, breakable,
  title={Text Box~S2: Swarm with central control, objectives of agent 5 over 6 iterations.}
]
\begin{minted}[fontsize=\footnotesize, breaklines, linenos, numbersep=5pt]{json}
"past_objectives": [
      "In Bar 5, intensify development of the A4–C5–E5–G#5–A5 motif in the upper voice with clear voice-leading from Bar 4 while the lower voice provides steady rhythmic grounding and strengthened A harmonic minor harmony (with a tasteful syncopation) to propel momentum into Bar 6.",
      "Develop the motif more contrapuntally in the upper voice with a varied contour/transposition (e.g., a brief stepwise expansion of A4–C5–E5–G#5–A5) while lowering the competing bass C4 to C3 and reworking the left hand into a stepwise A2–G#2–A2 (i–V6/5–i) with subtle syncopation to avoid masking the melody and to propel momentum into Bar 6.",
      "Develop the motif contrapuntally in the RH via a brief sequenced/inverted stepwise variant (avoiding extreme leaps), while removing or recontextualizing the isolated C3 (make it a passing tone or replace with A2/E2) and shaping the LH into a subtle syncopated A2–G#2–A2 (i–V6/5–i) to clarify harmony, preserve voice independence, and propel momentum into Bar 6/7 to prepare the dominant.",
      "In Bar 5, present a brief sequenced/inverted stepwise variant of the A–C–E–G#–A motif in the right hand (no large leaps) while simplifying the left hand to a clear syncopated A2–G#2–A2 pulse (i–V6/5–i), convert the isolated C3 into a passing tone or replace it with A2/E2, and increase spacing/voice independence to smoothly propel into the dominant preparation for bars 6–7.",
      "In Bar 5, present a stepwise sequenced/inverted variant of the A–C–E–G#–A motif in the right hand with increased spacing and an F as a passing/neighbor tone for harmonic-minor color, while the left hand adopts a varied syncopated A2–(rest)–G#2–A2 pulse (replacing the isolated C3 with A2/E2 or a passing tone), use tied/redistributed rhythms to smooth the 0.75+0.25 placement, avoid prolonged octave/unison parallels through contrary motion, and shape the voice-leading to clearly prepare the dominant for bars 6–7.",
      "In Bar 5, sequence/invert the A–C–E–G#–A motif in the right hand with increased spacing and an F-natural neighbor/pass tone to emphasize the harmonic-minor color, while the left hand provides a clearer preparatory bass line (A2 pulse moving toward E2, optionally via D2) using syncopation and tied rhythms to smooth the 0.75+0.25 placement, avoid parallel octaves/unisons, and shape voice-leading to firmly prepare the dominant for bars 6–7."
    ]
\end{minted}
\end{tcolorbox}

\clearpage

\begin{tcolorbox}[
  colback=gray!2,
  colframe=black!20,
  boxrule=0.5pt,
  enhanced, breakable,
  title={Text Box~S3: Swarm with central control, objectives of agent 8 over 6 iterations.}
]
\begin{minted}[fontsize=\footnotesize, breaklines, linenos, numbersep=5pt]{json}
"past_objectives": [
      "Provide a clear, final cadence in A harmonic minor that resolves the harmonic tension (E → A), restates or alludes to the A4–C5–E5–G#5–A5 motif in the upper voice for lyrical closure, and lets the lower voice supply rhythmic grounding and harmonic support to create a definitive, stylistically cohesive ending.",
      "Deliver a decisive V–i cadence by placing/sustaining A in the bass on the downbeat (A2), resolve the upper G#5→A5 while briefly alluding to the A–C–E–G#–A motif, and finish on a supported closed A–C–E sonority with clear rhythmic grounding.",
      "Deliver a conclusive V–i close by shaping bar 8 as an E major (V) → A minor (i) span: support the downbeat with sustained A2 for rhythmic grounding, present E/G# in the first half so the upper G# resolves to A, briefly allude to the A–C–E–G#–A motif in the upper voice, and finish on a clearly voiced A–C–E tonic sonority.",
      "Deliver a conclusive i cadence by sustaining A2 for rhythmic grounding, resolving the prior G# to A in the upper voice with stepwise, lyrical motion while briefly alluding to the A–C–E–G#–A motif, and finish on a clearly voiced A–C–E tonic sonority.",
      "Deliver a decisive i cadence by resolving the upper G#→A in a stepwise, lyrical line that briefly alludes to the A–C–E–G#–A motif, sustain A2 for grounding but introduce a subtle bass rhythmic variation (anticipation or short rest) to avoid monotony, use contrary motion to prevent prolonged parallel octaves, and close with a clearly voiced A–C–E tonic sonority (e.g., arpeggiated upper tones implying C5 and E5 before final A5).",
      "Deliver a decisive i cadence: place a grounded A2 on the downbeat (with a subtle anticipation or short rest for rhythmic interest), have the upper voice resolve G#→A stepwise and arpeggiate C5–E5 before a final A5 to confirm a clear A–C–E tonic sonority, use contrary motion to avoid parallel octaves, and briefly allude to the A–C–E–G#–A motif to close the arc."
    ]
\end{minted}
\end{tcolorbox}

\clearpage

\begin{figure}[h!]
    \centering
    \includegraphics[width=0.55\textwidth]{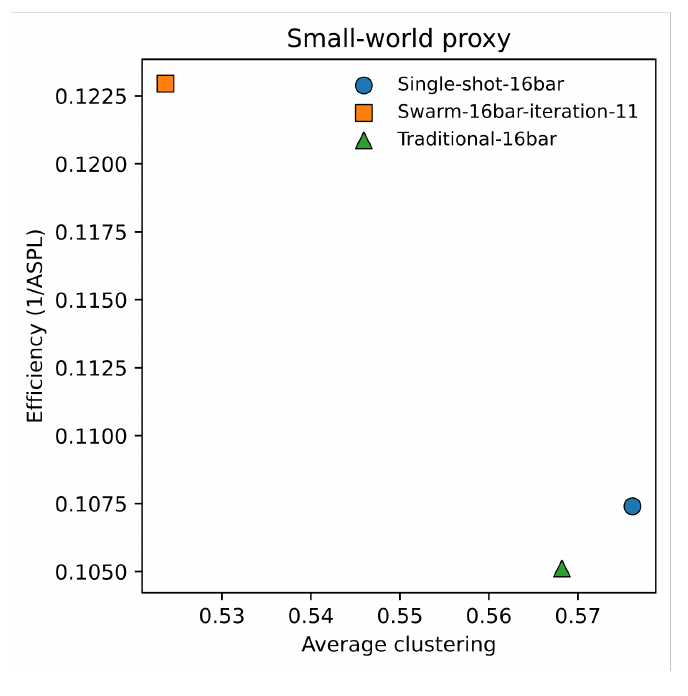}
    \caption{
    Small-world proxy plot showing the relationship between average clustering and efficiency (inverse of average shortest path length, 1/ASPL) for self-similarity graphs derived from swarm, single-shot, and traditional compositions. 
    All graphs lie in the small-world regime, combining above-random clustering with relatively short paths. 
    The swarm graph occupies the upper-left position, reflecting the shortest path lengths and thus the highest efficiency while maintaining substantial clustering, indicating globally efficient yet locally coherent organization.
    }
    \label{fig:graph_smallworld_proxy}
\end{figure}

\clearpage

\section{Deeper analysis of structural organization of the generated music}
\label{SI:deeper_structure}

\begin{figure}[t]
  \centering
  \includegraphics[width=\textwidth]{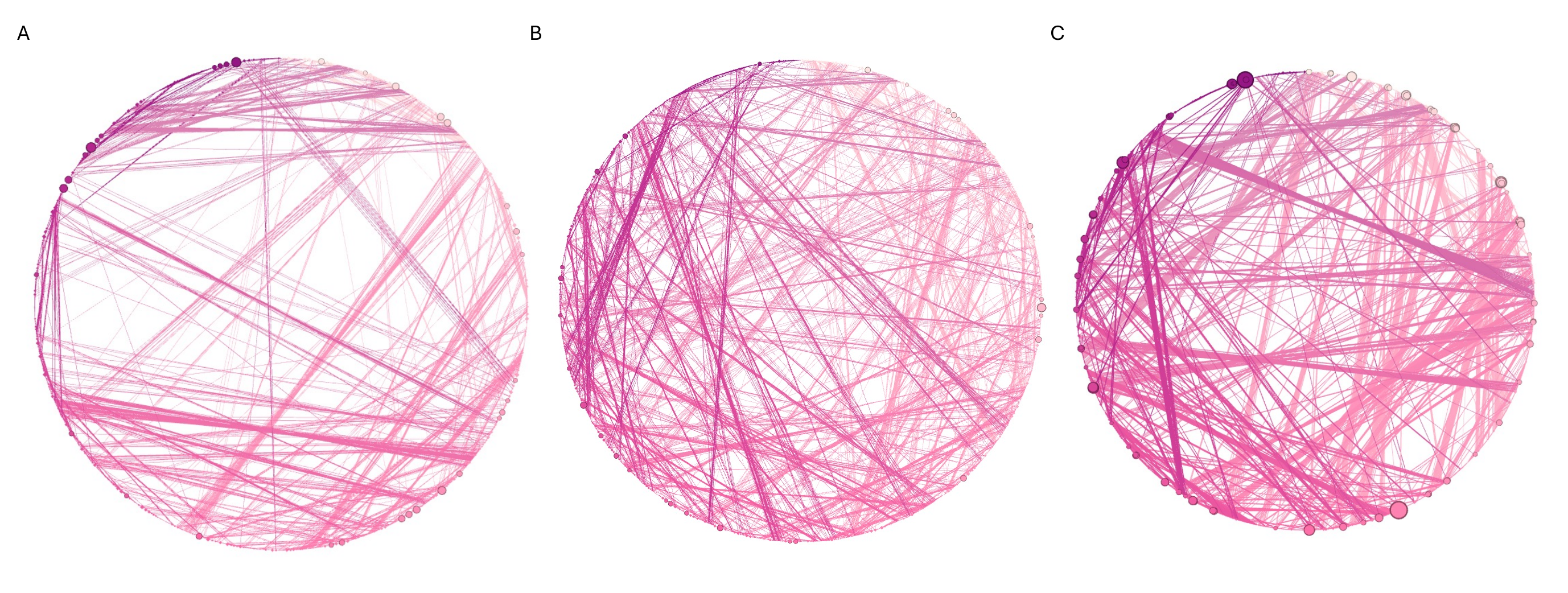}
  \caption{\textbf{Long-range musical self-similarity across generation paradigms.}
  Circular graph layouts for \textbf{(a)} traditional multi-agent, \textbf{(b)} swarm (decentralized, pheromone-mediated), and \textbf{(c)} single-shot baselines. 
  Nodes correspond to score positions and are arranged counterclockwise in time. 
  Undirected edges connect pairs with high musical self-similarity; edge opacity/width reflects similarity strength. 
  Node size encodes betweenness centrality. 
  The traditional system forms fragmented long-range structure with several local bottlenecks; the swarm shows distributed long-range links without dominant hubs, indicating balanced global coherence and role sharing; the single-shot output concentrates flow through a few hubs, yielding strong but motif-repetitive global links. 
  Together, these visualizations highlight that the swarm paradigm achieves broad long-range coherence while preserving variety, consistent with higher global efficiency and lower centrality inequality (see Methods/SI).}
  \label{fig:ssm_circle}
\end{figure}

\begin{figure}[t]
  \centering
  \includegraphics[width=0.8\textwidth]{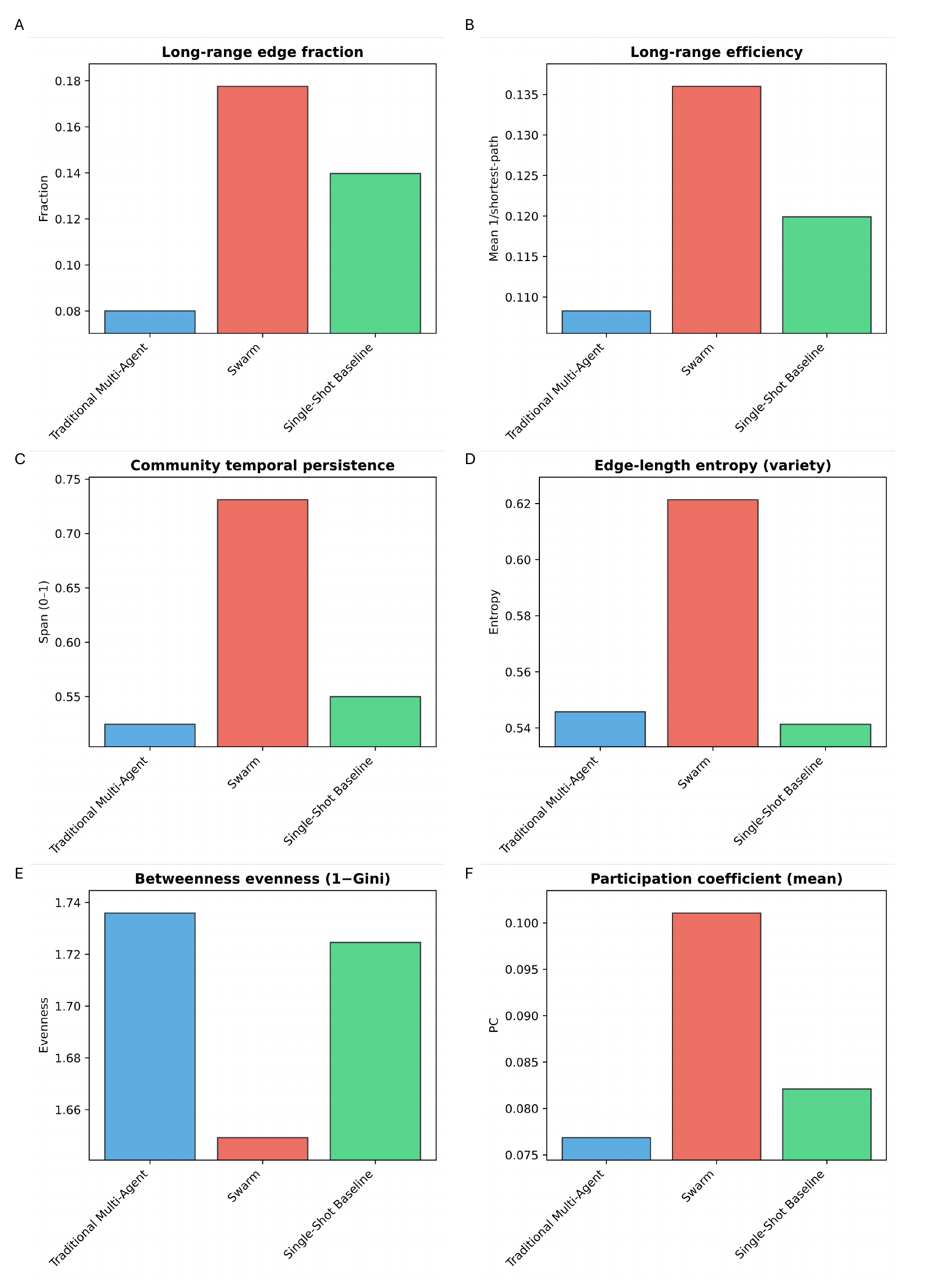}
  \caption{
  Long-range coherence and variety metrics across systems (see Table~\ref{table:simple_def} for definitions of metrics using simple terms).
  Bars compare Traditional Multi-Agent, Swarm, and Single-Shot Baseline systems using self-similarity graphs. 
  A:Long-range edge fraction: fraction of edges connecting events separated by large temporal lags. Swarm is highest, showing it more frequently links distant sections. 
  B:Long-range efficiency: mean inverse shortest-path length among distant pairs. Swarm is again highest, indicating remote sections are more efficiently bridged. 
  C: Community temporal persistence: fraction of the piece spanned by detected communities. Swarm yields the broadest spans, reflecting themes that persist across time. 
  D: Edge-length entropy: normalized entropy of edge-lag distribution. Swarm shows the greatest entropy, meaning a richer mix of short-, medium-, and long-range recurrences. 
  E: Betweenness evenness (1–Gini): evenness of centrality across nodes. Swarm is lower, suggesting specialized motifs act as structural bridges. 
  F: Participation coefficient (mean): average extent to which nodes connect across communities. Swarm is highest, showing stronger integration between themes. 
  Overall, the swarm paradigm produces more coherent long-range connectivity, more diverse recurrence structure, and greater thematic integration than either baseline.
  }
  \label{fig:longrange_variety}
\end{figure}

To probe the structural organization of the generated music, we evaluated six graph-theoretic quantities derived from self-similarity networks of each system (Fig.~\ref{fig:longrange_variety}). These metrics capture two overarching properties: \emph{long-range coherence}, the ability to sustain thematic connections across time, and \emph{variety}, the diversity of recurrence structures. Alternative visualizations are shown in Fig.~\ref{fig:ssm_circle}.

The swarm system consistently outperforms both baselines on measures of long-range coherence. Its \textbf{long-range edge fraction} (Fig.~\ref{fig:longrange_variety}A) is highest, showing that distant events are connected more frequently. Its \textbf{long-range efficiency} (Fig.~\ref{fig:longrange_variety}B) is also maximal, indicating that these distant links are not scattered but organized through efficient bridging motifs. Moreover, \textbf{community temporal persistence} (Fig.~\ref{fig:longrange_variety}C) is greatest for the swarm, revealing that musical communities span broader stretches of time and that thematic material recurs with greater durability. Together these three metrics demonstrate that swarm compositions integrate remote sections into a coherent whole, achieving narrative continuity that the multi-agent and single-shot approaches cannot sustain.

On the dimension of variety, the swarm again dominates. Its \textbf{edge-length entropy} (Fig.~\ref{fig:longrange_variety}D) is highest, meaning that it reuses material at a wide spectrum of timescales rather than concentrating recurrences at a narrow band. This produces more layered structural patterns and richer temporal interplay. The \textbf{participation coefficient} (Fig.~\ref{fig:longrange_variety}F) is also highest, showing that motifs connect across multiple communities and thereby weave different sections together. In contrast, baselines tend to produce more insular communities with limited cross-talk.

An interesting exception emerges in \textbf{betweenness evenness} (Fig.~\ref{fig:longrange_variety}E). The swarm shows slightly lower evenness, meaning centrality is not distributed uniformly but concentrated in certain motifs that act as structural bridges. Rather than a weakness, this reflects emergent specialization: a small set of passages assume connective roles, analogous to cadential or transitional motifs in human-composed music. Such specialization highlights how collective dynamics lead to division of labor among motifs, generating the scaffolding that supports long-range coherence.

In synthesis, Fig.~\ref{fig:longrange_variety} shows that swarm-based composition simultaneously increases long-range connectivity, persistence, and integrative variety, while allowing specialized motifs to emerge as structural anchors. This combination of global coherence, temporal richness, and role differentiation points to a qualitatively different mode of musical organization, moving beyond the local repetition of the multi-agent baseline and the hub-dominated structure of the single-shot baseline.

\subsection{Comparing traditional vs. swarm vs. single-shot}
 
Together with the earlier analyses (Fig.~\ref{fig:audio_creative_tension}) these complementary views sharpen the contrasts between swarm, multi-agent, and single-shot paradigms. The swarm framework emerges as the most exploratory and integrated system across both domains. Symbolic graph analysis showed that swarm outputs sustain stronger long-range coherence and thematic persistence, combine recurrence at diverse time scales, and weave motifs across sections, while allowing certain motifs to specialize as structural bridges (Fig.~\ref{fig:longrange_variety}). Audio analyses reinforce this picture: swarm maintains the highest average harmonic tension (panel A), introduces the greatest density of framewise novelty events (panel B), and produces irregular, off-meter creative leaps (panel C), resulting in fluid recombination of motifs that resists block-like segmentation. Timbre measures further confirm that swarm balances brightness and spread between extremes (panels D–F), yielding sound that is neither overly compact nor excessively diffuse.

The traditional multi-agent baseline, by contrast, occupies an intermediate but less compelling regime. Graph metrics revealed fragmented long-range structure and limited variety, with relatively siloed communities and weak long-range links. In audio space, multi-agent compositions sustain moderate tension (panel A) and display some bar-synchronous novelty (panel C), but generate the fewest framewise surprises (panel B). Timbrally, they are the darkest and most compact (panels D,F) yet also the most spectrally diffuse (panel E), producing music that is more sectionalized, safer, and less articulated than swarm outputs. The single-shot baseline remains consistently conservative: highly consonant and metrically predictable (panels A–C), brighter and broader in timbre (panels D,F), but structurally repetitive and hub-dominated in graph analyses.

Taken together, the combined symbolic and audio analyses demonstrate that swarm-generated music differs fundamentally from the other cases. It sustains global coherence through distributed linking of distant sections, introduces frequent micro-level novelty unconstrained by barlines, and balances harmonic instability with timbral richness. Multi-agent and single-shot systems each capture only fragments of this behavior: the former produces darker but fragmented sectional forms, while the latter yields bright but repetitive block structures. The swarm’s emergent dynamics instead achieve the ensemble-like balance of coherence and surprise that characterizes compelling musical organization.

\subsection{Computation of structural properties of generated music}

Long-range edge fraction (LR-EF) quantifies the proportion of edge weight linking distant events. Using circular lag $\Delta t(i,j)=\min(|i-j|,N-|i-j|)$ and threshold $\tau=N/4$, we define 
\[
\mathrm{LR\!-\!EF} = \frac{\sum_{(i,j)\in E} w_{ij}\,\mathbf{1}[\Delta t(i,j)>\tau]}{\sum_{(i,j)\in E} w_{ij}}.
\]

Long-range efficiency (LR-Eff) measures how directly distant sections are bridged. With edge lengths $\ell_{ij}=1/w_{ij}$ and shortest-path distances $d(i,j)$, 
\[
\mathrm{LR\!-\!Eff} = \frac{1}{|\mathcal{P}_\tau|}\sum_{(i,j)\in \mathcal{P}_\tau} \frac{1}{d(i,j)}, \quad \mathcal{P}_\tau=\{(i,j): \Delta t(i,j)>\tau\}.
\]

Community temporal persistence (CPT) captures the temporal extent of communities detected via greedy modularity. For each community $C$, the circular span is $\mathrm{span}(C)=1-\operatorname{maxgap}(C)/N$, where $\operatorname{maxgap}(C)$ is the largest gap between consecutive nodes on the timeline. CPT is the size-weighted mean: 
\[
\mathrm{CPT} = \frac{1}{N}\sum_{C} |C| \,\mathrm{span}(C).
\]

Edge-length entropy (ELE) measures the diversity of recurrence scales. With lag distribution $p(\ell)$ (weighted by $w_{ij}$), 
\[
\mathrm{ELE} = - \frac{1}{\log L} \sum_{\ell=1}^{L} p(\ell)\,\log p(\ell), \quad L=\lfloor N/2 \rfloor.
\]

Betweenness evenness evaluates how uniformly bridging roles are distributed. From betweenness values $\{b_i\}$ we compute the Gini index $G(b)$, and define 
\[
\mathrm{Evenness} = 1 - G(b).
\]

Participation coefficient (PC) quantifies cross-community integration. For each node $i$, with total degree $k_i$ and degree $k_{i,c}$ toward community $c$, 
\[
P_i = 1 - \sum_{c} \left(\frac{k_{i,c}}{k_i}\right)^2, \quad \mathrm{PC}=\frac{1}{N}\sum_i P_i.
\]

\begin{table}[t]
\centering
\caption{Simple interpretation of six structural metrics (Fig.~\ref{fig:longrange_variety}).}
\label{table:simple_def}
\begin{tabular}{p{5cm} p{5cm} p{6cm}}
\toprule
\textbf{Plot} & \textbf{What it measures} & \textbf{Plain-language interpretation} \\
\midrule
A. Long-range edge fraction & Frequency of links between far-apart moments & High values mean the piece frequently recalls or reconnects ideas from much earlier, not just the immediate past. \\
B. Long-range efficiency & Efficiency of connections across distant sections & High values indicate clear “bridges” that link remote sections without detours, yielding coherent long-horizon flow. \\
C. Community temporal persistence & Duration/extent of themes across time & High values mean themes or sections persist and return over broad spans, producing recognizable large-scale form. \\
D. Edge-length entropy & Diversity of recurrence timescales & High values indicate a balanced mix of short-, medium-, and long-lag recurrences, enriching structural variety. \\
E. Betweenness evenness & Uniformity of bridging roles & High values mean many events share connective roles; lower values imply a few specialized passages act as structural bridges. \\
F. Participation coefficient & Cross-community connectivity & High values indicate ideas flow between sections rather than staying siloed, weaving the piece into an integrated whole. \\
\bottomrule
\end{tabular}
\end{table}

\begin{table}[t]
\centering
\caption{Simple interpretation of audio-based measures from Figs.~\ref{fig:audio_creative_tension} and \ref{fig:graph_metrics}.}
\label{table:audio_measures}
\begin{tabular}{p{5cm} p{5cm} p{6cm}}
\toprule
\textbf{Plot} & \textbf{What it measures} & \textbf{Plain-language interpretation} \\
\midrule
Fig.~\ref{fig:audio_creative_tension}A. Mean harmonic tension & Average distance from consonant triads & High values mean the music sustains more dissonance and instability, producing greater tension and color. \\
Fig.~\ref{fig:audio_creative_tension}B. Framewise JS-novelty density & Frequency of local harmonic recontextualizations & High values mean frequent micro-level harmonic shifts, creating local surprises and richer textures. \\
Fig.~\ref{fig:audio_creative_tension}C. Beat-synchronous JS-novelty density & Novelty aligned to the meter & High values mean harmonic changes align with barlines; lower values mean surprises occur off-beat, yielding irregular leaps. \\
Fig.~\ref{fig:audio_creative_tension}D. Spectral bandwidth & Spread of frequency energy around the spectral centroid & High values mean a broad, diffuse timbre; low values mean a compact, focused sound. \\
Fig.~\ref{fig:audio_creative_tension}E. Spectral entropy & Evenness of energy distribution across frequencies & High values indicate noisy, flat timbres; low values indicate tonal clarity and ordered spectra. \\
Fig.~\ref{fig:audio_creative_tension}F. Spectral centroid & Brightness of the timbre & Higher values mean brighter, sharper sound; lower values mean darker, warmer tone. \\
\midrule
Fig.~\ref{fig:graph_metrics}A. Small-worldness index & Balance of clustering and efficiency in audio-derived similarity graphs & High values mean music is both locally cohesive and globally well connected, yielding coherent long-range flow. \\
Fig.~\ref{fig:graph_metrics}B. Graph modularity & Degree of partitioning into communities & High values mean the piece divides into separate sections; lower values mean more integrated structure. \\
Fig.~\ref{fig:graph_metrics}C. Number of detected communities & Structural fragmentation of the similarity graph & More communities mean sectionalized form; fewer mean more unified development. \\
Fig.~\ref{fig:graph_metrics}D. Average clustering coefficient & Local cohesiveness of motifs or sections & High values mean motifs strongly repeat within sections; lower values indicate looser local grouping. \\
\bottomrule
\end{tabular}
\end{table}

\clearpage

\begin{figure}[ht]
    \centering
    \includegraphics[width=1\textwidth]{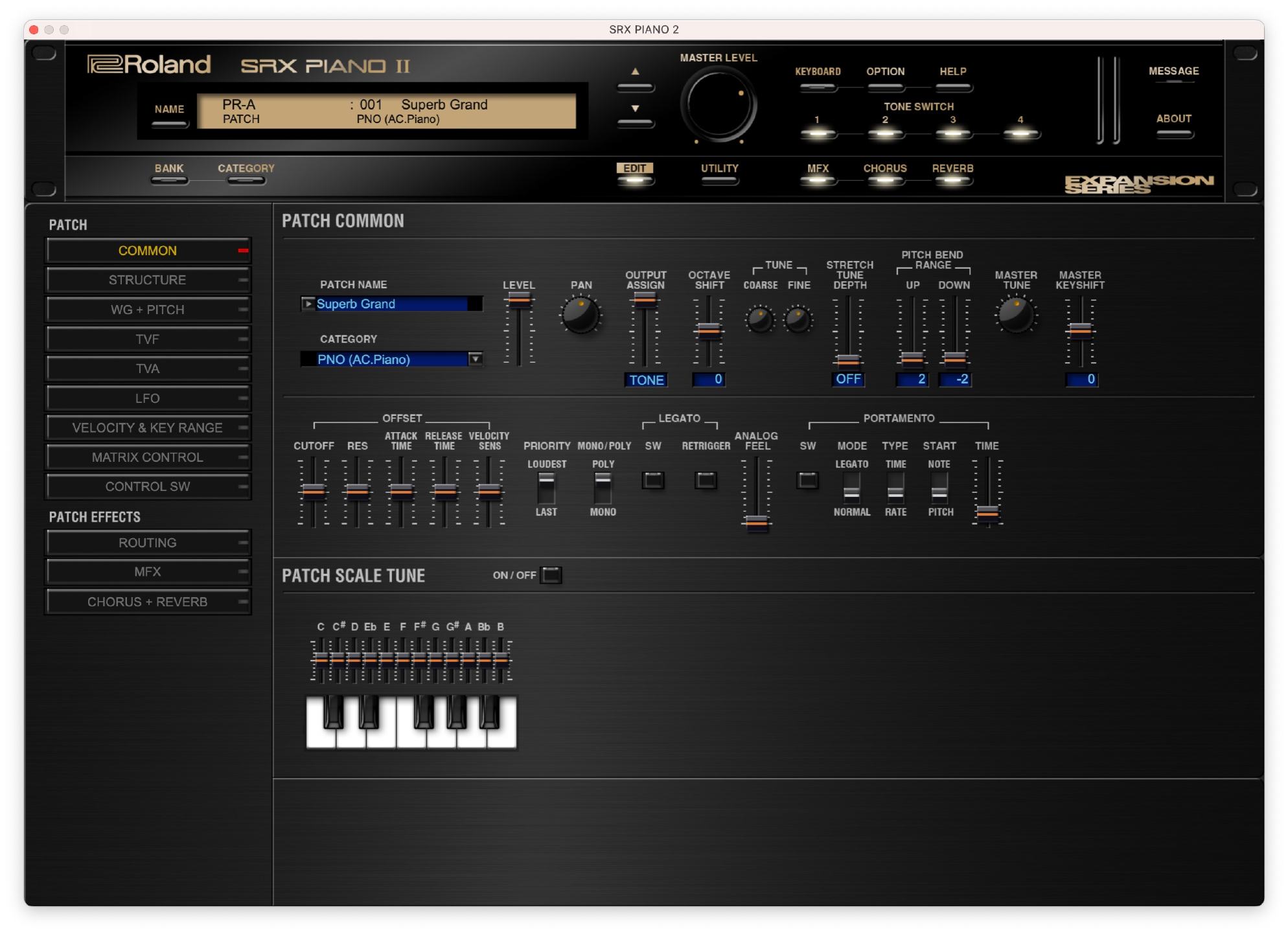}
    \caption{Graphical user interface of the Roland SRX Piano II software synthesizer, showing the Superb Grand patch used in this work. The interface enables detailed control over tone, dynamics, and effects, allowing consistent reproduction and customization of acoustic piano sounds for the experiments.}
    \label{fig:roland}
\end{figure}


\section{Multi-Scale Graph and Topological Analyses of Similarity Graphs}
\label{SI:multiscale_topological}

This section augments the single–scale network results with a cross–scale analysis that shows how musical structure emerges as we vary the similarity threshold applied to a fixed $k{=}6$ graph. Concretely, we treat the edge weights of the precomputed similarity graph as a scale parameter: high thresholds (strict) retain only the strongest links and reveal micro–motifs; progressively lower thresholds (looser) add weaker links and expose the fusion of motifs into meso– and macro–sections. Across this monotone edge–weight filtration, we (i) recompute community partitions and quantify their stability between successive levels via a best–match Jaccard index; (ii) visualize the consolidation pathways with a Sankey diagram whose columns are filtration levels and ribbons encode membership flow; and (iii) summarize the full graph (right–most scale) by complementary global and local signatures, including the normalized Laplacian spectrum and von Neumann entropy (global complexity), triangle $Z$–scores (motif enrichment), spectral gap $1-|{\lambda}_2|$ (integration/mixing), diffusion return–probability, role structure (core/bridge/periphery), and optional topological and geometric diagnostics (persistent homology up to $H_1$, Ollivier–Ricci curvature). Together, these readouts connect local repetition to long–range navigability and make explicit the timing and persistence of section formation across scales.

\subsection{Computational details}
We assume precomputed, time–indexed similarity graphs computed earlier (nodes = frames; undirected edges weighted by similarity $w_{ij}\in[0,1]$). All analyses operate directly on these graphs; no audio features are recomputed~\cite{EdelsbrunnerHarer2010,Ollivier2009,Bauer2021Ripser,Sankey1898ThermalEfficiency,RosvallBergstrom2010Alluvial}.

\textbf{Monotone weight filtration (no change to $k$).} To probe hierarchy without altering topology, we form a sequence of subgraphs by thresholding the \emph{existing} edge weights at $L$ levels $\tau_1>\cdots>\tau_L$ (quantiles of $\{w_{ij}\}$). Level $\ell$ keeps $E_\ell=\{(i,j): w_{ij}\ge \tau_\ell\}$ while preserving all nodes. This yields a nested filtration $G_1\subseteq G_2\subseteq\cdots\subseteq G_L$.

\textbf{Per–level network metrics.} For each $G_\ell$ we compute:
(i) size $|V|,|E|$, density, \#connected components and giant component size;
(ii) clustering coefficient (weighted where applicable), transitivity, and degree assortativity;
(iii) greedy–modularity communities and modularity $Q_\ell$;
(iv) small–worldness
\[
\sigma_\ell=\frac{C_\ell/C_\ell^{\mathrm{rand}}}{L_\ell/L_\ell^{\mathrm{rand}}},
\]
where $C_\ell$ is clustering and $L_\ell$ the (unweighted) average shortest path on the giant component, compared to Erdős–Rényi graphs with the same $n$ and $m$;
(v) degree–entropy (normalized), with $p_\ell(k)$ the degree histogram:
\[
H^{(\deg)}_\ell=-\sum_k p_\ell(k)\log p_\ell(k)\big/\log\big|\{k:p_\ell(k)>0\}\big|.
\]
We also compute a phase portrait across the filtration by plotting $(Q_\ell,\; H^{(\deg)}_\ell)$ with arrows $\ell\rightarrow \ell{+}1$.

\textbf{Community persistence across scales.} Let $\Pi_\ell$ be the partition at level $\ell$. We quantify cross–scale stability by a mean best–match Jaccard index
\[
J_{\ell\to \ell+1}=\frac{1}{|\Pi_\ell|}\sum_{C\in \Pi_\ell}\max_{D\in \Pi_{\ell+1}}\frac{|C\cap D|}{|C\cup D|},
\]
reported as a curve $J$ vs threshold. High/flat curves indicate persistent section identity; sharp rises indicate consolidation.

\textbf{Community–evolution Sankey.} Using the partitions $\{\Pi_\ell\}$ we draw a Sankey diagram~\cite{Sankey1898ThermalEfficiency,EdelsbrunnerHarer2010,RosvallBergstrom2010Alluvial} whose columns are thresholds $\tau_\ell$. Node heights equal community mass (\% of frames). To improve readability we retain the top-$K$ communities per level and aggregate the rest into \texttt{`Other'}. Ribbon widths are proportional to $|C\cap D|$ between adjacent levels; colors are consistent across levels.

The Sankey diagrams provide an intuitive view of how musical structure emerges as the similarity threshold is increased. Each vertical slice corresponds to a threshold: moving from left to right means keeping only stronger and stronger musical connections. At the left, most of the music appears as one large undifferentiated block (gray), indicating that everything is loosely connected. As the threshold tightens, this block fractures into smaller, distinct communities (colored bands), each representing a motif or recurring section. The flows show how these groups evolve: some communities remain stable, others divide into sub-motifs, and still others merge, revealing links between passages that share common material. By the final thresholds, the composition resolves into multiple well-formed clusters, some large and persistent, others smaller and fleeting. In engineering terms, this process resembles clustering in a noisy signal: at low resolution everything looks connected, but as the filter is raised, robust sub-networks appear while weaker, transient links fall away.

The Sankey representation makes visible the balance between stability and transformation in the music. A composition dominated by block-like flows and few crossovers emphasizes sectional repetition and hierarchy. In contrast, rich patterns of splitting and merging reveal a braided architecture, where motifs are reused, recombined, and linked across scales. Thus, the diagrams highlight whether musical form is rigid and modular, or flexible and interwoven, offering a direct visual correlate of structural creativity.

\textbf{Spectral fingerprint and entropy.} On the giant component of the full graph we compute the spectrum of the normalized Laplacian $\mathcal{L}$ (eigenvalues $\lambda_i\in[0,2]$) and report the von Neumann graph entropy
\[
H_{\mathrm{vN}}=-\sum_i p_i\log p_i,\qquad p_i=\lambda_i\Big/\sum_j \lambda_j,
\]
as a scalar summary of global structural complexity.

\textbf{Graph–signal smoothness.} Treating node attributes as signals on the graph, we report total variation
\[
\mathrm{TV}(x)=\sum_{(i,j)\in E} w_{ij}(x_i-x_j)^2,
\]
for two canonical signals: the time coordinate ($x_i=t_i$ if present; else node index) and node degree ($x_i=\deg(i)$). Lower TV indicates more coherent, motif–like neighborhoods.

\textbf{Roles (core/bridge/periphery).} We cluster standardized node features $[\deg,\;\text{strength},\;\text{clustering},\;\text{betweenness}]$ with $k\in\{2,3\}$ (K–means) to label structural roles. We also compute a \emph{bridge index}
\[
B_i=\frac{\sum_{j:\,c_j\neq c_i} w_{ij}}{\sum_{j} w_{ij}}\in[0,1],
\]
and plot $B_i$ over time to localize transitions (high values = inter–community connectors).

\textbf{Motif census and normalization.} We count triangles exactly and approximate 4–cycles; to assess enrichment we generate degree–preserving randomized graphs via double–edge swaps and report Z–scores relative to the null ensemble.

\textbf{Diffusion and mixing.} For the random–walk matrix $P=D^{-1}A$ we compute the mean return–probability curve
\[
r_t=\frac{1}{|V|}\operatorname{tr}\big(P^t\big),
\]
and the spectral gap $1-|\lambda_2(P)|$ (larger = faster mixing), capturing global navigability.

\textbf{Persistent homology (optional).} For graphs not exceeding a resource cap, we convert similarities to distances $D_{ij}=1-w_{ij}/\max w$ and compute persistent homology up to $H_1$ (ripser). We plot persistence diagrams/barcodes for $H_0$ (components) and $H_1$ (loops), interpreting long lifetimes as robust sections or cyclic structure across thresholds~\cite{Bauer2021Ripser}.

\textbf{Ricci curvature (optional).} We estimate Ollivier–Ricci curvature (transport parameter $\alpha=0.5$) per edge~\cite{Ollivier2009}. Positive curvature highlights cohesive interiors; negative curvature highlights bridges; we summarize by a histogram.

\textbf{Across–piece composites.} From per–piece CSVs we produce overlays of $J_{\ell\to\ell+1}$, $\sigma_\ell$, $Q_\ell$, and component count vs threshold, diffusion return–probability vs steps, and bar charts of motif Z–scores and spectral gaps.

\subsection{Analysis}

\begin{figure}[t]
  \centering
  \includegraphics[width=.7\linewidth]{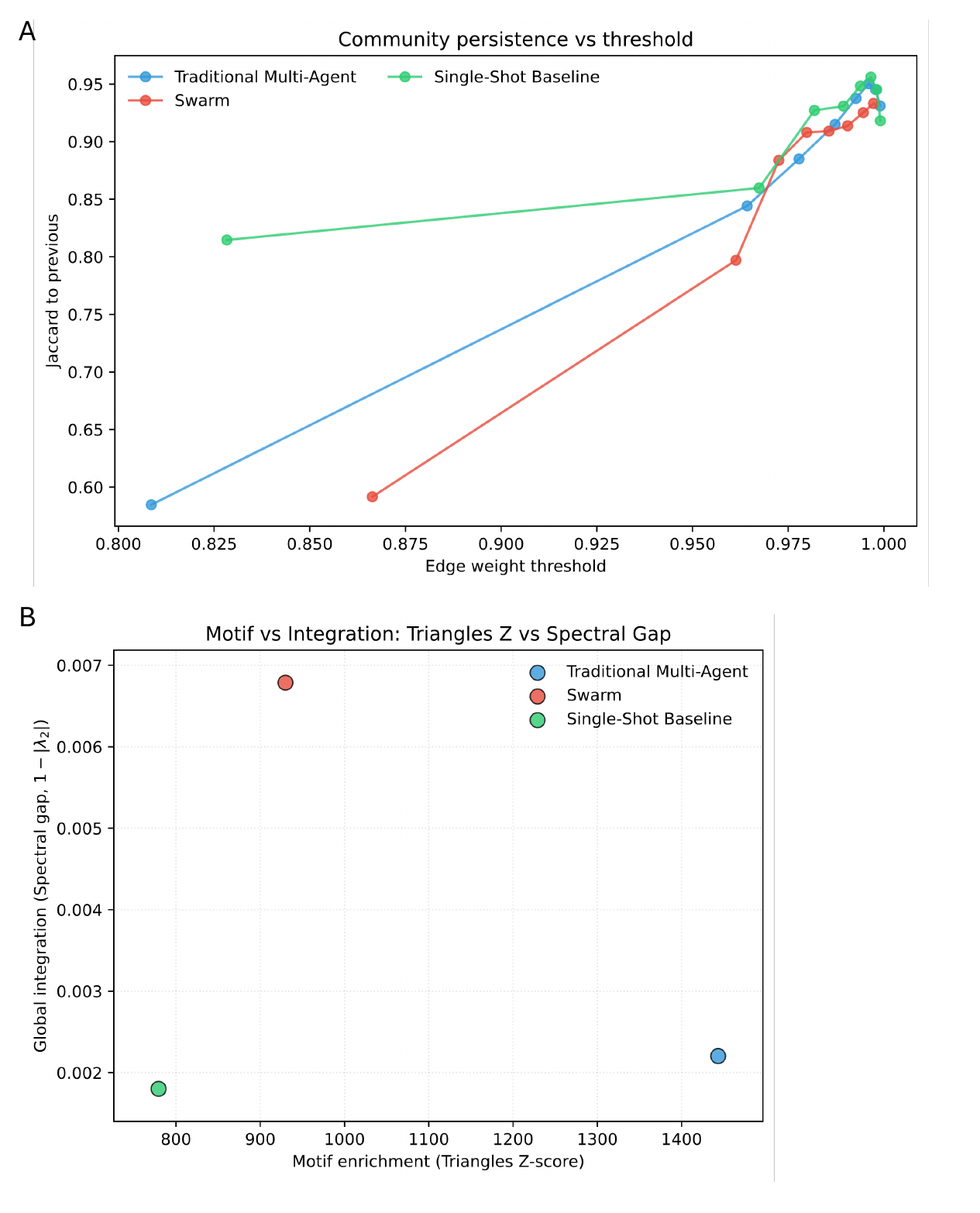}
  \caption{Two complementary views of structure.A: Community persistence (Jaccard similarity between successive edge–weight thresholds; left = strict (top-weight edges only); right = looser (weaker edges added)). Swarm (red) rises sharply from low to high values, indicating an active consolidation of micro-motifs into stable sections; Traditional Multi-Agent (blue) increases steadily, consistent with gradual consolidation; Single-Shot Baseline (green) is high and relatively flat at strict thresholds, reflecting rigid fragmentation that only locks into macro-sections near the loosest levels. B: Motif enrichment vs global integration for the full graph (right-most scale): the x-axis is triangle Z-score (richer local motifs to the right) and the y-axis is spectral gap $1-|{\lambda}_2|$ (better long-range integration upward). Swarm occupies the region of strong integration with solid motif enrichment; Traditional Multi-Agent shows the highest motif enrichment but lower integration; Single-Shot Baseline is weaker on both axes. Together these plots show that the swarm approach both stabilizes sections across scales and balances local coherence with global scaffolding.}
  \label{fig:fig19924}
\end{figure}

We quantify how musical form emerges and what kind of structure each system settles into (Fig.~\ref{fig:fig19924}). In Fig.~\ref{fig:fig19924}A, the community‐persistence curves track how stable the sectioning remains as we relax the similarity threshold (from strict, left, to looser, right). The swarm shows a pronounced consolidation phase—persistence climbs rapidly from low values to above 0.9—indicating that many small motifs cohere into durable sections across scales. Traditional Multi-Agent increases more steadily, consistent with gradual consolidation, whereas the Single-Shot Baseline starts high and flat, reflecting rigid fragmentation that only locks into macro-sections at the loosest thresholds. Fig.~\ref{fig:fig19924}B situates the three systems on the trade-off between local motif richness (triangle $Z$-score, $x$-axis) and global integration (spectral gap, $y$-axis). Swarm occupies the regime of strong integration while retaining substantial motif enrichment; Traditional emphasizes dense local motifs but with weaker long-range integration; Single-Shot is low on both. Taken together, these views show that the swarm not only stabilizes sections across scales but also balances local coherence with globally navigable structure.

\begin{figure}[p]
    \centering
    \includegraphics[angle=90,width=.4\textheight]{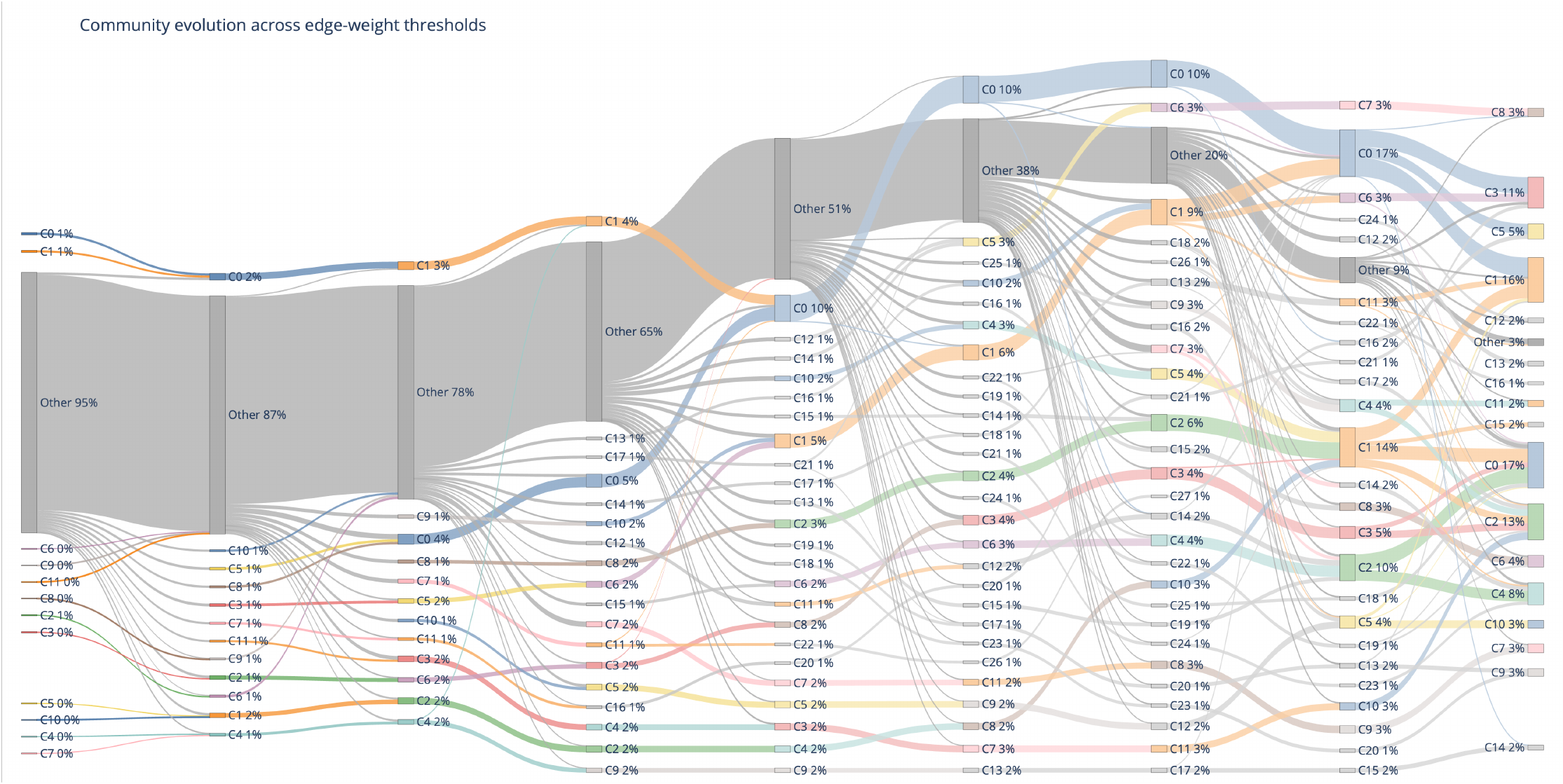}
    \caption{Sankey diagram of the traditional composition. Left shows only the strongest edges (highest threshold); right shows all connections as the threshold is relaxed. The flows indicate a block-like, hierarchical form: at very high thresholds many tiny fragments aggregate into \texttt{`Other'}, and as weaker links are admitted (moving right) these fuse late into a few large, stable motifs. Each vertical slice is an edge-weight filtration level; nodes are communities from greedy modularity at that level, and link widths are the number of frames transitioning between community labels across adjacent levels (i.e., membership overlap). Read the plot from left to right: the left side shows only the strongest connections, and progressively adding weaker links (moving right) causes fragments to join into larger groups. Straight bands mean motifs remain stable, splits show a motif breaking into parts, merges show separate motifs combining, and crossovers indicate reuse of material across communities.
}
    \label{fig:sankey_traditional}
\end{figure}

\begin{figure}[p]
    \centering
    \includegraphics[angle=90,width=.4\textheight]{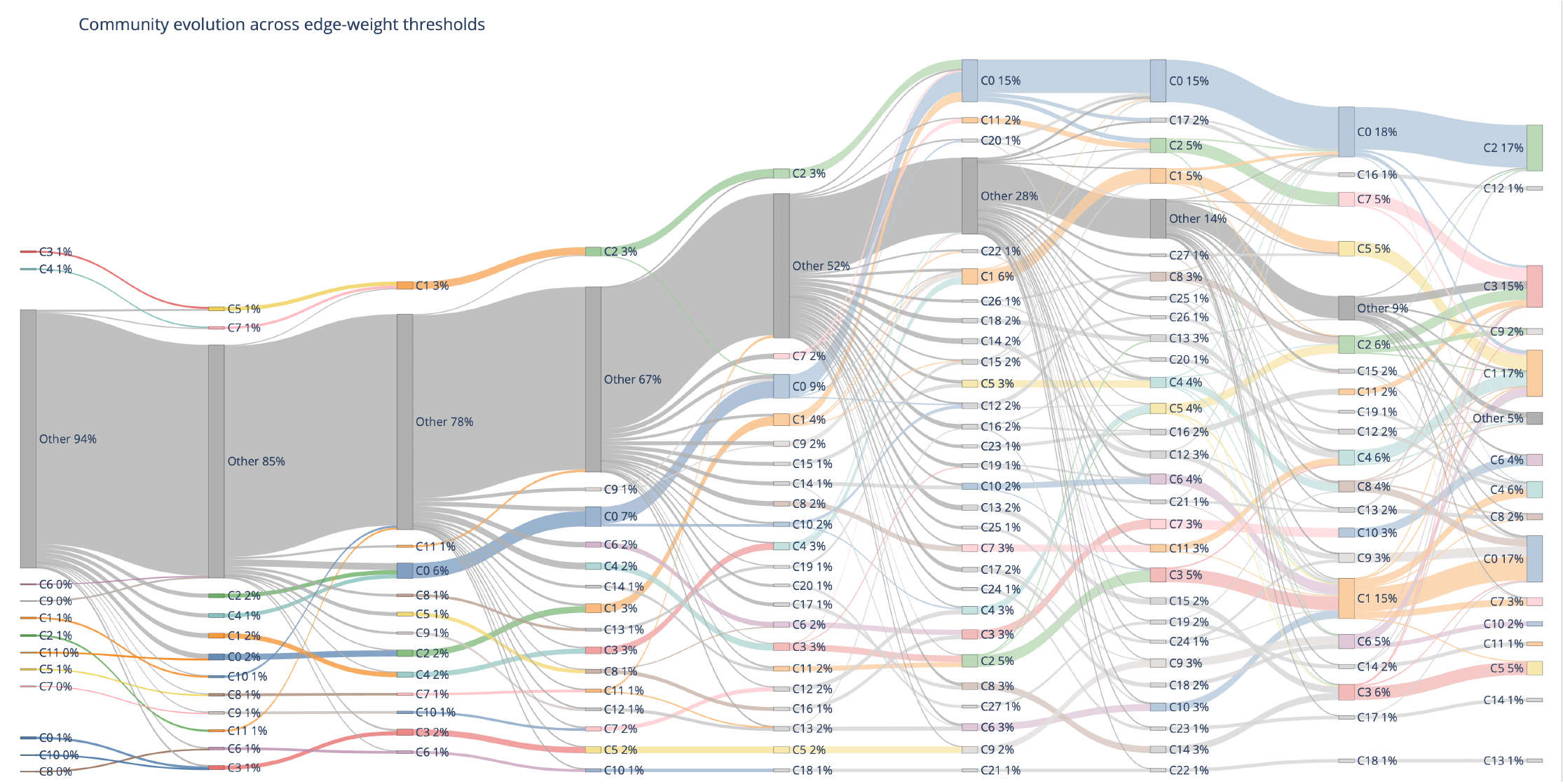}
    \caption{Sankey diagram of the swarm composition. Left uses only strongest edges; right includes all connections. Compared to traditional, the swarm exhibits earlier diversification and frequent splitting/merging, yielding multiple mid-sized motifs that braid and recombine as weaker links are added—an interwoven, small-world structure. Read left to right as decreasing threshold in an edge-weight filtration; communities are cluster assignments per level, and links quantify membership flow between consecutive partitions, revealing stability (straight bands) vs.\ reconfiguration (crossovers). Read the plot from left to right: the left side shows only the strongest connections, and progressively adding weaker links (moving right) causes fragments to join into larger groups. Straight bands mean motifs remain stable, splits show a motif breaking into parts, merges show separate motifs combining, and crossovers indicate reuse of material across communities.
}
    \label{fig:sankey_swarm}
\end{figure}

\begin{figure}[p]
    \centering
    \includegraphics[angle=90,width=.4\textheight]{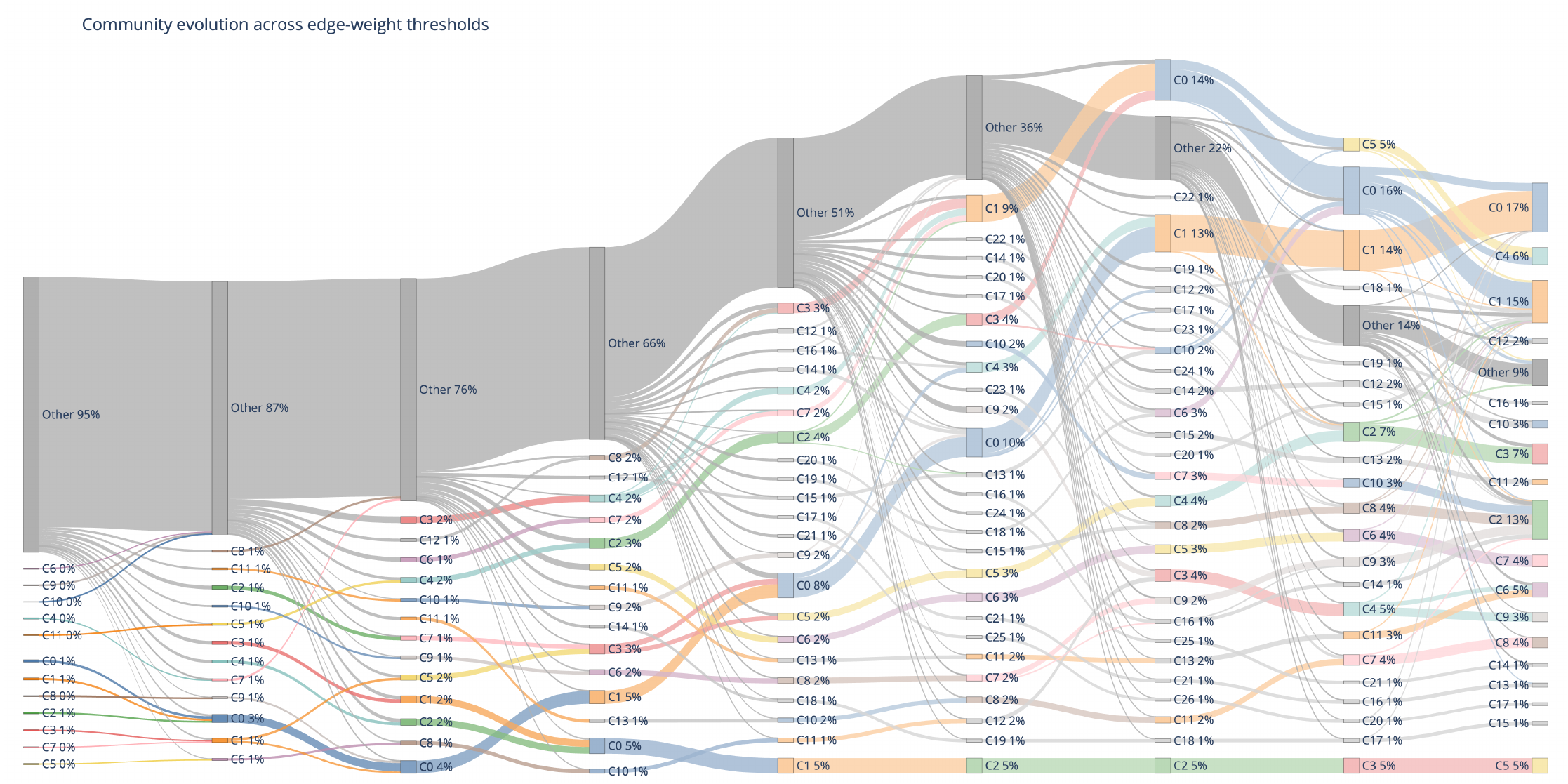}
    \caption{Sankey diagram of the single-shot composition. With strongest edges only (left), a few dominant motifs already appear and remain persistent as weaker connections are added (right), producing moderate diversity but limited recombination relative to the swarm. This filtration-based Sankey treats each threshold as a layer; community nodes come from modularity clustering per layer, and inter-layer links are proportional to the count of items retaining/altering community labels, providing a quantitative picture of motif persistence vs.\ branching. Read the plot from left to right: the left side shows only the strongest connections, and progressively adding weaker links (moving right) causes fragments to join into larger groups. Straight bands mean motifs remain stable, splits show a motif breaking into parts, merges show separate motifs combining, and crossovers indicate reuse of material across communities.
}
    \label{fig:sankey_singleshot}
\end{figure}

Figures~\ref{fig:sankey_traditional}--\ref{fig:sankey_singleshot} compare the evolution of musical communities in the traditional, swarm, and single-shot compositions. The traditional case shows a dominant block that fractures only late, producing a rigid and hierarchical structure. In contrast, the swarm case reveals early diversification into multiple medium-sized motifs, with frequent splitting and merging that indicate a braided and interwoven architecture. The single-shot composition lies in between: it maintains a few strong motifs with limited branching and weaker recombination, resulting in more variety than the traditional form but less structural richness than the swarm. Together, the three Sankey plots highlight the differences in how musical themes are organized, reused, and stabilized across compositional strategies.

\end{document}